\documentclass{article}

\PassOptionsToPackage{numbers, compress}{natbib}

\usepackage[preprint]{neurips_2026}

\usepackage[utf8]{inputenc} 
\usepackage[T1]{fontenc}    
\usepackage[hidelinks]{hyperref}       
\usepackage{url}            
\usepackage{booktabs}       
\usepackage{amsfonts}       
\usepackage{nicefrac}       
\usepackage{microtype}      
\usepackage{xcolor}         
\usepackage{graphicx}       
\usepackage{amsmath}
\usepackage{multirow}
\usepackage{enumitem}
\usepackage{cleveref}
\usepackage{colortbl}       
\usepackage{algorithm}
\usepackage{algorithmic}
\usepackage{pifont}
\usepackage{amssymb}
\usepackage{wrapfig}
\usepackage{bbm}
\usepackage{tcolorbox}

\title{

  \oursfull{}: Benchmarking the Expert-Lay Gap in Medical Vision-Language Models

}

\author{
  \textbf{Han Jang\textsuperscript{1,2,$\dagger$}},
  \textbf{Junhyeok Lee\textsuperscript{1,3,$\dagger$}},
  \textbf{Songsoo Kim\textsuperscript{2}}, 
  \textbf{Chae Young Lim\textsuperscript{2,4}},\\
  \textbf{Hyeonjin Goh\textsuperscript{2}}, 
  \textbf{Heeseong Eum\textsuperscript{1,3}}, 
  \textbf{Kyu Sung Choi\textsuperscript{1,2,3,*}}
\\
  \small \textsuperscript{1}Seoul National University \quad
  \small \textsuperscript{2}Seoul National University Hospital \\
  \textsuperscript{3}Seoul National University College of Medicine \\
  \small \textsuperscript{4}Sungkyunkwan University School of Medicine \\
  \small \texttt{hanjang, jhlee0619, ent1127}@snu.ac.kr
}
\newcommand{\blfootnote}[1]{%
  \begingroup
  \renewcommand\thefootnote{}\footnote{#1}%
  \addtocounter{footnote}{-1}%
  \endgroup
}

\definecolor{customblue}{RGB}{0, 0, 0}

\newcommand{\oursdata}{\textsc{MedLayXPlain}-122K}
\newcommand{\oursfull}{\textsc{MedLayXPlain}}

\newcommand{\ourseval}{\textsc{MedLayEval}}

\newcommand{\hover}{\textsc{HOVER}}
\definecolor{figpurple}{RGB}{180,0,180}
\tcbuselibrary{skins,breakable}
\usepackage{float}  
\newcommand{\ehigh}[1]{\textcolor{figpurple}{\textbf{#1}}}
\newcommand{\lhigh}[1]{\textcolor{figblue}{\textbf{#1}}}
\definecolor{figblue}{RGB}{0,70,200}

\begin{document}
\maketitle
\blfootnote{$^{\dagger}$Equal contribution. \quad $^{*}$Corresponding author}
\vspace{-24pt}

\begin{abstract}

Medical Vision-Language Models (Med-VLMs) achieve strong expert-level performance, yet their ability to generate patient-accessible descriptions remains underexplored.
With the 21st Century Cures Act now mandating immediate patient access to diagnostic imaging results, evaluating whether Med-VLMs can bridge this Expert-Lay Gap is both urgent and clinically consequential for patient education and shared decision-making.
To this end, we introduce \oursfull{}, the first large-scale multimodal benchmark and evaluation framework for Medical Lay Language Generation (MLLG).
\oursdata{} provides 122,789 region-grounded samples across 8 imaging modalities from 12 publicly source datasets, each comprising a medical image with paired expert and lay captions anchored in a three-level Unified Medical Language System (UMLS) ontology hierarchy spanning 7 semantic groups, 43 semantic types, and 2,411 medical concepts.
Lay captions are constructed via Hierarchical Ontology-VErified Refinement (\hover{}), a three-step pipeline combining patient-centric vocabulary mapping, LLM-based constrained rewriting, and cross-model visual verification to enforce semantic equivalence while preventing hallucination.
We further introduce \ourseval{}, a lightweight 3B evaluator distilled from a 27B verifier that scores expert-lay alignment across five clinically grounded attributes, addressing the poor correlation between standard NLG metrics and clinical judgment.
Benchmarking 33 VLMs on \oursdata{} reveals a systematic Expert-Lay Gap: medical VLMs achieve strong expert captioning but suffer significant lay-register degradation, while general-purpose VLMs produce more accessible language yet lack clinical precision, confirming that neither current paradigm adequately serves patient-facing communication.

\end{abstract}

\section{Introduction}

Medical Vision-Language Models (Med-VLMs) have driven rapid progress in clinical image understanding, achieving strong performance on tasks ranging from radiology report generation to medical visual question answering~\cite{moor2023med,moor2023foundation,li2023llava,xie2024medtrinity,sellergren2025medgemma,yao2026medical}.
Yet these models are trained almost exclusively on expert-level corpora such as radiology reports~\cite{johnson2019mimic}, pathology annotations~\cite{ding2025multimodal}, and biomedical literature~\cite{ruckert2024rocov2,lozano2025biomedica}, embedding visual features into a technical lexicon designed for clinician-to-clinician communication.
Meanwhile, the regulatory landscape is shifting toward direct patient engagement.
The United States 21st Century Cures Act now mandates immediate patient access to diagnostic imaging results~\cite{pollock2024patient,steitz2023perspectives}, yet 84.7\% of patients report needing plain-language definitions to comprehend their reports~\cite{vincoff2022patient}, and immediate release without such support has been linked to elevated patient anxiety and confusion~\cite{bhalla2024contemporary, dambrino2023unsolicited}.
Med-VLMs could serve a critical role in this emerging patient-facing workflow by translating complex imaging findings into accessible language, yet whether current models can actually generate patient-accessible descriptions remains largely underexplored.

Existing Med-VLM datasets, from visual question answering~\cite{lau2018dataset,liu2021slake,he2020pathvqa,yao2026medical} to large-scale captioning~\cite{ruckert2024rocov2,xie2024medtrinity,lozano2025biomedica}, have driven substantial progress in expert-level medical image understanding.
However, these datasets uniformly assume expert-level clinical text as the target output.
Whether Med-VLMs can express the same clinical findings in language that patients can understand remains unevaluated.
This patient-facing direction has been explored in the text-only domain through Medical Lay Language Generation (MLLG), including neural text simplification~\cite{shardlow2019neural} and biomedical lay summarization~\cite{goldsack2022biolaysumm,goldsack2024overview}.
Yet all existing MLLG methods operate on text without visual grounding.
MedLayBench-V~\cite{jang2026medlaybench} takes a first step toward the multimodal setting with 80K expert-lay pairs, but lacks region-of-interest grounding, a dedicated evaluator, and systematic VLM benchmarking.

The absence of suitable benchmark data is compounded by a parallel gap in evaluation methodology.
Standard NLG metrics such as BLEU~\cite{papineni2002bleu} and ROUGE~\cite{lin2004rouge} inherently penalize the vocabulary shifts that successful simplification requires~\cite{zhao2024x}, while medical-specific LLM-as-a-judge metrics such as GREEN~\cite{ostmeier2024green}, RaTEScore~\cite{zhao2024ratescore}, and RadEval~\cite{xu2025radeval} target expert-level factual correctness without assessing patient accessibility, and are predominantly scoped to chest X-ray reports with substantial inference costs.
No existing metric jointly evaluates clinical fidelity and lay readability across medical imaging modalities.
We refer to this unexamined dimension as the Expert-Lay Gap, and argue that closing it requires three components currently absent from the field.
First, a large-scale multimodal benchmark with region-grounded expert-lay pairs anchored in structured medical ontology.
Second, a lightweight evaluator that jointly captures clinical fidelity and patient accessibility across modalities.
Third, systematic benchmarking of existing Med-VLMs to quantify the gap.

\begin{figure}[!t]
  \centering
  \includegraphics[width=0.9\linewidth]{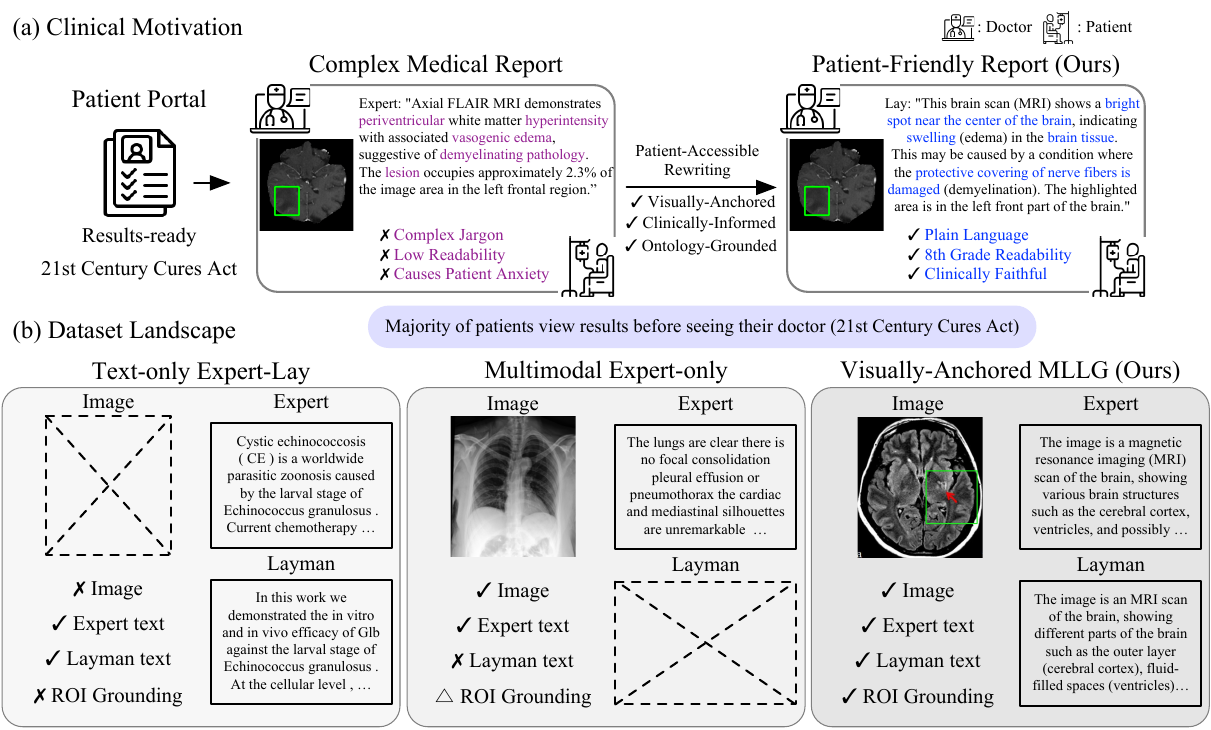}
    \caption{
    \textbf{Motivation and positioning of \oursfull{}.}
    \textbf{(a)} Clinical motivation for patient-accessible rewriting of medical reports.
    \textbf{(b)} Existing resources offer text-only expert-lay pairs (left) or expert-only multimodal data (center). \oursdata{} combines all four elements (right).
    }
    \label{fig:motivation}
\end{figure}

\begin{figure*}[!t]
    \centering
    \includegraphics[width=0.97\textwidth]{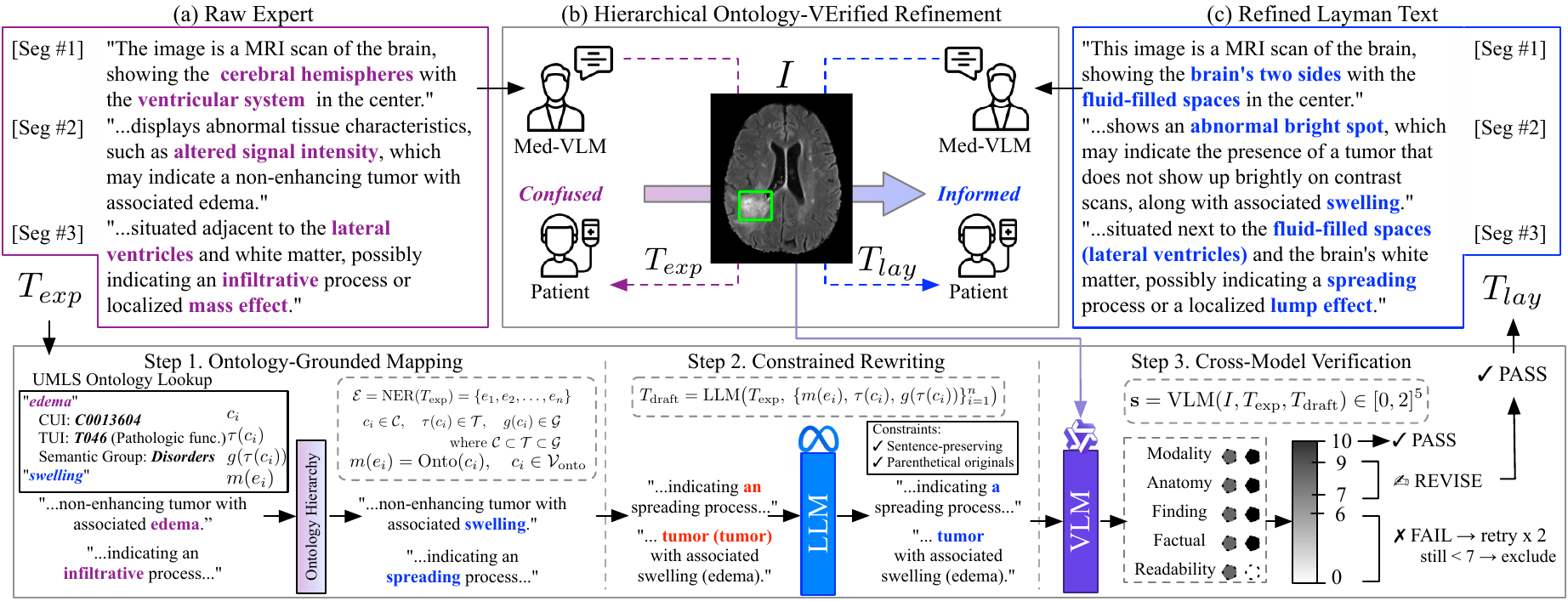}
    \vspace{-4pt}
    \caption{\textbf{Overview of the proposed \hover{} pipeline.}
    \textbf{(a)} Raw expert caption with jargon (\textcolor{figpurple}{highlighted}).
    \textbf{(b)} Step~1 maps medical entities to patient-friendly terms via three-level UMLS ontology ($\mathcal{C} \subset \mathcal{T} \subset \mathcal{G}$). Step~2 generates a constrained lay draft with full ontology context. Step~3 performs cross-model visual verification (pass/revise/fail).
    \textbf{(c)} Refined lay caption (\textcolor{blue}{highlighted}).}
    \label{fig:architecture}
    \vspace{-8pt}
\end{figure*}

To this end, we introduce \oursfull{}, a large-scale multimodal benchmark and evaluation framework for expert-lay medical image explanation (\Cref{fig:motivation}).
\oursdata{} provides 122,789 region-grounded samples spanning 8 imaging modalities from 12 publicly available datasets, each containing a medical image with paired expert and lay captions anchored in a three-level Unified Medical Language System~(UMLS)~\cite{bodenreider2004unified} ontology hierarchy (7 semantic groups, 43 semantic types, 2,411 medical concepts) to preserve medical semantics throughout the expert-to-lay translation process.
Lay captions are constructed via Hierarchical Ontology-VErified Refinement (\hover{}), a three-step pipeline (\Cref{fig:architecture}) that combines patient-centric vocabulary mapping, LLM-based constrained rewriting, and cross-model visual verification to enforce semantic equivalence while preventing hallucination.
To address the evaluation gap, we propose \ourseval{}, a lightweight 3B evaluator distilled from a 27B verifier that jointly scores clinical fidelity and patient accessibility across five attributes and all 8 imaging modalities.
Finally, we benchmark 33 state-of-the-art VLMs spanning medical and general-purpose categories on \oursdata{}, establishing the first measurement of the Expert-Lay Gap across model families, scales, and training paradigms.

Our contributions are as follows:
\vspace{-7pt}
\begin{enumerate}[leftmargin=*]
    \item We introduce \oursdata{}, the first large-scale multimodal benchmark with 122K region-grounded expert-lay caption pairs across 8 imaging modalities, anchored in UMLS ontology and verified through cross-model visual verification.
    \item We propose \ourseval{}, a reference-free 3B multimodal evaluator distilled from a 27B verifier that jointly scores clinical fidelity and patient accessibility across five ontology-aligned attributes.
    \item We benchmark 33 VLMs on \oursdata{} and formalize the Expert-Lay Gap, revealing that general-purpose VLMs outperform medical VLMs on patient-accessible description despite lacking domain-specific training.
\end{enumerate}
\vspace{-7pt}

\vspace{-4pt}
\section{Related Work}
\label{sec:related}
\vspace{-4pt}

\paragraph{Medical Vision-Language Models.}
Recent Med-VLMs span both contrastive and generative paradigms.
Contrastive models such as PubMedCLIP~\cite{eslami2023pubmedclip} and BiomedCLIP~\cite{zhang2025multimodal} learn joint image-text embeddings from biomedical literature, while generative Med-VLMs including LLaVA-Med~\cite{li2023llava}, Med-Flamingo~\cite{moor2023med}, and MedGemma~\cite{sellergren2025medgemma} extend this paradigm to open-ended tasks such as report generation and medical VQA.
General-purpose VLMs such as Qwen-VL~\cite{bai2025qwen3} and InternVL~\cite{zhu2025internvl3} exhibit competitive medical understanding without domain-specific training.
However, medical VLMs are optimized on expert-level corpora~\cite{johnson2016mimic,ruckert2024rocov2, lozano2025biomedica,xie2024medtrinity}, confining their output to the professional register, and general-purpose VLMs have not been evaluated on patient-facing tasks.

\paragraph{Medical Lay Language Generation.}
Translating clinical jargon into patient-accessible language has been studied under Medical Lay Language Generation (MLLG), spanning neural text simplification~\cite{shardlow2019neural} and biomedical lay summarization~\cite{goldsack2022making,goldsack2024overview,xiao2025overview}, with benchmarks such as BioLaySumm and MedAgentBoard~\cite{zhu2025medagentboard} driving progress.
Magical~\cite{liao2025magical} proposes an asymmetric LoRA architecture with semantic invariance constraints for factually faithful register transfer.
However, existing MLLG methods operate exclusively on text without visual grounding, and naive LLM-based simplification introduces factual errors in up to 7\% of clinical reports~\cite{prucker2025prospective}, highlighting the need for visually grounded and structured verification.

\paragraph{Multimodal Benchmarks and Knowledge Grounding.}
Large-scale medical image-text datasets have driven Med-VLM development across diverse tasks, from visual question answering~\cite{lau2018dataset,liu2021slake,he2020pathvqa} to captioning~\cite{ruckert2024rocov2,lozano2025biomedica,xie2024medtrinity}.
However, these resources uniformly assume expert-level output and often suffer from quality issues such as LLM-generated error cascading, 3D-to-2D slice inflation, and modality mislabeling~\cite{xie2024medtrinity}.
Layman's RRG~\cite{zhao2024x} incorporates lay descriptions but is limited to chest X-rays with restricted scale and hallucination risks~\cite{moor2023med,zhu2025can}.
MedLayBench-V~\cite{jang2026medlaybench} introduced the first multimodal expert-lay benchmark with 80K samples across 7 modalities via UMLS-grounded ontology mapping~\cite{bodenreider2004unified}, showing that knowledge-constrained rewriting preserves retrieval performance while naive simplification severely degrades it.
Despite these advances, limitations remain, including the absence of region-of-interest grounding and systematic benchmarking of existing VLMs on the lay generation task.
We address the data gap through a large-scale region-grounded benchmark with cross-model visual verification and systematic benchmarking of existing VLMs (\Cref{app:benchmark}).

\paragraph{Evaluation of Expert-Lay Alignment.}
Evaluating expert-to-lay translation remains challenging.
Standard n-gram metrics~\cite{papineni2002bleu,lin2004rouge,banerjee2005meteor} inherently penalize the vocabulary shifts required for simplification~\cite{zhao2024x}, as replacing clinical terms with plain-language equivalents reduces lexical overlap with expert references even when clinical meaning is preserved.
Medical-specific metrics have adopted the LLM-as-a-judge paradigm to address this limitation.
GREEN~\cite{ostmeier2024green} leverages language models to identify and explain clinically significant errors, RaTEScore~\cite{zhao2024ratescore} uses entity-aware matching robust to medical synonyms and negation, and RadEval~\cite{xu2025radeval} provides a multi-dimensional evaluation framework with interpretable sub-scores.
However, these metrics evaluate exclusively for expert-level factual correctness without assessing patient accessibility, are predominantly designed for chest X-ray reports, and rely on 7B or larger language models that incur substantial inference costs~\cite{wang2023automated}.
These gaps motivate the development of a dedicated lightweight evaluator that jointly assesses clinical fidelity and patient accessibility across diverse imaging modalities.

\section{Method}
\label{sec:method}

We present \oursfull{}, a framework consisting of two components: (1)~\oursdata{}, a large-scale region-grounded benchmark with paired expert-lay captions anchored in structured medical ontology~(\S\ref{sec:benchmark}), and (2)~\ourseval{}, a lightweight multimodal evaluator for reference-free expert-lay alignment scoring~(\S\ref{sec:medlayeval}). \Cref{fig:architecture} illustrates the data construction pipeline.

\subsection{\oursdata{}: Benchmark Construction}
\label{sec:benchmark}

We construct \oursdata{} through a three-step pipeline that curates high-quality region-grounded samples from 12~source datasets and enriches them with UMLS ontology grounding and verified lay descriptions, yielding 122,789 paired samples across 8 imaging modalities (\Cref{alg:hover}).

\paragraph{Quality-Driven Source Curation.}
\label{sec:curation}

We aggregate medical image-caption pairs from 12~publicly available source datasets spanning 8~imaging modalities, including pathology, MRI, CT, X-Ray, PET, dermoscopy, endoscopy, and ultrasound.
For sources where region-of-interest~(ROI) annotations or detailed textual descriptions are insufficient, we supplement annotations using the multigranular image-ROI-description triplets from MedTrinity-25M~\cite{xie2024medtrinity}, which provides bounding-box-level ROI localization and expert captions generated via retrieval-augmented multimodal LLMs.

To ensure benchmark quality across all 12 source datasets, we apply three filtering criteria:
(i)~captions shorter than 20~words are removed as they lack sufficient diagnostic detail;
(ii)~3D volumetric datasets with more than 100~slices per volume are excluded to prevent redundant near-duplicate samples from a single scan;
(iii)~samples with missing or inconsistent modality labels are discarded.
For approximately 30K samples originally labeled as mixed modality, we assign modality labels by matching keywords in the caption text and remove 806~non-medical samples (e.g., protein structures).

\paragraph{UMLS Ontology Grounding.}
\label{sec:umls}
To establish a structured semantic layer over the curated samples, we ground each expert caption in the UMLS~\cite{bodenreider2004unified}, the most comprehensive biomedical ontology linking clinical terminology to semantically meaningful concept units.
We perform Named Entity Recognition~(NER) and entity linking using scispaCy~\cite{neumann2019scispacy} with the UMLS entity linker, mapping expert caption spans to UMLS concepts.
Each extracted mention is resolved to a Concept Unique Identifier~(CUI), which is further associated with a Type Unique Identifier~(TUI) and aggregated into one of 7~coarse-grained Semantic Groups~(e.g., \textit{Anatomy}, \textit{Disorders}, \textit{Procedures}).
This three-level hierarchy serves a dual purpose.
First, it provides concept-level anchoring for lay caption generation, where CUI-to-lay-term mappings enable systematic translation of expert jargon into patient-accessible equivalents.
Second, it supplies structured semantic metadata for downstream evaluation, enabling attribute-level scoring aligned with ontology categories~(\S\ref{sec:medlayeval}).
Across the full benchmark, scispaCy extracts 2,306~unique CUIs spanning 43~semantic types across 7~semantic groups (\Cref{fig:statistics}d).

\paragraph{Hierarchical Ontology-Verified Refinement.}
\label{sec:hover}

Generating lay descriptions by naive LLM rewriting risks hallucination, semantic drift, and omission of clinically important details~\cite{prucker2025prospective,alabed2026large}. We address this through \hover{}, a three-step pipeline (\Cref{fig:architecture}b) that enforces strict semantic equivalence between expert and lay captions via deterministic ontology mapping, LLM-based constrained rewriting, and cross-model visual verification to ensure clinical-semantic fidelity.

\textbf{Step 1: Ontology-Grounded Mapping.}
Given an expert caption $T_{\text{exp}}$, we extract medical entities $\mathcal{E} = \text{NER}(T_{\text{exp}}) = \{e_1, \ldots, e_n\}$ using scispaCy and resolve each to a UMLS concept through the three-level ontology hierarchy.
For example, ``edema'' is resolved to CUI $c_i$ = C0013604, TUI $\tau(c_i)$ = T046 (Pathologic Function), and Semantic Group $g(\tau(c_i))$ = \textit{Disorders}:
\begin{equation}
  c_i \in \mathcal{C}, \quad \tau(c_i) \in \mathcal{T}, \quad g(\tau(c_i)) \in \mathcal{G}, \quad \text{where } \mathcal{C} \subset \mathcal{T} \subset \mathcal{G}
\end{equation}
For each resolved concept, we retrieve a patient-friendly synonym primarily from the Consumer Health Vocabulary (CHV)~\cite{zeng2006exploring}, with fallback to four additional UMLS source vocabularies (Appendix~\ref{app:vocab}).
The resulting deterministic mapping is formulated as:
\begin{equation}
  m(e_i) = \text{Onto}(c_i), \quad c_i \in \mathcal{V}_{\text{onto}}
\end{equation}
These mappings are passed to the LLM in Step~2 as soft references, where the model adopts each mapping only when it improves patient readability (\Cref{app:chv-applied}).

\textbf{Step 2: Constrained Rewriting.}
Step~1 produces word-level mappings $\{m(e_i)\}_{i=1}^{n}$, but directly substituting these into the expert caption would yield ungrammatical or unnatural text (e.g., ``an infiltrative'' becomes ``an spreading'').
We therefore prompt Llama-3.1-70B-Instruct~\cite{grattafiori2024llama} with the expert caption $T_{\text{exp}}$, the ontology mappings, and their associated semantic hierarchy ($c_i, \tau(c_i), g(\tau(c_i))$), instructing the model to selectively adopt each mapping while preserving the ontology-grounded semantic structure:
\begin{equation}
  T_{\text{draft}} = \text{LLM}\bigl(T_{\text{exp}},\; \{m(e_i),\, \tau(c_i),\, g(\tau(c_i))\}_{i=1}^{n}\bigr)
\end{equation}
The generation is subject to three constraints: sentence-count preservation ($|\text{sent}(T_{\text{draft}})| = |\text{sent}(T_{\text{exp}})|$), parenthetical retention of original terms on first mention (e.g., ``swelling (edema)''), and ontology-aware substitution that prevents replacements across semantic group boundaries.

\textbf{Step 3: Cross-Model Visual Verification.}
Each draft $T_{\text{draft}}$ is verified against the original image $I$, ROI bounding box $b$, and expert caption $T_{\text{exp}}$ using Qwen3.5-27B~\cite{bai2025qwen3} as an independent visual judge from a different model family than the generator:
\begin{equation}
  \mathbf{s} = \text{VLM}(I, b, T_{\text{exp}}, T_{\text{draft}}) \in [0, 2]^5
\end{equation}
The verifier scores five ontology-aligned attributes (modality consistency, anatomical accuracy, finding completeness, factual correctness, and lay readability), yielding a total score $S = \sum_{j=1}^{5} s_j$ out of~10.
Based on this score, each sample receives a three-tier verdict.
\textsc{Pass}~($S$=10) accepts the draft as-is.
\textsc{Revise}~($7 \leq S < 10$) prompts the verifier to produce a corrected lay caption guided by its own attribute-level feedback, which is then re-verified.
\textsc{Fail}~($S < 7$) triggers up to two additional revision loops, and the sample is excluded if it remains below threshold.
Using Llama for generation and Qwen for verification mitigates the self-confirmation bias inherent in single-model pipelines~\cite{panickssery2024llm} (Appendix~\ref{app:cross_family}).
The verifier additionally identifies expert caption errors inherited from upstream annotations, which are corrected as described in Appendix~\ref{app:hover}.
Qualitative examples across all 8 modalities are provided in \Cref{app:qualitative}.

\subsection{\ourseval{}: Multimodal Lay Quality Evaluator}
\label{sec:medlayeval}

Existing evaluation metrics for medical text generation either require ground-truth references or operate exclusively on text without considering the source image.
We propose \ourseval{}, an annotation-free multimodal evaluator that scores expert-to-lay translation quality by jointly examining the medical image, expert caption, and generated lay caption.

\paragraph{Formulation.}
Given a medical image $I$, an expert caption $T_{\text{exp}}$, and a candidate lay caption $\hat{T}_{\text{lay}}$ produced by any VLM, \ourseval{} predicts five continuous quality scores:
\begin{equation}
  \mathbf{s} = f_{\theta}(I, T_{\text{exp}}, \hat{T}_{\text{lay}}) \in [0, 1]^5
\end{equation}
where $\mathbf{s} = (s_{\text{mod}}, s_{\text{anat}}, s_{\text{find}}, s_{\text{fact}}, s_{\text{read}})$ assesses modality consistency, anatomical accuracy, finding completeness, factual correctness, and lay readability.
The overall quality score is $S = \frac{1}{5}\sum_{j} s_j$.
These five attributes are designed to capture both clinical fidelity (first four) and patient accessibility (readability), enabling interpretable diagnosis of where a translation fails.
Incorporating the source image enables the evaluator to verify whether a generated lay description faithfully represents the actual imaging finding, an assessment that text-only metrics fundamentally cannot perform.

\begin{wrapfigure}{r}{0.5\textwidth}
  \vspace{-12pt}
  \centering
  \includegraphics[width=0.5\textwidth]{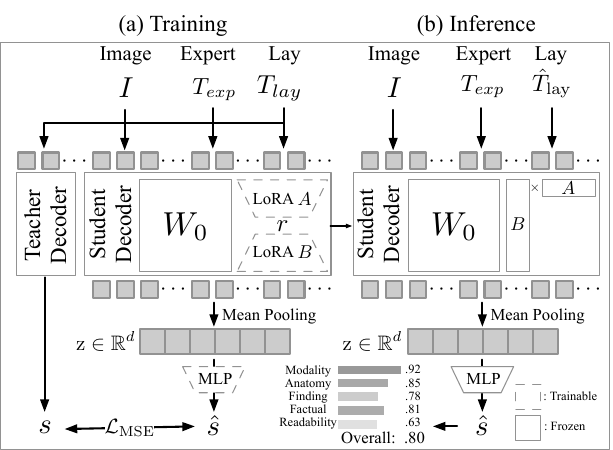}
  \caption{\textbf{\ourseval{} architecture.}
  (a)~Training: the student evaluator is distilled from a teacher verifier via per-attribute weighted MSE.
  (b)~Inference: the evaluator predicts five attribute scores from an image, expert caption, and candidate lay caption. LoRA weights are merged at inference.}  \label{fig:medlayeval}
  \vspace{-10pt}
\end{wrapfigure}

\paragraph{Architecture.}
We build \ourseval{} on Qwen2.5-VL-3B-Instruct~\cite{bai2025qwen3}, adapting it from a generative VLM into a regression-based evaluator through two modifications.
First, we apply Low-Rank Adaptation (LoRA)~\cite{hu2022lora} to the backbone's weight matrices.
For each pretrained weight matrix $W_0 \in \mathbb{R}^{d \times k}$, LoRA reparameterizes the forward pass as:
\begin{equation}
  \mathbf{h} = (W_0 + \frac{\alpha}{r} BA)\mathbf{x}, \quad B \in \mathbb{R}^{d \times r},\; A \in \mathbb{R}^{r \times k}
\end{equation}
where $W_0$ remains frozen, $A$ is initialized from $\mathcal{N}(0, \sigma^2)$, $B$ is initialized to zero so that $\Delta W = 0$ at the start of training, and $\alpha / r$ controls the magnitude of the adaptation.
We apply LoRA to all attention and feed-forward projections, enabling the backbone to adapt its visual and textual representations for quality assessment while preserving pretrained knowledge.

Second, we replace the language modeling head with a regression head $h_\phi$ that maps the backbone's hidden representation to five continuous attribute scores.
Given the multimodal input sequence $\mathbf{x} = [I; T_{\text{exp}}; \hat{T}_{\text{lay}}]$, the LoRA-adapted backbone produces contextualized hidden states $\mathbf{H} = [\mathbf{h}_1, \ldots, \mathbf{h}_L] \in \mathbb{R}^{L \times d}$.
We aggregate these into a fixed-dimensional representation via attention-masked mean pooling:
\begin{equation}
  \mathbf{z} = \frac{\sum_{l=1}^{L} \alpha_l \, \mathbf{h}_l}{\sum_{l=1}^{L} \alpha_l} \in \mathbb{R}^{d}
\end{equation}
where $\alpha_l \in \{0, 1\}$ is the attention mask.
The pooled representation is then projected through a two-layer MLP with GELU activation~\cite{hendrycks2016gaussian} and dropout regularization:
\begin{equation}
  \mathbf{s} = \sigma\bigl(W_2 \cdot \text{GELU}(\text{Dropout}(W_1 \, \mathbf{z} + \mathbf{b}_1)) + \mathbf{b}_2\bigr) \in [0, 1]^5
\end{equation}
where $W_1 \in \mathbb{R}^{256 \times d}$, $W_2 \in \mathbb{R}^{5 \times 256}$, and $\sigma$ is the element-wise sigmoid function.
This regression-based design directly predicts continuous scores without text generation, avoiding the mode collapse observed when training models to output score tokens (\S\ref{app:medlayeval_ablation}).

\paragraph{Training via Score Distillation.}
We distill the scoring capability of the 27B verifier used in \hover{} Step~3 (\S\ref{sec:hover}) into \ourseval{} through supervised regression.
Each teacher attribute score is normalized from $[0,2]$ to $[0,1]$ and used as the training target.
To ensure coverage across the full score range, training data combines four complementary sources:
(1)~\textsc{pass}-verdict lay captions with high teacher scores;
(2)~pre-revision drafts from \textsc{revise}-verdict samples, providing mid-range scores;
(3)~\textsc{fail}-verdict samples with naturally low clinical scores;
and (4)~augmented negatives including expert-as-lay copies and generic captions (details in \Cref{app:medlayeval_ablation}).
The model is optimized with per-attribute inverse-frequency weighted MSE loss.
Let $s_j$ and $\hat{s}_j$ denote the predicted and teacher score for the $j$-th attribute, respectively:
\begin{equation}
  \mathcal{L} = \sum_{j=1}^{5} w_j \cdot \bigl(s_j - \hat{s}_j\bigr)^2, \quad w_j = \frac{N}{n_j}
\end{equation}
where $n_j$ denotes the number of training samples sharing the same target score value as $\hat{s}_j$ for the $j$-th attribute.
This weighting assigns higher loss to underrepresented score values, preventing the model from defaulting to the majority class.
The distilled evaluator achieves Pearson $r = 0.880$ against the teacher on held-out validation (\Cref{app:bench}).

\begin{figure*}[!ht]
    \centering
    \includegraphics[width=\textwidth]{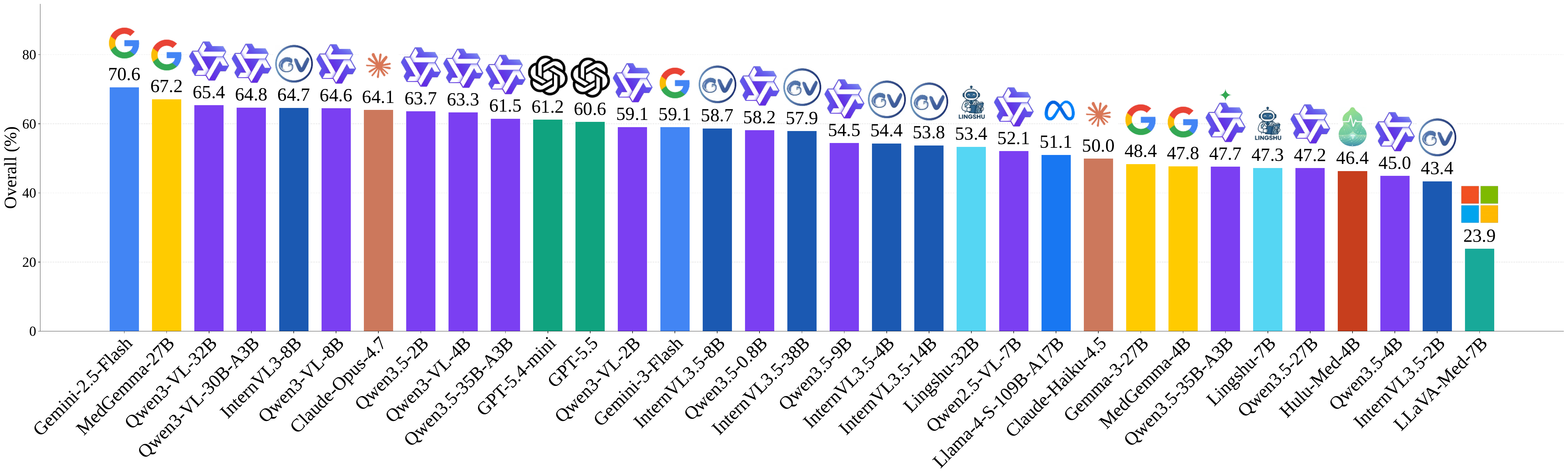}
    \caption{\textbf{VLM leaderboard on \oursdata{}, ranked by \ourseval{} overall score (lay register, $n$\,=\,5,000).}
    A 46.7-point spread separates the top- and bottom-ranked models, confirming that expert-to-lay medical image description remains a substantial open challenge.}
    \label{fig:leaderboard}
\end{figure*}

\vspace{-4pt}
\section{Experiments}
\label{sec:experiments}
\vspace{-4pt}

\subsection{Experimental Setup}
\label{sec:setup}

\paragraph{Dataset Statistics.}
\oursdata{} is constructed from 12 publicly available source datasets spanning 8 imaging modalities, with ROI annotations and expert captions supplemented from MedTrinity-25M~\cite{xie2024medtrinity} where insufficient (\Cref{tab:source_composition}).
From an initial pool of approximately 143K candidates, the \hover{} pipeline produces 122,789 verified expert-lay pairs (93.2\% yield) with an average verification score of 9.32/10.
Of these, 48.1\% pass verification directly and 51.9\% are accepted after automatic revision.
The final corpus spans 8 imaging modalities (Pathology 42.2\%, MRI 28.4\%, CT 27.9\%, and 5 others) and is split into train, validation, and test (65/15/20) with modality-stratified sampling (\Cref{tab:dataset_umls}).
Intrinsic quality metrics confirm high semantic preservation (BERTScore-F\,=\,0.89) with a 2.3-grade Flesch--Kincaid reduction (\Cref{app:intrinsic}).

\paragraph{Implementation Details.}
For the \hover{} pipeline, Step~1 uses scispaCy~\cite{neumann2019scispacy} with the UMLS entity linker for NER and concept resolution, Step~2 employs Llama-3.1-70B-Instruct~\cite{grattafiori2024llama} for constrained rewriting (temperature 0.3), and Step~3 uses Qwen3.5-27B~\cite{bai2025qwen3} for cross-model visual verification (temperature 0.1).
For \ourseval{}, Qwen2.5-VL-3B-Instruct is adapted with LoRA ($r$=16, $\alpha$=32) and a two-layer MLP regression head, trained for 3 epochs with AdamW (lr $1 \times 10^{-5}$) on 4$\times$B200 GPUs (\Cref{app:repro}).

\paragraph{Task Definition.}
We evaluate VLMs on expert-to-lay medical image description (Task~A): given a medical image with ROI bounding box, the model generates both an expert and a lay description following a standardized three-segment structure.
All 33 models are evaluated on a shared modality-stratified 5,000-sample subset of the test split, ensuring fair comparison across proprietary and open-source categories.
We additionally define two ablation tasks in \Cref{app:task_bc}: Task~B provides the ground-truth expert caption alongside the image, isolating register transfer capability, and Task~C provides only the expert caption without the image, quantifying the contribution of visual grounding.

\paragraph{Evaluation Protocol.}
We adopt reference-based metrics including n-gram overlap (BLEU~\cite{papineni2002bleu}, ROUGE-L~\cite{lin2004rouge}), semantic similarity (BERTScore-F~\cite{zhang2019bertscore}), medical entity preservation (RaTEScore~\cite{zhao2024ratescore}), LLM-based clinical error detection (GREEN~\cite{ostmeier2024green}), text simplification quality (SARI~\cite{xu2016optimizing}), and readability (Flesch--Kincaid grade~\cite{flesch1948new}) alongside \ourseval{} as the primary reference-free ranking metric.
We additionally conduct a human evaluation with two medical professionals (board-certified radiologists, 3 and 7 years experience) and two non-medical undergraduates on 100 modality-stratified samples (\Cref{tab:human_eval}).

\paragraph{Evaluated Models.}
We benchmark 33 VLM configurations spanning proprietary (6), open-source generalist (21, 0.8B--109B), and medical (6, 4B--32B) categories (\Cref{tab:leaderboard_full}).
All models use identical prompts, greedy decoding, and max 1,024 tokens; full details are in \Cref{app:repro}.
\ourseval{} belongs to a different model generation than any participant, mitigating evaluation circularity.

\begin{table*}[t]
\centering
\caption{\textbf{Benchmark results on \oursdata{} (Task~A, lay register, $n$\,=\,5,000).}
All 33 models evaluated on the same modality-stratified subset with identical \ourseval{} evaluator and prompts.
Proprietary and medical models ranked by $S$; open-source models grouped by model family (newest first).
Best in \textbf{bold}, second \underline{underlined} overall.
$\dagger$ = thinking enabled. BL\,=\,BLEU-4; R-L\,=\,ROUGE-L; GRN\,=\,GREEN; BS-F\,=\,BERTScore-F. FK\,=\,Flesch--Kincaid grade.}
\label{tab:leaderboard_full}
\resizebox{\textwidth}{!}{%
\begin{tabular}{ll c ccccc c ccccccc}
\toprule
 & & & \multicolumn{5}{c}{\textbf{\ourseval{} Attributes}} & & \multicolumn{7}{c}{\textbf{Reference-Based}} \\
\cmidrule(lr){4-8} \cmidrule(lr){10-16}
\textbf{Model} & \textbf{Params} & \textbf{$S$} & Mod & Anat & Find & Fact & Read & & BL & R-L & GRN & RaTE & BS-F & FK$\downarrow$ & SARI \\
\midrule
\multicolumn{16}{l}{\textit{Proprietary}} \\
Gemini 2.5 Flash$^\dagger$~\cite{google2025gemini25}  & undiscl. & \textbf{70.6} & \underline{67.6} & \textbf{75.4} & \textbf{78.5} & \textbf{62.0} & \textbf{69.6} & & 1.8 & 17.1 & 18.5 & 49.8 & 58.9 & 9.8 & 28.0 \\
Claude Opus 4.7~\cite{anthropic2026opus47}            & undiscl. & 64.1 & 60.3 & 67.3 & 71.6 & 55.4 & 65.9 & & 2.0 & 17.4 & 13.6 & 49.4 & 58.2 & 12.0 & 28.8 \\
GPT-5.4-mini~\cite{openai2026gpt54}                   & undiscl. & 61.2 & 59.0 & 64.4 & 68.1 & 53.4 & 61.3 & & 3.6 & 22.3 & 16.3 & 49.9 & 62.1 & 9.9 & 27.8 \\
GPT-5.5~\cite{openai2026gpt55}                       & undiscl. & 60.6 & 59.4 & 64.4 & 67.8 & 53.7 & 57.7 & & 3.6 & 21.9 & 14.8 & 49.8 & 61.9 & 10.6 & 27.7 \\
Gemini 3 Flash$^\dagger$~\cite{google2026gemini3}     & undiscl. & 59.1 & 56.6 & 63.0 & 66.6 & 50.7 & 58.5 & & 2.3 & 19.9 & 15.0 & 49.7 & 59.7 & 10.9 & 27.8 \\
Claude Haiku 4.5~\cite{anthropic2025haiku45}          & undiscl. & 50.0 & 45.2 & 52.1 & 55.9 & 41.5 & 55.1 & & 1.8 & 18.1 & 8.2 & 44.5 & 58.9 & 11.8 & 25.4 \\
\midrule
\multicolumn{16}{l}{\textit{Open-Source Generalist VLMs}} \\
Qwen3.6-35B-A3B~\cite{qwen36_35b_a3b}                    & 35B  & 61.5 & 57.7 & 65.6 & 69.4 & 50.7 & 64.2 & & 2.6 & 20.3 & \textbf{28.7} & 50.1 & 60.4 & 9.0 & 28.6 \\
Qwen3.6-35B-A3B$^\dagger$~\cite{qwen36_35b_a3b}          & 35B  & 47.7 & 48.4 & 54.9 & 56.2 & 43.4 & 35.4 & & 1.4 & 12.9 & 13.7 & 52.6 & 53.4 & \underline{6.5} & 29.2 \\
\addlinespace[4pt]
Qwen3.5-27B$^\dagger$~\cite{qwen3.5}              & 27B  & 47.2 & 45.3 & 51.1 & 53.8 & 40.7 & 45.2 & & 1.8 & 16.0 & 16.4 & 53.6 & 54.0 & \underline{6.5} & 27.3 \\
Qwen3.5-9B$^\dagger$~\cite{qwen3.5}               & 9B   & 54.5 & 54.7 & 61.2 & 63.1 & 50.1 & 43.3 & & 1.8 & 14.4 & 21.9 & \textbf{55.9} & 54.2 & 6.8 & 29.3 \\
Qwen3.5-4B$^\dagger$~\cite{qwen3.5}               & 4B   & 45.0 & 44.1 & 50.4 & 52.3 & 39.6 & 38.4 & & 1.5 & 14.2 & 17.6 & 54.0 & 53.9 & \textbf{6.4} & 28.6 \\
Qwen3.5-2B$^\dagger$~\cite{qwen3.5}               & 2B   & 63.7 & \textbf{68.1} & \underline{70.8} & \underline{73.5} & \underline{61.7} & 44.3 & & 4.6 & 23.4 & \underline{26.3} & \underline{55.0} & 64.0 & 11.6 & 29.3 \\
Qwen3.5-0.8B$^\dagger$~\cite{qwen3.5}             & 0.8B & 58.2 & 64.9 & 66.5 & 68.3 & 58.9 & 32.5 & & 5.3 & \underline{24.3} & 24.8 & 54.9 & 62.7 & 10.1 & 29.3 \\
\addlinespace[4pt]
Qwen3-VL-32B$^\dagger$~\cite{bai2025qwen3}             & 32B  & 65.4 & 61.0 & 69.2 & 72.4 & 55.3 & \underline{69.2} & & 2.4 & 18.4 & 15.8 & 49.9 & 59.8 & 9.3 & 29.3 \\
Qwen3-VL-30B-A3B$^\dagger$~\cite{bai2025qwen3}         & 30B  & 64.8 & 63.2 & 68.8 & 72.4 & 56.8 & 62.7 & & 3.6 & 21.7 & 20.6 & 51.4 & 62.7 & 9.8 & 28.6 \\
Qwen3-VL-8B$^\dagger$~\cite{bai2025qwen3}              & 8B   & 64.6 & 62.0 & 68.4 & 71.7 & 55.9 & 64.9 & & 2.9 & 20.4 & 17.2 & 49.9 & 61.8 & 9.9 & 27.5 \\
Qwen3-VL-4B$^\dagger$~\cite{bai2025qwen3}              & 4B   & 63.3 & 60.6 & 67.4 & 70.9 & 54.7 & 63.1 & & 2.8 & 20.2 & 19.2 & 51.1 & 62.0 & 10.7 & 27.7 \\
Qwen3-VL-2B$^\dagger$~\cite{bai2025qwen3}              & 2B   & 59.1 & 57.0 & 62.5 & 67.0 & 49.5 & 59.6 & & 5.0 & 24.2 & 17.0 & 50.5 & 63.1 & 8.3 & \underline{30.1} \\
\addlinespace[4pt]
InternVL3.5-38B~\cite{zhu2025internvl3}                 & 38B  & 57.9 & 59.0 & 62.3 & 65.9 & 52.2 & 50.4 & & 4.5 & 23.5 & 24.7 & 52.2 & 64.0 & 10.3 & 28.3 \\
InternVL3.5-14B~\cite{zhu2025internvl3}                 & 14B  & 53.8 & 54.3 & 57.9 & 61.8 & 48.2 & 46.8 & & 3.5 & 21.7 & 23.4 & 53.0 & 62.1 & 11.4 & 28.7 \\
InternVL3.5-8B~\cite{zhu2025internvl3}                  & 8B   & 58.7 & 59.4 & 62.6 & 66.5 & 52.7 & 52.5 & & 5.3 & 23.7 & 23.1 & 52.3 & 64.0 & 9.7 & 28.9 \\
InternVL3.5-4B~\cite{zhu2025internvl3}                  & 4B   & 54.4 & 54.1 & 57.6 & 62.3 & 47.6 & 50.4 & & 4.5 & 23.3 & 20.6 & 52.7 & 63.7 & 8.6 & 28.2 \\
InternVL3.5-2B~\cite{zhu2025internvl3}                  & 2B   & 43.4 & 43.8 & 45.8 & 50.7 & 38.5 & 38.3 & & \textbf{6.4} & \textbf{26.1} & 19.0 & 53.7 & \textbf{64.1} & 7.8 & 29.2 \\
\addlinespace[4pt]
InternVL3-8B~\cite{zhu2025internvl3}                    & 8B   & 64.7 & 66.7 & 70.5 & 73.4 & 60.7 & 52.1 & & 5.0 & 23.4 & 22.7 & 52.8 & 63.9 & 10.0 & \textbf{30.1} \\
Gemma-4-31B~\cite{gemmateam2025gemma3technicalreport}   & 31B  & 48.4 & 45.9 & 51.1 & 54.9 & 41.0 & 49.3 & & 2.6 & 20.4 & 12.8 & 48.9 & 59.6 & 10.4 & 27.5 \\
Llama-4-Scout~\cite{meta2025llama4}                     & 109B & 51.1 & 52.2 & 55.6 & 58.6 & 47.5 & 41.5 & & 3.0 & 21.0 & 13.9 & 48.2 & 60.6 & 12.8 & 28.8 \\
Qwen2.5-VL-7B~\cite{wang2024qwen2}                     & 7B   & 52.1 & 53.5 & 56.2 & 60.2 & 47.8 & 43.0 & & \underline{5.4} & 24.1 & 16.4 & 50.7 & 63.3 & 9.3 & 28.9 \\
\midrule
\multicolumn{16}{l}{\textit{Medical VLMs}} \\
MedGemma-27B~\cite{sellergren2025medgemma}              & 27B  & \underline{67.2} & 66.5 & \underline{71.8} & \underline{75.2} & 59.8 & 62.7 & & 3.2 & 21.6 & 19.6 & 49.5 & 62.0 & 9.8 & 28.6 \\
Lingshu-32B~\cite{xu2025lingshu}                        & 32B  & 53.4 & 56.6 & 58.6 & 62.4 & 50.4 & 39.0 & & 4.5 & 23.0 & 15.3 & 50.6 & 63.1 & 11.0 & 28.3 \\
MedGemma-4B~\cite{sellergren2025medgemma}               & 4B   & 47.8 & 45.4 & 49.7 & 53.5 & 39.1 & 51.1 & & 3.8 & 23.0 & 15.3 & 50.0 & 62.0 & 8.4 & 27.4 \\
Lingshu-7B~\cite{xu2025lingshu}                         & 7B   & 47.3 & 50.6 & 51.6 & 55.8 & 44.7 & 33.8 & & \textbf{5.5} & \textbf{25.0} & 17.4 & 52.8 & \underline{64.1} & 9.7 & 28.7 \\
Hulu-Med-4B~\cite{jiang2025hulu}                        & 4B   & 46.4 & 46.1 & 49.0 & 52.8 & 40.3 & 43.9 & & 2.7 & 20.5 & 15.3 & 49.6 & 61.4 & 8.6 & 25.3 \\
LLaVA-Med-7B~\cite{li2023llava}                         & 7B   & 23.9 & 24.8 & 24.6 & 26.8 & 24.5 & 18.9 & & 2.6 & 17.0 & 12.4 & 49.9 & 60.0 & 16.3 & 24.6 \\
\bottomrule
\end{tabular}%
}
\end{table*}

\vspace{-10pt}
\subsection{Main Results}
\label{sec:results}
\vspace{-4pt}

\paragraph{VLM Leaderboard.}
\Cref{tab:leaderboard_full} presents the benchmark results across 33 VLMs on a modality-stratified subset of the \oursdata{} test set ($n$\,=\,5,000; \Cref{app:task_bc}), ranked by \ourseval{} overall score $S$.
Gemini~2.5~Flash achieves the highest score (70.6), followed by MedGemma-27B (67.2) and Qwen3-VL-32B (65.4), while the lowest-scoring model, LLaVA-Med (23.9), yields a 46.7-point spread that confirms expert-to-lay medical image description remains a substantial open challenge.
Even within proprietary systems, scores range from 70.6 (Gemini~2.5~Flash) to 50.0 (Claude~Haiku~4.5), revealing considerable variance among frontier models.
The top three models span three distinct training paradigms, a proprietary generalist, a medical-specialized VLM, and an open-source generalist, indicating that no single paradigm dominates the expert-to-lay task.

\vspace{-6pt}
\paragraph{Expert-Lay Gap.}
A consistent trade-off between clinical fidelity and readability emerges across model families, with Finding completeness consistently the highest-scoring attribute and Factual correctness the lowest among clinical dimensions, indicating that models more easily preserve findings than avoid introducing inaccuracies during register transfer.
Among open-source generalists, Qwen3.5-2B achieves the highest modality score (68.1) and factual correctness (61.7) but the lowest readability (44.3) in the top tier, while Qwen3-VL-32B achieves the second-highest readability (69.2) with lower factual correctness (55.3).
MedGemma-27B uniquely balances both dimensions (Fact\,=\,59.8, Read\,=\,62.7), explaining its strong second-place ranking.
In contrast, most medical VLMs exhibit pronounced readability deficits: Lingshu-32B scores 39.0 and Lingshu-7B scores 33.8 on readability despite competitive factual scores (50.4 and 44.7), confirming that domain specialization biases models toward expert-register output at the expense of patient accessibility.

\paragraph{Metric Disagreement.}
A key finding is the systematic divergence between reference-based metrics and \ourseval{}.
GREEN, trained to detect errors in expert-level reports, penalizes successful register transfer: Qwen3.6-35B-A3B leads GREEN (28.7) despite mid-tier \ourseval{} (61.5), while the overall leader Gemini~2.5~Flash scores only 18.5.
BLEU-4 and ROUGE-L are inherently depressed by simplification, yet InternVL3.5-2B achieves the highest scores (6.4 / 26.1) by copying expert terminology verbatim, ranking last among generalists by \ourseval{} (43.4).
RaTEScore compresses 27 of 33 models into a narrow 48--55 band, providing minimal discriminative power across registers.
These patterns confirm that no existing reference-based metric captures both clinical fidelity and patient accessibility, motivating \ourseval{} as a dedicated evaluator (\Cref{app:metric_disagreement}).

\vspace{-4pt}

\paragraph{Scaling and Reasoning.}
Within the Qwen3-VL family, scaling yields diminishing returns, rising from 59.1 at 2B to 63.3 at 4B, 64.6 at 8B, and 65.4 at 32B, with most gains concentrated at smaller scales.
The relationship is non-monotonic across families: InternVL3-8B (64.7) outperforms InternVL3.5-38B (57.9), and Qwen3.5-2B (63.7) surpasses Qwen3.5-27B (47.2), suggesting that model architecture and instruction-tuning quality matter more than scale alone.
Thinking mode proves counterproductive, as Qwen3.6-35B-A3B with thinking enabled (47.7) scores 13.8 points below its non-thinking variant (61.5), with chain-of-thought deliberation consuming the 1,024-token budget before producing the patient-facing answer (\Cref{app:thinking-trace}).

\vspace{-4pt}

\paragraph{Medical vs General VLMs.}
The relationship between medical specialization and lay captioning performance is nuanced.
MedGemma-27B (67.2) ranks second overall, outperforming all open-source generalists, demonstrating that medical training can benefit lay generation when combined with strong general instruction-following capabilities inherited from its Gemma backbone.
However, this is the exception: the remaining five medical VLMs rank in the lower half of the leaderboard, with MedGemma-4B (47.8), Lingshu-7B (47.3), Hulu-Med-4B (46.4), and LLaVA-Med (23.9) all scoring below the generalist median.
This asymmetry suggests that medical domain adaptation alone is insufficient, and the quality of the base model's instruction-following and language generation capabilities is the primary determinant of lay captioning performance.
Side-by-side model predictions are compared in \Cref{app:model_comparison}.

\begin{wraptable}{r}{0.48\textwidth}
\vspace{-12pt}
\centering
\caption{Human evaluation (5-pt Likert). $\kappa$: inter-annotator agreement (Cohen's kappa).}
\label{tab:human_eval}
\footnotesize
\setlength{\tabcolsep}{3pt}
\begin{tabular}{lcc|lcc}
\toprule
\multicolumn{3}{c|}{\textit{Expert (2 Medical Doctors)}} & \multicolumn{3}{c}{\textit{Layperson (2 Students)}} \\
Attribute & Score & $\kappa$ & Attribute & Score & $\kappa$ \\
\midrule
Factual Corr.  & 4.02 & 0.50 & Understand.   & 4.73 & 0.84 \\
Completeness   & 4.71 & 0.81 & Helpfulness   & 4.54 & 0.76 \\
Simplicity     & 4.29 & 0.71 & Trustworth.   & 4.39 & 0.76 \\
Fluency        & 4.83 & 0.90 & Fluency       & 4.83 & 0.90 \\
\bottomrule
\end{tabular}
\vspace{-10pt}
\end{wraptable}

\vspace{-4pt}

\paragraph{Human Evaluation.}
Radiologist evaluation confirms that lay captions largely preserve clinical content, with completeness (4.71, $\kappa$\,=\,0.81) and fluency (4.83, $\kappa$\,=\,0.90) rated highly, while simplicity scores (4.29, $\kappa$\,=\,0.71) confirm successful register transfer.
Factual correctness receives a moderate score (4.02, $\kappa$\,=\,0.50), reflecting the challenge of assessing clinical precision in simplified descriptions across diverse imaging modalities.
From a patient perspective, laypersons rate understandability (4.73) and fluency (4.83) highly ($\kappa$\,=\,0.76--0.90), though trustworthiness scores slightly lower (4.39), suggesting residual caution toward AI-generated medical explanations.
However, the gap between moderate expert-rated factual correctness and high layperson-rated understandability highlights a comprehension--accuracy disconnect: patients may perceive simplified text as fully understood even when clinical precision is imperfect, underscoring the need for expert verification in patient-facing deployment.

\vspace{-4pt}
\section{Conclusion}
\vspace{-4pt}

We introduced \oursfull{}, the first large-scale multimodal benchmark and evaluation framework for Medical Lay Language Generation.
\oursdata{} provides 122,789 region-grounded samples across 8 imaging modalities with paired expert-lay captions anchored in a three-level UMLS ontology hierarchy, constructed via Hierarchical Ontology-Verified Refinement (\hover{}).
We further proposed \ourseval{}, a lightweight 3B multimodal evaluator that jointly scores clinical fidelity and patient accessibility across five attributes without requiring reference captions.
Benchmarking 33 VLMs revealed a systematic Expert-Lay Gap, where medical specialization alone does not guarantee patient-accessible output, and the trade-off between clinical fidelity and readability persists across all model families and scales.
These findings confirm that patient accessibility must be treated as a first-class training objective alongside clinical precision, despite current limitations in English-only coverage and modality imbalance inherited from upstream sources.

\bibliographystyle{plainnat}
\bibliography{ref}

@article{goldsack2022biolaysumm,
  title={Making science simple: Corpora for the lay summarisation of scientific literature},
  author={Goldsack, Tomas and Zhang, Zhihao and Lin, Chenghua and Scarton, Carolina},
  journal={arXiv preprint arXiv:2210.09932},
  year={2022}
}

@article{shi2025universal,
  title={Universal lesion segmentation challenge 2023: a comparative research of different algorithms},
  author={Shi, Kaiwen and Li, Yifei and Ho, Binh and Wang, Jovian and Guo, Kobe},
  journal={arXiv preprint arXiv:2502.10608},
  year={2025}
}

@article{cooper2023machine,
  title={Machine learning in computational histopathology: Challenges and opportunities},
  author={Cooper, Michael and Ji, Zongliang and Krishnan, Rahul G},
  journal={Genes, Chromosomes and Cancer},
  volume={62},
  number={9},
  pages={540--556},
  year={2023},
  publisher={Wiley Online Library}
}

@article{johnson2019mimic,
  title={MIMIC-CXR, a de-identified publicly available database of chest radiographs with free-text reports},
  author={Johnson, Alistair EW and Pollard, Tom J and Berkowitz, Seth J and Greenbaum, Nathaniel R and Lungren, Matthew P and Deng, Chih-ying and Mark, Roger G and Horng, Steven},
  journal={Scientific data},
  volume={6},
  number={1},
  pages={317},
  year={2019},
  publisher={Nature Publishing Group UK London}
}

@article{johnson2016mimic,
  title={MIMIC-III, a freely accessible critical care database},
  author={Johnson, Alistair EW and Pollard, Tom J and Shen, Lu and Lehman, Li-wei H and Feng, Mengling and Ghassemi, Mohammad and Moody, Benjamin and Szolovits, Peter and Anthony Celi, Leo and Mark, Roger G},
  journal={Scientific data},
  volume={3},
  number={1},
  pages={1--9},
  year={2016},
  publisher={Nature Publishing Group}
}

@misc{bai2025qwen25vltechnicalreport,
      title={Qwen2.5-VL Technical Report}, 
      author={Shuai Bai and Keqin Chen and Xuejing Liu and Jialin Wang and Wenbin Ge and Sibo Song and Kai Dang and Peng Wang and Shijie Wang and Jun Tang and Humen Zhong and Yuanzhi Zhu and Mingkun Yang and Zhaohai Li and Jianqiang Wan and Pengfei Wang and Wei Ding and Zheren Fu and Yiheng Xu and Jiabo Ye and Xi Zhang and Tianbao Xie and Zesen Cheng and Hang Zhang and Zhibo Yang and Haiyang Xu and Junyang Lin},
      year={2025},
      eprint={2502.13923},
      archivePrefix={arXiv},
      primaryClass={cs.CV},
      url={https://arxiv.org/abs/2502.13923}, 
}

@inproceedings{moor2023med,
  title={Med-flamingo: a multimodal medical few-shot learner},
  author={Moor, Michael and Huang, Qian and Wu, Shirley and Yasunaga, Michihiro and Dalmia, Yash and Leskovec, Jure and Zakka, Cyril and Reis, Eduardo Pontes and Rajpurkar, Pranav},
  booktitle={Machine learning for health (ML4H)},
  pages={353--367},
  year={2023},
  organization={PMLR}
}

@article{yan2018deeplesion,
  title={DeepLesion: automated mining of large-scale lesion annotations and universal lesion detection with deep learning},
  author={Yan, Ke and Wang, Xiaosong and Lu, Le and Summers, Ronald M},
  journal={Journal of medical imaging},
  volume={5},
  number={3},
  pages={036501--036501},
  year={2018},
  publisher={Society of Photo-Optical Instrumentation Engineers}
}

@article{cheng2023sam,
  title={Sam-med2d},
  author={Cheng, Junlong and Ye, Jin and Deng, Zhongying and Chen, Jianpin and Li, Tianbin and Wang, Haoyu and Su, Yanzhou and Huang, Ziyan and Chen, Jilong and Jiang, Lei and others},
  journal={arXiv preprint arXiv:2308.16184},
  year={2023}
}

@article{ma2024automatic,
  title={Automatic organ and pan-cancer segmentation in abdomen ct: the flare 2023 challenge},
  author={Ma, Jun and Zhang, Yao and Gu, Song and Ge, Cheng and Wang, Ershuai and Zhou, Qin and Huang, Ziyan and Lyu, Pengju and He, Jian and Wang, Bo},
  journal={arXiv preprint arXiv:2408.12534},
  year={2024}
}

@article{menze2014multimodal,
  title={The multimodal brain tumor image segmentation benchmark (BRATS)},
  author={Menze, Bjoern H and Jakab, Andras and Bauer, Stefan and Kalpathy-Cramer, Jayashree and Farahani, Keyvan and Kirby, Justin and Burren, Yuliya and Porz, Nicole and Slotboom, Johannes and Wiest, Roland and others},
  journal={IEEE transactions on medical imaging},
  volume={34},
  number={10},
  pages={1993--2024},
  year={2014},
  publisher={IEEE}
}

@article{liao2025magical,
  title={Magical: Medical lay language generation via semantic invariance and layperson-tailored adaptation},
  author={Liao, Weibin and Wang, Tianlong and Zhu, Yinghao and Wang, Yasha and Gao, Junyi and Ma, Liantao},
  journal={arXiv preprint arXiv:2508.08730},
  year={2025}
}

@inproceedings{xiao2025overview,
  title={Overview of the biolaysumm 2025 shared task on lay summarization of biomedical research articles and radiology reports},
  author={Xiao, Chenghao and Zhao, Kun and Wang, Xiao and Wu, Siwei and Yan, Sixing and Goldsack, Tomas and Ananiadou, Sophia and Al Moubayed, Noura and Zhan, Liang and Cheung, William K and others},
  booktitle={Proceedings of the 24th Workshop on Biomedical Language Processing},
  pages={365--377},
  year={2025}
}

@article{goldsack2024overview,
  title={Overview of the biolaysumm 2024 shared task on the lay summarization of biomedical research articles},
  author={Goldsack, Tomas and Scarton, Carolina and Shardlow, Matthew and Lin, Chenghua},
  journal={arXiv preprint arXiv:2408.08566},
  year={2024}
}

@inproceedings{yao2024readme,
  title={Readme: Bridging medical jargon and lay understanding for patient education through data-centric nlp},
  author={Yao, Zonghai and Kantu, Nandyala Siddharth and Wei, Guanghao and Tran, Hieu and Duan, Zhangqi and Kwon, Sunjae and Yang, Zhichao and Yu, Hong},
  booktitle={Findings of the Association for Computational Linguistics: EMNLP 2024},
  pages={12609--12629},
  year={2024}
}

@inproceedings{shardlow2019neural,
  title={Neural text simplification of clinical letters with a domain specific phrase table},
  author={Shardlow, Matthew and Nawaz, Raheel},
  booktitle={Proceedings of the 57th annual meeting of the association for computational linguistics},
  pages={380--389},
  year={2019}
}

@article{zhao2024x,
  title={X-ray Made Simple: Lay Radiology Report Generation and Robust Evaluation},
  author={Zhao, Kun and Xiao, Chenghao and Yan, Sixing and Tang, Haoteng and Cheung, William K and Moubayed, Noura Al and Zhan, Liang and Lin, Chenghua},
  journal={arXiv preprint arXiv:2406.17911},
  year={2024}
}

@article{ruckert2024rocov2,
  title={Rocov2: Radiology objects in context version 2, an updated multimodal image dataset},
  author={R{\"u}ckert, Johannes and Bloch, Louise and Br{\"u}ngel, Raphael and Idrissi-Yaghir, Ahmad and Sch{\"a}fer, Henning and Schmidt, Cynthia S and Koitka, Sven and Pelka, Obioma and Abacha, Asma Ben and G. Seco de Herrera, Alba and others},
  journal={Scientific Data},
  volume={11},
  number={1},
  pages={688},
  year={2024},
  publisher={Nature Publishing Group UK London}
}

@inproceedings{lozano2025biomedica,
  title={Biomedica: An open biomedical image-caption archive, dataset, and vision-language models derived from scientific literature},
  author={Lozano, Alejandro and Sun, Min Woo and Burgess, James and Chen, Liangyu and Nirschl, Jeffrey J and Gu, Jeffrey and Lopez, Ivan and Aklilu, Josiah and Rau, Anita and Katzer, Austin Wolfgang and others},
  booktitle={Proceedings of the Computer Vision and Pattern Recognition Conference},
  pages={19724--19735},
  year={2025}
}

@article{sellergren2025medgemma,
  title={Medgemma technical report},
  author={Sellergren, Andrew and Kazemzadeh, Sahar and Jaroensri, Tiam and Kiraly, Atilla and Traverse, Madeleine and Kohlberger, Timo and Xu, Shawn and Jamil, Fayaz and Hughes, C{\'\i}an and Lau, Charles and others},
  journal={arXiv preprint arXiv:2507.05201},
  year={2025}
}

@article{moor2023foundation,
  title={Foundation models for generalist medical artificial intelligence},
  author={Moor, Michael and Banerjee, Oishi and Abad, Zahra Shakeri Hossein and Krumholz, Harlan M and Leskovec, Jure and Topol, Eric J and Rajpurkar, Pranav},
  journal={Nature},
  volume={616},
  number={7956},
  pages={259--265},
  year={2023},
  publisher={Nature Publishing Group UK London}
}

@inproceedings{zhu2025can,
  title={Can we trust AI doctors? a survey of medical hallucination in large language and large vision-language models},
  author={Zhu, Zhihong and Zhang, Yunyan and Zhuang, Xianwei and Zhang, Fan and Wan, Zhongwei and Chen, Yuyan and QingqingLong, QingqingLong and Zheng, Yefeng and Wu, Xian},
  booktitle={Findings of the Association for Computational Linguistics: ACL 2025},
  pages={6748--6769},
  year={2025}
}

@inproceedings{goldsack2022making,
  title={Making science simple: Corpora for the lay summarisation of scientific literature},
  author={Goldsack, Tomas and Zhang, Zhihao and Lin, Chenghua and Scarton, Carolina},
  booktitle={Proceedings of the 2022 Conference on Empirical Methods in Natural Language Processing},
  pages={10589--10604},
  year={2022}
}

@article{bodenreider2004unified,
  title={The unified medical language system (UMLS): integrating biomedical terminology},
  author={Bodenreider, Olivier},
  journal={Nucleic acids research},
  volume={32},
  number={suppl\_1},
  pages={D267--D270},
  year={2004},
  publisher={Oxford University Press}
}

@inproceedings{neumann2019scispacy,
  title={ScispaCy: fast and robust models for biomedical natural language processing},
  author={Neumann, Mark and King, Daniel and Beltagy, Iz and Ammar, Waleed},
  booktitle={Proceedings of the 18th BioNLP workshop and shared task},
  pages={319--327},
  year={2019}
}

@inproceedings{papineni2002bleu,
  title={Bleu: a method for automatic evaluation of machine translation},
  author={Papineni, Kishore and Roukos, Salim and Ward, Todd and Zhu, Wei-Jing},
  booktitle={Proceedings of the 40th annual meeting of the Association for Computational Linguistics},
  pages={311--318},
  year={2002}
}

@inproceedings{ostmeier2024green,
  title={Green: Generative radiology report evaluation and error notation},
  author={Ostmeier, Sophie and Xu, Justin and Chen, Zhihong and Varma, Maya and Blankemeier, Louis and Bluethgen, Christian and Md, Arne Edward Michalson and Moseley, Michael and Langlotz, Curtis and Chaudhari, Akshay S and others},
  booktitle={Findings of the association for computational linguistics: EMNLP 2024},
  pages={374--390},
  year={2024}
}

@inproceedings{zhao2024ratescore,
  title={Ratescore: A metric for radiology report generation},
  author={Zhao, Weike and Wu, Chaoyi and Zhang, Xiaoman and Zhang, Ya and Wang, Yanfeng and Xie, Weidi},
  booktitle={Proceedings of the 2024 Conference on Empirical Methods in Natural Language Processing},
  pages={15004--15019},
  year={2024}
}

@inproceedings{lin2004rouge,
  title={Rouge: A package for automatic evaluation of summaries},
  author={Lin, Chin-Yew},
  booktitle={Text summarization branches out},
  pages={74--81},
  year={2004}
}

@inproceedings{banerjee2005meteor,
  title={METEOR: An automatic metric for MT evaluation with improved correlation with human judgments},
  author={Banerjee, Satanjeev and Lavie, Alon},
  booktitle={Proceedings of the acl workshop on intrinsic and extrinsic evaluation measures for machine translation and/or summarization},
  pages={65--72},
  year={2005}
}

@article{zhang2019bertscore,
  title={Bertscore: Evaluating text generation with bert},
  author={Zhang, Tianyi and Kishore, Varsha and Wu, Felix and Weinberger, Kilian Q and Artzi, Yoav},
  journal={arXiv preprint arXiv:1904.09675},
  year={2019}
}

@article{zhu2025medagentboard,
  title={Medagentboard: Benchmarking multi-agent collaboration with conventional methods for diverse medical tasks},
  author={Zhu, Yinghao and He, Ziyi and Hu, Haoran and Zheng, Xiaochen and Zhang, Xichen and Wang, Zixiang and Gao, Junyi and Ma, Liantao and Yu, Lequan},
  journal={arXiv preprint arXiv:2505.12371},
  year={2025}
}

@article{flesch1948new,
  title={A new readability yardstick.},
  author={Flesch, Rudolph},
  journal={Journal of applied psychology},
  volume={32},
  number={3},
  pages={221},
  year={1948},
  publisher={American Psychological Association}
}

@article{prucker2025prospective,
  title={A Prospective Controlled Trial of Large Language Model--based Simplification of Oncologic CT Reports for Patients with Cancer},
  author={Prucker, Philipp and Bressem, Keno K and Peeken, Jan and Jukic, Mateo and Marka, Alexander W and Strenzke, Maximilian and Kim, Su Hwan and Mertens, Christian J and Weller, Dominik and Lemke, Tristan and others},
  journal={Radiology},
  volume={317},
  number={2},
  pages={e251844},
  year={2025},
  publisher={Radiological Society of North America}
}

@inproceedings{eslami2023pubmedclip,
  title={Pubmedclip: How much does clip benefit visual question answering in the medical domain?},
  author={Eslami, Sedigheh and Meinel, Christoph and De Melo, Gerard},
  booktitle={Findings of the Association for Computational Linguistics: EACL 2023},
  pages={1181--1193},
  year={2023}
}

@article{zhang2025multimodal,
  title={A multimodal biomedical foundation model trained from fifteen million image--text pairs},
  author={Zhang, Sheng and Xu, Yanbo and Usuyama, Naoto and Xu, Hanwen and Bagga, Jaspreet and Tinn, Robert and Preston, Sam and Rao, Rajesh and Wei, Mu and Valluri, Naveen and others},
  journal={Nejm Ai},
  volume={2},
  number={1},
  pages={AIoa2400640},
  year={2025},
  publisher={Massachusetts Medical Society}
}

@article{xie2024medtrinity,
  title={Medtrinity-25m: A large-scale multimodal dataset with multigranular annotations for medicine},
  author={Xie, Yunfei and Zhou, Ce and Gao, Lang and Wu, Juncheng and Li, Xianhang and Zhou, Hong-Yu and Liu, Sheng and Xing, Lei and Zou, James and Xie, Cihang and others},
  journal={arXiv preprint arXiv:2408.02900},
  year={2024}
}

@article{zhang2023pmc,
  title={Pmc-vqa: Visual instruction tuning for medical visual question answering},
  author={Zhang, Xiaoman and Wu, Chaoyi and Zhao, Ziheng and Lin, Weixiong and Zhang, Ya and Wang, Yanfeng and Xie, Weidi},
  journal={arXiv preprint arXiv:2305.10415},
  year={2023}
}

@article{li2023llava,
  title={Llava-med: Training a large language-and-vision assistant for biomedicine in one day},
  author={Li, Chunyuan and Wong, Cliff and Zhang, Sheng and Usuyama, Naoto and Liu, Haotian and Yang, Jianwei and Naumann, Tristan and Poon, Hoifung and Gao, Jianfeng},
  journal={Advances in Neural Information Processing Systems},
  volume={36},
  pages={28541--28564},
  year={2023}
}

@article{jang2026medlaybench,
  title={MedLayBench-V: A Large-Scale Benchmark for Expert-Lay Semantic Alignment in Medical Vision Language Models},
  author={Jang, Han and Lee, Junhyeok and Eum, Heeseong and Choi, Kyu Sung},
  journal={arXiv preprint arXiv:2604.05738},
  year={2026}
}

@article{lau2018dataset,
  title={A dataset of clinically generated visual questions and answers about radiology images},
  author={Lau, Jason J and Gayen, Soumya and Ben Abacha, Asma and Demner-Fushman, Dina},
  journal={Scientific data},
  volume={5},
  number={1},
  pages={180251},
  year={2018},
  publisher={Nature Publishing Group}
}

@inproceedings{liu2021slake,
  title={Slake: A semantically-labeled knowledge-enhanced dataset for medical visual question answering},
  author={Liu, Bo and Zhan, Li-Ming and Xu, Li and Ma, Lin and Yang, Yan and Wu, Xiao-Ming},
  booktitle={2021 IEEE 18th international symposium on biomedical imaging (ISBI)},
  pages={1650--1654},
  year={2021},
  organization={IEEE}
}

@article{he2020pathvqa,
  title={Pathvqa: 30000+ questions for medical visual question answering},
  author={He, Xuehai and Zhang, Yichen and Mou, Luntian and Xing, Eric and Xie, Pengtao},
  journal={arXiv preprint arXiv:2003.10286},
  year={2020}
}

@article{hu2022lora,
  title={Lora: Low-rank adaptation of large language models.},
  author={Hu, Edward J and Shen, Yelong and Wallis, Phillip and Allen-Zhu, Zeyuan and Li, Yuanzhi and Wang, Shean and Wang, Liang and Chen, Weizhu and others},
  journal={Iclr},
  volume={1},
  number={2},
  pages={3},
  year={2022}
}

@article{hendrycks2016gaussian,
  title={Gaussian error linear units (gelus)},
  author={Hendrycks, Dan and Gimpel, Kevin},
  journal={arXiv preprint arXiv:1606.08415},
  year={2016}
}

@article{xu2016optimizing,
  title={Optimizing statistical machine translation for text simplification},
  author={Xu, Wei and Napoles, Courtney and Pavlick, Ellie and Chen, Quanze and Callison-Burch, Chris},
  journal={Transactions of the Association for Computational Linguistics},
  volume={4},
  pages={401--415},
  year={2016}
}

@article{vincoff2022patient,
  title={The patient-friendly radiology report: history, evolution, challenges and opportunities},
  author={Vincoff, Nina S and Barish, Matthew A and Grimaldi, Gregory},
  journal={Clinical Imaging},
  volume={89},
  pages={128--135},
  year={2022},
  publisher={Elsevier}
}

@article{zeng2006exploring,
  title={Exploring and developing consumer health vocabularies},
  author={Zeng, Qing T and Tse, Tony},
  journal={Journal of the American Medical Informatics Association},
  volume={13},
  number={1},
  pages={24--29},
  year={2006},
  publisher={BMJ Group BMA House, Tavistock Square, London, WC1H 9JR}
}

@inproceedings{xu2025radeval,
  title={RadEval: A framework for radiology text evaluation},
  author={Xu, Justin and Zhang, Xi and Abderezaei, Javid and Bauml, Julie and Boodoo, Roger and Haghighi, Fatemeh and Ganjizadeh, Ali and Brattain, Eric and Van Veen, Dave and Meng, Zaiqiao and others},
  booktitle={Proceedings of the 2025 Conference on Empirical Methods in Natural Language Processing: System Demonstrations},
  pages={546--557},
  year={2025}
}

@article{steitz2023perspectives,
  title={Perspectives of patients about immediate access to test results through an online patient portal},
  author={Steitz, Bryan D and Turer, Robert W and Lin, Chen-Tan and MacDonald, Scott and Salmi, Liz and Wright, Adam and Lehmann, Christoph U and Langford, Karen and McDonald, Samuel A and Reese, Thomas J and others},
  journal={JAMA Network Open},
  volume={6},
  number={3},
  pages={e233572},
  year={2023}
}

@article{bhalla2024contemporary,
  title={Contemporary trends in reviewing test results through the electronic patient portal among patients with cancer},
  author={Bhalla, Sheena and Prasad, Tanushree and Xie, Donglu and Gerber, David E},
  journal={JAMA oncology},
  volume={10},
  number={1},
  pages={139--140},
  year={2024}
}

@inproceedings{dambrino2023unsolicited,
  title={Unsolicited patient complaints following the 21st Century Cures Act information-blocking rule},
  author={Dambrino IV, Robert J and Domenico, Henry J and Graves, John A and Buntin, Melinda JB and Martinez, William and Rosenbloom, S Trent and Cooper, William O},
  booktitle={JAMA Health Forum},
  volume={4},
  pages={e233244},
  year={2023}
}

@article{wang2024qwen2,
  title={Qwen2-vl: Enhancing vision-language model's perception of the world at any resolution},
  author={Wang, Peng and Bai, Shuai and Tan, Sinan and Wang, Shijie and Fan, Zhihao and Bai, Jinze and Chen, Keqin and Liu, Xuejing and Wang, Jialin and Ge, Wenbin and others},
  journal={arXiv preprint arXiv:2409.12191},
  year={2024}
}

@misc{meta2025llama4,
  author       = {{Meta}},
  title        = {The {Llama} 4 Herd: The Beginning of a New Era 
                  of Natively Multimodal {AI} Innovation},
  year         = {2025},
  month        = apr,
  howpublished = {\url{https://ai.meta.com/blog/llama-4-multimodal-intelligence/}},
  note         = {Accessed: 2026-05-06}
}

@article{grattafiori2024llama,
  title={The llama 3 herd of models},
  author={Grattafiori, Aaron and Dubey, Abhimanyu and Jauhri, Abhinav and Pandey, Abhinav and Kadian, Abhishek and Al-Dahle, Ahmad and Letman, Aiesha and Mathur, Akhil and Schelten, Alan and Vaughan, Alex and others},
  journal={arXiv preprint arXiv:2407.21783},
  year={2024}
}

@misc{gemmateam2025gemma3technicalreport,
      title={Gemma 3 Technical Report}, 
      author={Google DeepMind},
      year={2025},
      eprint={2503.19786},
      archivePrefix={arXiv},
      primaryClass={cs.CL},
      url={https://arxiv.org/abs/2503.19786}, 
}

@article{xu2025lingshu,
  title={Lingshu: A generalist foundation model for unified multimodal medical understanding and reasoning},
  author={Xu, Weiwen and Chan, Hou Pong and Li, Long and Aljunied, Mahani and Yuan, Ruifeng and Wang, Jianyu and Xiao, Chenghao and Chen, Guizhen and Liu, Chaoqun and Li, Zhaodonghui and others},
  journal={arXiv preprint arXiv:2506.07044},
  year={2025}
}

@article{zhu2025internvl3,
  title={Internvl3: Exploring advanced training and test-time recipes for open-source multimodal models},
  author={Zhu, Jinguo and Wang, Weiyun and Chen, Zhe and Liu, Zhaoyang and Ye, Shenglong and Gu, Lixin and Tian, Hao and Duan, Yuchen and Su, Weijie and Shao, Jie and others},
  journal={arXiv preprint arXiv:2504.10479},
  year={2025}
}

@article{jiang2025hulu,
  title={Hulu-med: A transparent generalist model towards holistic medical vision-language understanding},
  author={Jiang, Songtao and Wang, Yuan and Song, Sibo and Hu, Tianxiang and Zhou, Chenyi and Pu, Bin and Zhang, Yan and Yang, Zhibo and Feng, Yang and Zhou, Joey Tianyi and others},
  journal={arXiv preprint arXiv:2510.08668},
  year={2025}
}

@article{alabed2026large,
  title={Large language models for simplifying radiology reports: a systematic review and meta-analysis of patient, public, and clinician evaluations},
  author={Alabed, Samer and Anderson, Abigail and Maiter, Ahmed and Hughes, Anthony and McAnenly, Niamh and Salehi, Mahan and Sharkey, Michael and Dwivedi, Krit and Hokmabadi, Alireza and Alahdab, Fares and others},
  journal={The Lancet Digital Health},
  year={2026},
  publisher={Elsevier}
}

@article{yao2026medical,
  title={Medical thinking with multiple images},
  author={Yao, Zonghai and Wang, Benlu and Zhang, Yifan and Wang, Junda and Xia, Iris and Tang, Zhipeng and Han, Shuo and Ouyang, Feiyun and Yang, Zhichao and Cohan, Arman and others},
  journal={arXiv preprint arXiv:2604.16506},
  year={2026}
}

@article{bai2025qwen3,
  title={Qwen3-vl technical report},
  author={Bai, Shuai and Cai, Yuxuan and Chen, Ruizhe and Chen, Keqin and Chen, Xionghui and Cheng, Zesen and Deng, Lianghao and Ding, Wei and Gao, Chang and Ge, Chunjiang and others},
  journal={arXiv preprint arXiv:2511.21631},
  year={2025}
}

@misc{qwen3.5,
    title  = {{Qwen3.5}: Towards Native Multimodal Agents},
    author = {{Qwen Team}},
    month  = {February},
    year   = {2026},
    url    = {https://qwen.ai/blog?id=qwen3.5}
}

@misc{qwen36_35b_a3b,
    title = {{Qwen3.6-35B-A3B}: Agentic Coding Power, Now Open to All},
    url = {https://qwen.ai/blog?id=qwen3.6-35b-a3b},
    author = {{Qwen Team}},
    month = {April},
    year = {2026}
}

@article{pollock2024patient,
  title={Patient access of their radiology reports before and after implementation of 21st Century Cures Act information-blocking provisions at a large multicampus health system},
  author={Pollock, Jordan R and Petty, Skye A Buckner and Schmitz, John J and Varner, Jacob and Metcalfe, Allie M and Tan, Nelly},
  journal={American Journal of Roentgenology},
  volume={222},
  number={6},
  pages={e2330343},
  year={2024},
  publisher={American Roentgen Ray Society}
}

@inproceedings{wang2023automated,
  title={Automated metrics for medical multi-document summarization disagree with human evaluations},
  author={Wang, Lucy Lu and Otmakhova, Julia and DeYoung, Jay and Truong, Thinh Hung and Kuehl, Bailey and Bransom, Erin and Wallace, Byron C},
  booktitle={Proceedings of the 61st Annual Meeting of the Association for Computational Linguistics (Volume 1: Long Papers)},
  pages={9871--9889},
  year={2023}
}

@misc{pmcoa,
  author       = {{National Library of Medicine}},
  title        = {{PMC} Open Access Subset},
  howpublished = {\url{https://pmc.ncbi.nlm.nih.gov/tools/openftlist/}},
  year         = {2003--},
  note         = {Accessed: 2026-05-06}
}

@inproceedings{wang2017chestx,
  title={Chestx-ray8: Hospital-scale chest x-ray database and benchmarks on weakly-supervised classification and localization of common thorax diseases},
  author={Wang, Xiaosong and Peng, Yifan and Lu, Le and Lu, Zhiyong and Bagheri, Mohammadhadi and Summers, Ronald M},
  booktitle={Proceedings of the IEEE conference on computer vision and pattern recognition},
  pages={2097--2106},
  year={2017}
}

@inproceedings{kwon2023efficient,
  title={Efficient memory management for large language model serving with pagedattention},
  author={Kwon, Woosuk and Li, Zhuohan and Zhuang, Siyuan and Sheng, Ying and Zheng, Lianmin and Yu, Cody Hao and Gonzalez, Joseph and Zhang, Hao and Stoica, Ion},
  booktitle={Proceedings of the 29th symposium on operating systems principles},
  pages={611--626},
  year={2023}
}

@article{zheng2024sglang,
  title={Sglang: Efficient execution of structured language model programs},
  author={Zheng, Lianmin and Yin, Liangsheng and Xie, Zhiqiang and Sun, Chuyue and Huang, Jeff and Yu, Cody H and Cao, Shiyi and Kozyrakis, Christos and Stoica, Ion and Gonzalez, Joseph E and others},
  journal={Advances in neural information processing systems},
  volume={37},
  pages={62557--62583},
  year={2024}
}

@article{bos2006snomed,
  title={SNOMED-CT: The advanced terminology and coding system for eHealth},
  author={Bos, L and Donnelly, K},
  journal={Stud Health Technol Inform},
  volume={121},
  pages={279--290},
  year={2006}
}

@article{sioutos2007nci,
  title={NCI Thesaurus: a semantic model integrating cancer-related clinical and molecular information},
  author={Sioutos, Nicholas and de Coronado, Sherri and Haber, Margaret W and Hartel, Frank W and Shaiu, Wen-Ling and Wright, Lawrence W},
  journal={Journal of biomedical informatics},
  volume={40},
  number={1},
  pages={30--43},
  year={2007},
  publisher={Elsevier}
}

@misc{mesh2024,
  title        = {Medical Subject Headings ({MeSH})},
  author       = {{National Library of Medicine}},
  year         = {2024},
  howpublished = {\url{https://www.nlm.nih.gov/mesh/}},
}

@misc{openai2026gpt55,
  title   = {{GPT-5.5} System Card},
  author  = {{OpenAI}},
  year    = {2026},
  month   = apr,
  url     = {https://openai.com/index/gpt-5-5-system-card/},
}

@misc{openai2026gpt54,
  title   = {{GPT-5.4} and {GPT-5.4-mini} Model Card},
  author  = {{OpenAI}},
  year    = {2026},
  month   = feb,
  url     = {https://platform.openai.com/docs/models},
}

@misc{anthropic2026opus47,
  title   = {Introducing {Claude Opus 4.7}},
  author  = {{Anthropic}},
  year    = {2026},
  month   = apr,
  url     = {https://www.anthropic.com/news/claude-opus-4-7},
}

@misc{anthropic2025haiku45,
  title   = {{Claude} Models Overview},
  author  = {{Anthropic}},
  year    = {2025},
  url     = {https://docs.anthropic.com/en/docs/about-claude/models},
}

@misc{google2026gemini3,
  title   = {Gemini 3 Flash Model Card},
  author  = {{Google DeepMind}},
  year    = {2026},
  url     = {https://ai.google.dev/gemini-api/docs/models#gemini-3-flash-preview},
}

@misc{google2025gemini25,
  title   = {Gemini 2.5 Flash Model Card},
  author  = {{Google DeepMind}},
  year    = {2025},
  url     = {https://ai.google.dev/gemini-api/docs/models#gemini-2.5-flash},
}

@article{ding2025multimodal,
  title={A multimodal whole-slide foundation model for pathology},
  author={Ding, Tong and Wagner, Sophia J and Song, Andrew H and Chen, Richard J and Lu, Ming Y and Zhang, Andrew and Vaidya, Anurag J and Jaume, Guillaume and Shaban, Muhammad and Kim, Ahrong and others},
  journal={Nature medicine},
  pages={1--13},
  year={2025},
  publisher={Nature Publishing Group US New York}
}

@article{panickssery2024llm,
  title={Llm evaluators recognize and favor their own generations},
  author={Panickssery, Arjun and Bowman, Samuel R and Feng, Shi},
  journal={Advances in Neural Information Processing Systems},
  volume={37},
  pages={68772--68802},
  year={2024}
}

@article{wataoka2024self,
  title={Self-preference bias in llm-as-a-judge},
  author={Wataoka, Koki and Takahashi, Tsubasa and Ri, Ryokan},
  journal={arXiv preprint arXiv:2410.21819},
  year={2024}
}

@article{verga2024replacing,
  title={Replacing judges with juries: Evaluating llm generations with a panel of diverse models},
  author={Verga, Pat and Hofstatter, Sebastian and Althammer, Sophia and Su, Yixuan and Piktus, Aleksandra and Arkhangorodsky, Arkady and Xu, Minjie and White, Naomi and Lewis, Patrick},
  journal={arXiv preprint arXiv:2404.18796},
  year={2024}
}

\clearpage
\appendix
\setcounter{figure}{0}
\setcounter{table}{0}
\makeatletter
\@addtoreset{figure}{section}
\@addtoreset{table}{section}
\makeatother
\renewcommand{\thefigure}{\thesection.\arabic{figure}}
\renewcommand{\thetable}{\Alph{section}.\arabic{table}}

\section{Dataset Statistics}
\label{app:dataset}

Table~\ref{tab:dataset_umls} summarizes the modality distribution and UMLS ontology statistics for the released 122,789 samples.
Table~\ref{tab:source_composition} lists the 12 source datasets and their contributions.

\begin{table}[h!]
\centering
\caption{Dataset overview.
(Left) Modality distribution with modality-stratified splits (65/15/20).
(Right) UMLS ontology statistics showing top-4 entries per level; 99.8\% of samples contain $\geq$1 CUI.}
\label{tab:dataset_umls}
\begin{minipage}[t]{0.52\linewidth}
\centering
\footnotesize
\begin{tabular}{lrrrr}
\toprule
Modality & Train & Val & Test & Total (\%) \\
\midrule
Pathology   & 33,615 & 7,801  & 10,346 & 51,762 (42.2) \\
MRI         & 22,630 & 5,240  & 6,996  & 34,866 (28.4) \\
CT          & 22,293 & 5,154  & 6,864  & 34,311 (27.9) \\
X-Ray       & 988    & 242    & 327    & 1,557 (1.3)   \\
PET         & 108    & 26     & 34     & 168 (0.1)     \\
Dermoscopy  & 52     & 13     & 15     & 80 ($<$0.1)   \\
Endoscopy   & 22     & 6      & 6      & 34 ($<$0.1)   \\
Ultrasound  & 7      & 2      & 2      & 11 ($<$0.1)   \\
\midrule
Total & 79,715 & 18,484 & 24,590 & 122,789 \\
\bottomrule
\end{tabular}
\end{minipage}%
\hfill
\begin{minipage}[t]{0.46\linewidth}
\centering
\footnotesize
\begin{tabular}{llr}
\toprule
Level & Entry & Count \\
\midrule
\multirow{4}{*}{Sem.\ Group}
 & Anatomy            & 265,927 \\
 & Disorders           & 229,811 \\
 & Procedures          & 127,028 \\
 & Chem.\ \& Drugs    & 20,605  \\
\midrule
\multirow{4}{*}{TUI (43)}
 & T060 Diag.\ Proc.  & 126,450 \\
 & T033 Finding        & 103,435 \\
 & T023 Body Part      & 79,642  \\
 & T024 Tissue         & 64,589  \\
\midrule
\multirow{4}{*}{CUI (2,411)}
 & C0221198 Lesion     & 48,285 \\
 & C0024485 MRI        & 43,473 \\
 & C0475358 Tumor      & 35,999 \\
 & C0030660 Path.\ Proc. & 35,154 \\
\bottomrule
\end{tabular}
\end{minipage}
\end{table}

\begin{table}[ht]
\centering
\caption{Source composition of \textsc{MedLayXPlain}-122K, curated from 12 publicly available datasets via MedTrinity-25M~\cite{xie2024medtrinity} (60/67 shards).
$N$ denotes pre-filtering counts; 122,789 samples (93.2\%) are retained after \hover{} verification.}
\label{tab:source_composition}
\footnotesize
\begin{tabular}{@{}llrr@{\hskip 12pt}llrr@{}}
\toprule
\textbf{Source} & \textbf{Mod.} & \textbf{$N$} & \textbf{\%}
& \textbf{Source} & \textbf{Mod.} & \textbf{$N$} & \textbf{\%} \\
\midrule
SAM-Med~\cite{cheng2023sam}        & MRI  & 24,880 & 18.8
& PMC-VQA~\cite{zhang2023pmc}       & Mix  & 9,869  & 7.5  \\
DeepLesion~\cite{yan2018deeplesion}  & CT   & 17,348 & 13.1
& TCGA                         & Path & 9,622  & 7.3  \\
PTCGA                          & Path & 17,223 & 13.0
& ULS23~\cite{shi2025universal}           & CT   & 5,576  & 4.2  \\
PMC-OA~\cite{pmcoa}           & Mix  & 12,165 & 9.2
& LLaVA-Med~\cite{li2023llava}   & Mix  & 4,597  & 3.5  \\
FLARE23~\cite{ma2024automatic}        & CT   & 10,000 & 7.6
& NIH CXR~\cite{wang2017chestx}     & XR   & 983    & 0.7  \\
CISC~\cite{cooper2023machine}               & Path & 10,000 & 7.6
& & & & \\
BraTS~\cite{menze2014multimodal}             & MRI  & 10,000 & 7.6
& \multicolumn{2}{@{}l}{\textbf{Total}} & \textbf{132,263} & \textbf{100} \\
\bottomrule
\end{tabular}
\end{table}

\begin{figure*}[h!]
\centering
\includegraphics[width=\textwidth]{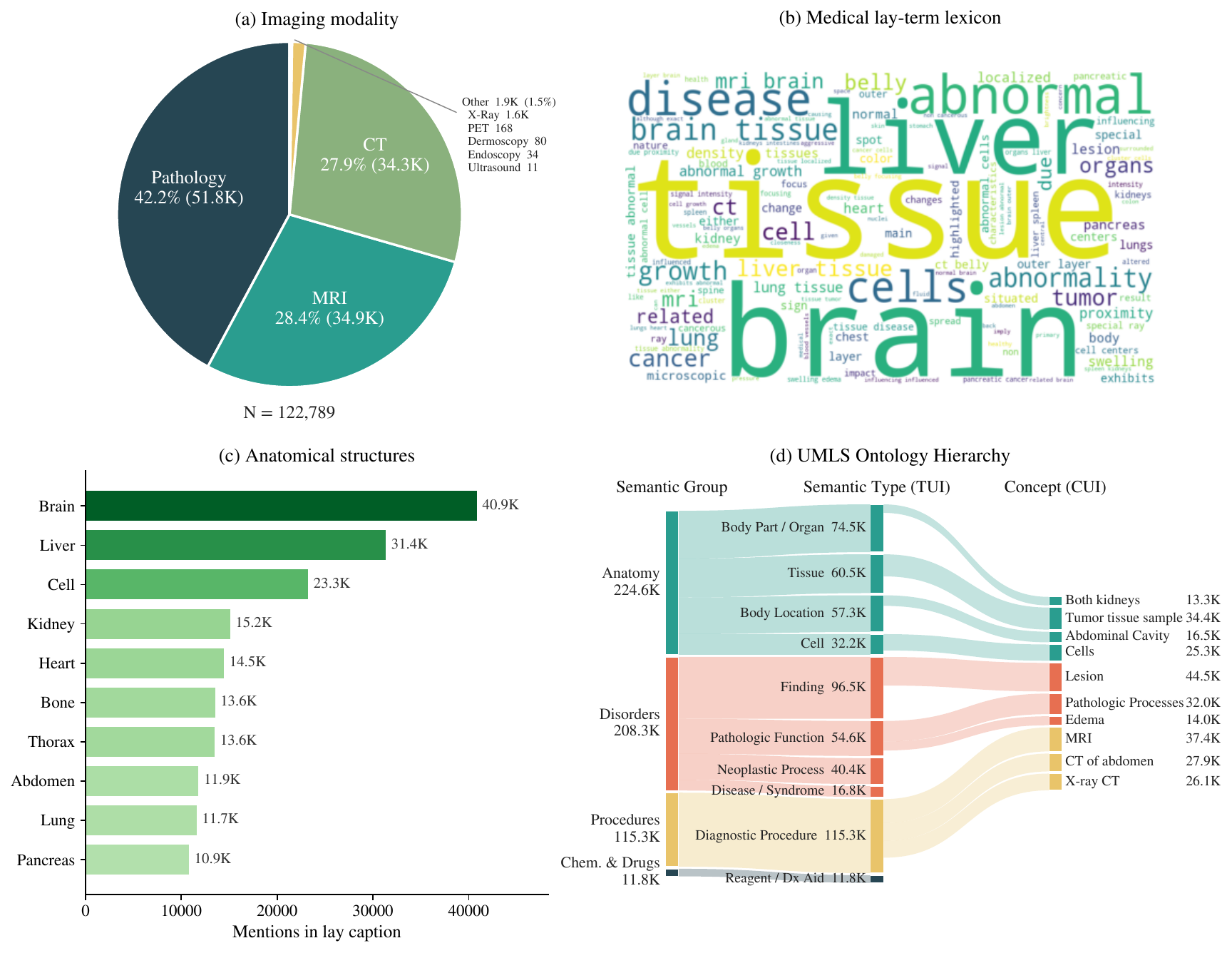}
\caption{\textbf{MedLayXPlain-122K dataset statistics.}
(a)~Distribution across 8 imaging modalities, with Pathology
(42.2\%), MRI (28.4\%), and CT (27.9\%) comprising the majority.
(b)~Word cloud of the most frequent terms in the final lay
captions, reflecting patient-accessible vocabulary produced
by the HOVER pipeline.
(c)~Top-10 anatomical structures mentioned in lay captions.
(d)~Three-level UMLS ontology hierarchy showing the top-10
of 43 semantic types and top-10 of 2,306 unique concepts
across 7 semantic groups; flow widths represent mention
counts in lay captions.}
\label{fig:statistics}
\end{figure*}

\section{Benchmark Comparison}
\label{app:benchmark}

Table~\ref{tab:benchmark_comparison} compares \textsc{MedLayXPlain}-122K against existing medical benchmarks along six dimensions: scale, modality coverage, image availability, expert captions, lay captions, and ROI grounding.
To our knowledge, \textsc{MedLayXPlain}-122K is the first resource to provide all six simultaneously.

\begin{table}[h!]
\centering
\caption{Comparison with existing medical benchmarks.
\#Mod.\ = number of imaging modalities.}
\label{tab:benchmark_comparison}
\resizebox{\linewidth}{!}{%
\begin{tabular}{lcccccc}
\toprule
Dataset & Scale & \#Mod. & Image & Expert Cap. & Lay Cap. & ROI \\
\midrule
Cochrane~\cite{goldsack2022making}         & 4K    & -- & \ding{55} & \checkmark & \checkmark & \ding{55} \\
eLife~\cite{goldsack2022making}            & 5K    & -- & \ding{55} & \checkmark & \checkmark & \ding{55} \\
Plos~\cite{yao2024readme}                  & 4K    & -- & \ding{55} & \checkmark & \checkmark & \ding{55} \\
ROCOv2~\cite{ruckert2024rocov2}            & 80K   & 7  & \checkmark & \checkmark & \ding{55} & \ding{55} \\
MIMIC-CXR~\cite{johnson2016mimic}          & 227K  & 1  & \checkmark & \checkmark & \ding{55} & \ding{55} \\
BIOMEDICA~\cite{lozano2025biomedica}       & 24M   & 6  & \checkmark & \checkmark & \ding{55} & \ding{55} \\
MedTrinity-25M~\cite{xie2024medtrinity}    & 25M   & 10 & \checkmark & \checkmark & \ding{55} & \checkmark \\
MedLayBench-V~\cite{jang2026medlaybench}  & 80K   & 7  & \checkmark & \checkmark & \checkmark & \ding{55} \\
\rowcolor{blue!4}
\textsc{MedLayXPlain}-122K                 & 122K  & 8  & \checkmark & \checkmark & \checkmark & \checkmark \\
\bottomrule
\end{tabular}%
}
\end{table}

\section{HOVER Verification Statistics}
\label{app:hover}

\begin{table}[h!]
\centering
\caption{HOVER Step~3 verification statistics on the full dataset ($N$\,=\,131,678).
Per-attribute scores are on a 0--2 scale.}
\label{tab:hover_quality}
\footnotesize
\begin{tabular}{lr}
\toprule
\textbf{Metric} & \textbf{Value} \\
\midrule
Total candidates & 131,678 \\
PASS (10/10) & 59,062 (44.9\%) \\
REVISE (7--9/10) & 63,727 (48.4\%) \\
FAIL ($<$7/10) & 8,889 (6.7\%) \\
\midrule
Released samples (PASS+REVISE) & 122,789 (93.2\%) \\
Avg.\ total score & 9.32/10 \\
\midrule
Modality consistency & 1.96/2 \\
Anatomical accuracy & 1.96/2 \\
Finding completeness & 1.99/2 \\
Factual correctness & 1.92/2 \\
Lay readability & 1.49/2 \\
\midrule
Upstream expert captions corrected & 42,996 (35.0\%) \\
\bottomrule
\end{tabular}
\end{table}

\Cref{tab:hover_quality} summarizes the verification outcomes of the \hover{} pipeline (\Cref{alg:hover}).
Of the 122,789 released samples, 48.1\% passed verification on the first attempt and the remaining 51.9\% were accepted after automatic revision guided by attribute-level feedback.
The cross-model verification also served as upstream quality assurance, flagging 35.0\% of source expert captions (42,996 rows) as containing factual errors inherited from MedTrinity-25M such as modality mislabeling or incorrect anatomical references.
For these rows the verifier generated a corrected expert caption, which the released dataset stores to maintain internal consistency across the expert/lay/image triple.
This correction rate is consistent across splits (35.2\% train, 35.4\% val, 34.0\% test), confirming that the errors originate from the upstream source rather than from split-specific artifacts.

\begin{algorithm}[h]
\caption{HOVER: Hierarchical Ontology-Verified Refinement}
\label{alg:hover}
\begin{algorithmic}[1]
\REQUIRE Dataset $\mathcal{D}$ with $(I, b, T_{\text{exp}})$ tuples; max revision rounds $R=2$
\ENSURE Verified expert-lay pairs $\mathcal{D}_{\text{out}}$

\FOR{each $(I, b, T_{\text{exp}})$ in $\mathcal{D}$}

\STATE \textcolor{gray}{\textit{// Step 1: Ontology-Grounded Mapping}}
\STATE $\mathcal{E} \leftarrow \text{NER}(T_{\text{exp}})$ \hfill $\triangleright$ scispaCy entity extraction
\FOR{each $e_i \in \mathcal{E}$}
    \STATE Resolve $e_i \rightarrow c_i \in \mathcal{C},\; \tau(c_i) \in \mathcal{T},\; g(\tau(c_i)) \in \mathcal{G}$ \hfill $\triangleright$ UMLS 3-level hierarchy
    \STATE $m(e_i) \leftarrow \text{Onto}(c_i)$ \hfill $\triangleright$ CHV lookup with fallback
\ENDFOR

\STATE \textcolor{gray}{\textit{// Step 2: Constrained Rewriting}}
\STATE $T_{\text{draft}} \leftarrow \text{LLM}\bigl(T_{\text{exp}},\; \{m(e_i), \tau(c_i), g(\tau(c_i))\}_{i=1}^{n}\bigr)$ \hfill $\triangleright$ Llama-3.1-70B

\STATE \textcolor{gray}{\textit{// Step 3: Cross-Model Visual Verification}}
\STATE $r \leftarrow 0$
\REPEAT
    \STATE $\mathbf{s} \leftarrow \text{VLM}(I, b, T_{\text{exp}}, T_{\text{draft}}) \in [0,2]^5$ \hfill $\triangleright$ Qwen3.5-27B
    \STATE $S \leftarrow \sum_{j=1}^{5} s_j$
    \IF{$S = 10$}
        \STATE verdict $\leftarrow$ \textsc{Pass}; \quad $T_{\text{lay}} \leftarrow T_{\text{draft}}$
    \ELSIF{$S \geq 7$}
        \STATE verdict $\leftarrow$ \textsc{Revise}
        \STATE $T_{\text{draft}} \leftarrow \text{VLM}_{\text{correct}}(I, b, T_{\text{exp}}, T_{\text{draft}}, \mathbf{s})$ \hfill $\triangleright$ Verifier-guided revision
    \ELSE
        \STATE verdict $\leftarrow$ \textsc{Fail}
        \STATE $T_{\text{draft}} \leftarrow \text{VLM}_{\text{retry}}\bigl(I, b, T_{\text{exp}}, T_{\text{draft}}, \mathbf{s}\bigr)$ \hfill $\triangleright$ Verifier-guided re-generation
    \ENDIF
    \STATE $r \leftarrow r + 1$
\UNTIL{verdict $=$ \textsc{Pass} \textbf{or} $r > R$}

\IF{verdict $\neq$ \textsc{Fail}}
    \STATE $\mathcal{D}_{\text{out}} \leftarrow \mathcal{D}_{\text{out}} \cup \{(I, b, T_{\text{exp}}, T_{\text{lay}})\}$
\ENDIF

\ENDFOR
\end{algorithmic}
\end{algorithm}

\section{Cross-Family Design Rationale}
\label{app:cross_family}

The \hover{} pipeline assigns generation (Step~2) and verification (Step~3) to models from disjoint families: Llama-3.1-70B-Instruct for rewriting and Qwen3.5-27B for visual verification.
This cross-family design is motivated by empirical findings that LLM judges exhibit self-preference bias, systematically favoring their own outputs via stylistic self-recognition~\cite{panickssery2024llm}, and that this bias extends to family-level preference, where judges inflate scores for architecturally related models from the same provider~\cite{wataoka2024self}.

Using disjoint families for generation and verification follows the Panel of LLM Evaluators (PoLL) principle~\cite{verga2024replacing}, which demonstrates that diverse model ensembles exhibit less intra-model bias than single-model evaluation.
\ourseval{} is further validated against independent human judgments from board-certified radiologists (\Cref{tab:human_eval}), confirming that the distilled evaluator captures clinically meaningful quality distinctions beyond the teacher's own biases.

\section{CHV Analysis}
\label{app:chv}

\subsection{Vocabulary Sources and Coverage}
\label{app:vocab}

The primary source for patient-friendly synonyms is the Consumer Health Vocabulary (CHV)~\cite{zeng2006exploring}.
When CHV lacks a lay synonym for an extracted concept, we sequentially consult four fallback sources:
(1)~UMLS preferred names~\cite{bodenreider2004unified},
(2)~Medical Subject Headings (MeSH)~\cite{mesh2024},
(3)~Systematized Nomenclature of Medicine Clinical Terms (SNOMED~CT)~\cite{bos2006snomed}, and
(4)~National Cancer Institute (NCI) Thesaurus~\cite{sioutos2007nci}.
These are standard clinical vocabularies adopted in electronic health records and biomedical indexing systems.

The HOVER Step~1 mapping table contains 2,411 CUI entries.
CHV directly covers 1,266 of them (52.5\%); the remaining entries fall back through the four additional sources before being marked unmapped (Table~\ref{tab:chv-coverage}).
All 2,306 unique CUIs that actually appear in the released dataset are present in the mapping table.

\begin{table}[h!]
\centering
\caption{Patient-friendly term sourcing in the released mapping table.}
\label{tab:chv-coverage}
\begin{tabular}{lrr}
\toprule
Source & \# CUIs & \% of 2,411 \\
\midrule
CHV               & 1,266 & 52.5\% \\
NCI Thesaurus     & 228   & 9.5\%  \\
UMLS preferred    & 225   & 9.3\%  \\
SNOMED CT (US)    & 172   & 7.1\%  \\
MeSH              & 113   & 4.7\%  \\
MedlinePlus       & 3     & 0.1\%  \\
\midrule
Mapped (any source)   & 2,007 & 83.2\% \\
Unmapped (no rewrite) & 404   & 16.8\% \\
\bottomrule
\end{tabular}
\end{table}

Although CHV covers only 52.5\% of unique CUIs, it accounts for 85.8\% of entity tokens in the released captions because CHV-mapped terms (e.g., MRI, lesion, tumor) are disproportionately frequent.
Across all captions, the average per-caption mapping rate is 0.963.

\subsection{Top Patient-Friendly Substitutions Actually Applied}
\label{app:chv-applied}

The CHV mapping is provided as a soft reference: rule~6 of the rewriting prompt instructs the LLM to ignore entries that are not genuinely simpler (\S\ref{app:prompt}).
To verify that the pipeline produces meaningful register shifts in practice, we directly audit the released expert/lay caption pairs.
For each medical term $e$, we count how many captions contain $e$ in the expert version and how many successfully replace it with a patient-friendly alternative $l$ in the lay version; the replacement rate is the ratio of the two.
We exclude morphological variants (e.g., singular/plural) and report only substitutions with at least 100 occurrences and a replacement rate above 50\%.

\begin{table}[h!]
\centering
\caption{Top 15 register-shifting substitutions realized in the released lay captions.
Source indicates whether the $e \to l$ pair appears in the CHV/UMLS lookup table (CHV) or was produced freely by the LLM (LLM).}
\label{tab:chv-top15}
\begin{tabular}{rllllrrr}
\toprule
\# & Expert term & $\to$ & Lay term & Source & in\_exp & Replaced & Rate \\
\midrule
 1 & magnetic resonance imaging & $\to$ & MRI           & CHV & 40,363 & 34,273 & 85\% \\
 2 & parenchyma      & $\to$ & tissue          & LLM & 23,509 & 22,523 & 96\% \\
 3 & cerebral         & $\to$ & brain           & LLM & 17,755 & 13,513 & 76\% \\
 4 & hepatic          & $\to$ & liver           & LLM & 12,618 & 12,327 & 98\% \\
 5 & thoracic         & $\to$ & chest           & CHV &  6,384 &  6,010 & 94\% \\
 6 & atypical         & $\to$ & abnormal        & LLM &  6,707 &  5,052 & 75\% \\
 7 & renal            & $\to$ & kidney          & LLM &  4,350 &  3,653 & 84\% \\
 8 & benign           & $\to$ & non-cancerous   & LLM &  6,186 &  3,405 & 55\% \\
 9 & pulmonary        & $\to$ & lung            & LLM &  3,244 &  2,621 & 81\% \\
10 & neoplasm         & $\to$ & tumor           & LLM &  4,212 &  2,585 & 61\% \\
11 & cranial cavity   & $\to$ & brain cavity    & CHV &  4,322 &  2,443 & 57\% \\
12 & cardiac          & $\to$ & heart           & LLM &  1,970 &  1,786 & 91\% \\
13 & vasculature      & $\to$ & blood vessels   & LLM &    824 &    708 & 86\% \\
14 & gastric          & $\to$ & stomach         & LLM &    668 &    566 & 85\% \\
15 & cervical         & $\to$ & neck            & LLM &    536 &    341 & 64\% \\
\bottomrule
\end{tabular}
\end{table}

\paragraph{LLM is the dominant source of patient-friendly register.}
Eleven of the fifteen entries are not CHV/UMLS-derived: the LLM freely rewrites ``parenchyma''~$\to$~``tissue'', ``hepatic''~$\to$~``liver'', ``cardiac''~$\to$~``heart'', ``cerebral''~$\to$~``brain'', etc., because the consumer vocabularies either lack a clean entry or provide an awkward one.
The CHV table contributes only the high-frequency anchors ``magnetic resonance imaging''~$\to$~``MRI'' and ``thoracic''~$\to$~``chest''; the rest of the register shift is the LLM's own work, with the soft term map acting as guard-rails rather than a substitution dictionary.

\paragraph{Rejection of awkward CHV mappings.}
The soft-reference design allows the LLM to override unsuitable CHV suggestions.
For instance, the CHV maps ``lesion'' to ``lesions'' (a mere plural change), but the LLM instead substitutes patient-friendly alternatives such as ``growth'' or ``abnormal area'' in 98\% of cases.
Likewise, ``tumor'' is mapped by CHV to ``tissue tumor'', yet the LLM retains the original ``tumor'' in 88\% of cases as it is already widely understood by patients.

\paragraph{Unmapped concepts.}
The 404 CUIs (16.8\%) that no consumer vocabulary covers are passed through unchanged at Step~2, retaining the original wording.
These are dominated by rare anatomical sub-structures and modality-specific imaging descriptors (e.g., MR sequence names) that have no widely accepted plain-language equivalent.

\section{Dataset Intrinsic Quality Metrics}
\label{app:intrinsic}
 
To validate that the \hover{} pipeline preserves clinical semantics while improving readability, we evaluate the released expert-lay caption pairs ($n$\,=\,122,789) using four independent metrics computed on the dataset itself.
 
\begin{table}[h!]
\centering
\caption{Intrinsic quality metrics on the released \oursdata{} expert-lay pairs.
GREEN, BERTScore, and RaTEScore measure semantic preservation between expert and lay captions; readability deltas measure simplification effectiveness.}
\label{tab:intrinsic_quality}
\footnotesize
\begin{tabular}{llccc}
\toprule
\textbf{Metric} & \textbf{Aspect} & \textbf{Train} & \textbf{Val} & \textbf{Test} \\
\midrule
GREEN~\cite{ostmeier2024green} & Clinical fidelity & 0.774 & 0.769 & 0.759 \\
BERTScore-F~\cite{zhang2019bertscore} & Semantic similarity & 0.890 & 0.886 & 0.878 \\
RaTEScore~\cite{zhao2024ratescore} & Entity preservation & 0.779 & 0.774 & 0.762 \\
\midrule
$\Delta$FK grade$\downarrow$ & Readability gain & $-$2.30 & $-$2.22 & $-$2.28 \\
$\Delta$Flesch ease$\uparrow$ & Readability gain & $+$16.9 & $+$16.7 & $+$16.8 \\
$\Delta$Word count & Length change & $+$0.7 & $+$0.6 & $+$0.8 \\
\bottomrule
\end{tabular}
\end{table}

The results confirm that the \hover{} pipeline achieves high semantic preservation between expert and lay captions: GREEN (0.759--0.774), BERTScore-F (0.878--0.890), and RaTEScore (0.762--0.779) all indicate strong clinical fidelity across splits.
Simultaneously, the readability deltas show that lay captions are consistently easier  to read than their expert counterparts, with a Flesch--Kincaid~\cite{flesch1948new} grade reduction of 2.3 levels and a Flesch Reading Ease increase of 16.8 points on average, while maintaining nearly identical word counts ($\Delta$\,=\,$+$0.7 words).
This validates that the pipeline improves patient accessibility without sacrificing clinical content or inflating caption length.

\section{Prompt Templates}
\label{app:prompt}

This section reproduces every prompt used in the data construction and benchmark evaluation pipelines.
All prompts match the released codebase.
We use standardized three-segment prompt templates for all VLM evaluations, instructing the model to generate descriptions covering
(1)~modality and anatomical context,
(2)~region-of-interest findings, and
(3)~clinical interpretation.
For lay captioning, the prompt additionally specifies 8th-grade readability and parenthetical retention of original medical terms on first mention.

\subsection{HOVER Step~2: LLM-Based Constrained Rewriting}

\textbf{Model.} Llama-3.1-70B-Instruct~\cite{grattafiori2024llama}, served via vLLM~\cite{kwon2023efficient}.
\textbf{Decoding.} Temperature~$0.3$, top-$p$~$0.95$, maximum generation length~$768$ tokens, with up to three retries on failure and a 180\,s timeout per request.

\paragraph{System prompt.}
The CHV-derived term map is treated as a soft reference, not as hard substitutions: rules~5--6 explicitly instruct the rewriter to ignore mappings that introduce errors or remain technical.

{\small
\begin{verbatim}
You are a medical communication specialist trained in health
literacy. Your task is to rewrite expert radiology/pathology
descriptions into patient-friendly language.

ABSOLUTE RULES -- STRUCTURE PRESERVATION:
1. Keep the EXACT same number of sentences and the EXACT
   same order.
2. Do NOT delete, merge, split, or rearrange any sentences.
3. Every sentence in the original MUST have a corresponding
   sentence in your output, in the same position.
4. Do NOT add new sentences or new information not present
   in the original.

LANGUAGE SIMPLIFICATION:
5. Replace medical jargon with plain, everyday language a
   middle school student can understand.
6. You are provided REFERENCE term mappings from UMLS. These
   may contain errors or still-technical terms (e.g.,
   "hepatopathy", "intraabdominal", "catting"). Use a
   mapping ONLY if it is genuinely simpler. Otherwise,
   IGNORE it and use your own plain language.
7. Include the original expert term in parentheses after the
   simplified term on FIRST mention only.
   Example: "enlarged lymph nodes (lymphadenomegaly)"
8. Maintain anatomical specificity -- do not generalize
   locations.

CONTENT RULES:
9. Do NOT provide medical advice, prognosis, or reassurance.
10. Write in third person descriptive style, not addressing
    the patient directly.
11. Reference the highlighted region (green box) when the
    original does.

Output ONLY the rewritten caption. No preamble, no
explanation.
\end{verbatim}
}

\paragraph{User prompt template.}
The full Step~1 metadata (term mappings, unmapped terms, modality, semantic groups) is exposed in clearly labelled markdown sections:

{\small
\begin{verbatim}
Rewrite the following expert description into
patient-friendly language.

IMPORTANT: Keep the EXACT same sentence structure and order.
Do NOT delete or rearrange sentences.

## Expert Caption
{expert_caption}

## Reference Term Mappings (UMLS)
These are REFERENCE ONLY. They may contain errors or
still-technical terms. Use them ONLY if genuinely simpler.
Otherwise, use your own plain language.
- "abdominal ct scan" -> "abdomen ct"
- "hepatic lesion" -> "liver mass"
- ... (one per CUI matched in this caption) ...

Additional terms with no reference (simplify freely):
  signal intensity, ground-glass opacity, ...

## Context
- Modality: CT
- Semantic Groups: Procedures, Anatomy, Disorders

Return ONLY the rewritten caption.
Same sentences, same order, simpler words.
\end{verbatim}
}

Mappings annotated in Step~1 as unmapped are passed as \texttt{(no reference, simplify freely)} so the model is explicitly invited to produce a plain-language substitute.

\subsection{HOVER Step~3: Cross-Model Visual Verification}

\textbf{Model.} Qwen3.5-27B, served via SGLang~\cite{zheng2024sglang} with 8-way tensor parallelism and a 131,072-token context window.
\textbf{Decoding.} Temperature~$0.1$, maximum generation length~$2,048$ tokens, with thinking mode disabled.
The verifier runs asynchronously with a concurrency of~32 and a 300\,s per-request timeout (3~retries).

\paragraph{Image preprocessing.}
Each image is loaded, converted to RGB, resized so the longest edge does not exceed $1,024$\,px (Lanczos), then JPEG-encoded at quality~85 and passed to the verifier as a base64 image.

\paragraph{System prompt.}

{\small
\begin{verbatim}
You are a senior radiologist performing quality assurance on
medical image descriptions. You will receive a medical image
along with two text descriptions:
1. Expert Caption: the original expert-level description
2. Lay Caption: a patient-friendly rewrite of the expert
   caption

Your task is to verify the lay caption by examining the
image and comparing both texts.
\end{verbatim}
}

\paragraph{User prompt (multimodal).}
The verifier receives the image together with a structured rubric containing explicit $0$--$2$ anchors per attribute, the verdict policy, an expert-error flag, and a strict JSON output schema:

{\small
\begin{verbatim}
## Verification Task
Examine the medical image and evaluate the Lay Caption
against the Expert Caption.

## Expert Caption (reference)
{expert_caption}

## Lay Caption (to verify)
{lay_caption}

## Stated Modality: {modality}

## Evaluate the following 5 criteria. Score each 0-2:

1. **Modality Consistency** (0-2)
   - Does the lay caption correctly identify the imaging
     modality?
   - Does the stated modality match what you see in the
     image?
   - 2=correct, 1=vague but acceptable, 0=wrong modality

2. **Anatomical Accuracy** (0-2)
   - Are all organs/structures from the expert caption
     preserved?
   - Are anatomical locations correct based on the image?
   - 2=all preserved, 1=minor omission,
     0=major omission or error

3. **Finding Completeness** (0-2)
   - Are all abnormal findings from the expert caption
     included?
   - No clinical observation should be lost in
     simplification.
   - 2=all findings present, 1=minor finding missing,
     0=key finding missing

4. **Factual Correctness** (0-2)
   - Does the lay caption introduce any factually incorrect
     information?
   - Does it contradict what you see in the image?
   - 2=no errors, 1=minor inaccuracy, 0=factual error

5. **Lay Readability** (0-2)
   - Is the lay caption genuinely easier to understand than
     the expert version?
   - Are expert terms properly replaced with lay
     equivalents?
   - 2=clearly patient-friendly, 1=partially simplified,
     0=still too technical

## Instructions:
- If total_score is 10: set verdict to PASS, leave
  revised_lay_caption empty.
- If total_score is 7-9: set verdict to REVISE, and
  provide a revised_lay_caption that fixes the identified
  issues while keeping the same sentence structure.
- If total_score < 7: set verdict to FAIL, and provide a
  revised_lay_caption that is a complete rewrite in
  patient-friendly language.

## Expert Caption Quality Flag
Additionally, flag if the EXPERT caption itself contains
errors:
- expert_error: true/false
- expert_error_reason: (brief explanation if true)

## Output Format (JSON only, no other text)
{
    "scores": {
        "modality_consistency": <0-2>,
        "anatomical_accuracy": <0-2>,
        "finding_completeness": <0-2>,
        "factual_correctness": <0-2>,
        "lay_readability": <0-2>
    },
    "total_score": <0-10>,
    "verdict": "<PASS|REVISE|FAIL>",
    "fail_reasons": ["<reason1>", ...],
    "revised_lay_caption": "<improved lay caption or empty
                             string if PASS>",
    "expert_error": <true|false>,
    "expert_error_reason": "<reason or empty string>",
    "brief_note": "<one-sentence summary, max 30 words>"
}
\end{verbatim}
}

\paragraph{Verdict policy and re-verification loop.}
Total score $S = \sum_{j=1}^{5} s_j \in [0,10]$.
Samples receive PASS when $S = 10$, REVISE when $7 \le S < 10$, and FAIL when $S < 7$.
The verifier returns a \texttt{revised\_lay\_caption} on every non-PASS verdict; FAIL rows enter a re-verification loop that re-scores the revised caption for up to two rounds.
Rows that remain below the threshold after both rounds are excluded from the released dataset (cf.\ \S\ref{app:hover}).
Each row's final lay caption is the verifier's revised draft when the verdict is REVISE or FAIL-resolved, and the Step~2 draft otherwise.

\subsection{Task~A: Image-to-Lay Prompt}

Used by all 33 benchmarked VLMs.
The model receives only the medical image with ROI bounding box.

{\small
\begin{verbatim}
Explain this medical image to a patient with no medical
background.
The green bounding box highlights the area your doctor wants
you to understand.

Structure your response in three parts:
1. What type of image this is and what body part is shown.
2. What the highlighted area shows and where it is located.
3. What this finding means in simple terms.

When using a medical term, add a simple definition in
parentheses.
\end{verbatim}
}

\subsection{Task~B: Expert-Guided Lay Prompt}

The model receives the medical image with ROI bounding box \emph{and} the ground-truth expert caption.

{\small
\begin{verbatim}
You are given a medical image with a green bounding box
highlighting the region of interest, along with the
expert-level clinical description below.

=== Expert Description ===
{expert_caption}
=========================

Rewrite this expert description for a patient with no
medical background.

Structure your response in three parts:
1. What type of image this is and what body part is shown.
2. What the highlighted area shows and where it is located.
3. What this finding means in simple terms.

When using a medical term, add a simple definition in
parentheses.
\end{verbatim}
}

\subsection{Task~C: Text-Only Lay Prompt}

The model receives only the expert caption; no image is provided.

{\small
\begin{verbatim}
You are given the expert-level clinical description of a
medical image finding. No image is provided.

=== Expert Description ===
{expert_caption}
=========================

Rewrite this expert description for a patient with no
medical background.

Structure your response in three parts:
1. What type of image this is and what body part is shown.
2. What the highlighted area shows and where it is located.
3. What this finding means in simple terms.

When using a medical term, add a simple definition in
parentheses.
\end{verbatim}
}

\subsection{MedLayEval (Distilled Evaluator) Input Format}

\textbf{Model.} Qwen2.5-VL-3B-Instruct + LoRA + 5-attribute regression head.
The student does not generate text; it returns five sigmoid scores from the attention-masked mean-pooled last hidden state.
The chat-template-formatted input is:

{\small
\begin{verbatim}
[IMAGE]
<expert>{expert_caption}</expert>
<lay>{candidate_lay_caption}</lay>
\end{verbatim}
}

Both caption fields are truncated to 1,500 characters; the full message is processed with \texttt{max\_length}\,=\,2,048 and image resolution capped at $448 \times 448$\,px.

\section{\ourseval{}: Medical Expert-to-Lay Evaluator}
\label{app:medlayeval_ablation}

\paragraph{Training Data.}

The \hover{} pipeline produces $\sim$132K candidate samples, of which 122,789 pass verification and form the released \oursdata{} benchmark.
Rather than discarding the remaining samples, we leverage the full pipeline output to train \ourseval{}: PASS-verdict samples provide high-scoring positive examples via their verified final captions, REVISE-verdict samples yield informative mid-range examples via their pre-revision Step~2 drafts, and FAIL-verdict samples provide naturally low-scoring examples that are excluded from the benchmark but valuable for calibrating the evaluator's lower score range.
We further construct two synthetic sources to cover extreme cases: expert-as-lay copies (clinical fidelity preserved but readability penalized) and image-unrelated text pairs (all attributes penalized).

\ourseval{} is distilled from the 27B teacher verifier using 105,100 training samples and 12,061 validation samples, both drawn exclusively from the benchmark's train and val splits.
The held-out test split (26,453 samples) is never seen during training, ensuring unbiased evaluation on the benchmark.
All labels are derived from the 27B teacher's attribute-level scores, normalized to $[0,1]$, and the training loss uses inverse-frequency weighting per attribute to prevent majority-class dominance.
\Cref{tab:medlayeval_data} summarizes the data composition.

\begin{table}[h!]
\centering
\caption{\ourseval{} data composition.
The test split is held out entirely from training and used only for performance assessment (\Cref{tab:test_performance}).}
\label{tab:medlayeval_data}
\footnotesize
\begin{tabular}{llrl}
\toprule
\textbf{Role} & \textbf{Source} & $n$ & \textbf{Description} \\
\midrule
\multirow{5}{*}{Train}
  & \textsc{Pass-Final}    & 36,805 & Verified lay captions (score 10/10); high scores \\
  & \textsc{Draft}         & 50,354 & Pre-revision Step~2 outputs; mid-range scores \\
  & \textsc{Fail}          &  7,041 & HOVER-rejected samples; low clinical scores \\
  & \textsc{Expert-Copy}   &  7,229 & Expert caption as lay; readability\,=\,0 \\
  & \textsc{Adv-Negative}  &  3,671 & Unrelated text as lay; all scores\,=\,0 \\
\midrule
  & Total                  & 105,100 & \\
\midrule
Val  & Same five sources    & 12,061 & Disjoint from training \\
\midrule
Test & Held-out benchmark test & 26,453 & Never seen during training \\
\bottomrule
\end{tabular}
\end{table}

\paragraph{Test Discrimination.}

\Cref{tab:test_performance} evaluates discrimination on the held-out test split (26,453 samples), which the evaluator has never seen during training.
The evaluator cleanly separates accepted and rejected samples: verified captions (PASS+REVISE) receive a mean predicted score of 0.70, while rejected samples (FAIL) score 0.36, yielding a separation margin of 0.34.
Among rejected samples, only 2.4\% exceed a predicted score of 0.7, while 41.4\% fall below 0.3, confirming that the evaluator reliably identifies low-quality expert-to-lay translations.

\begin{table}[h!]
\centering
\caption{\ourseval{} discrimination on pre-filtering test candidates ($n$\,=\,26,453), including 1,857 HOVER-rejected samples excluded from the released benchmark.}
\label{tab:test_performance}
\footnotesize
\begin{tabular}{lrrrrr}
\toprule
Verdict & $n$ & Pred mean & \% pred $>$0.7 & \% pred $<$0.3 \\
\midrule
Accepted (PASS+REVISE) & 24,596 & 0.70 & 67.3 & 1.9 \\
Rejected (FAIL)        &  1,857 & 0.36 &  2.4 & 41.4 \\
\bottomrule
\end{tabular}
\end{table}

\paragraph{Design Ablation.}

We validate the regression-based design by comparing against a generative alternative that outputs score tokens (e.g., ``Modality: 0.8, Anatomy: 0.7'').
The generative variant suffers from mode collapse, defaulting to a narrow score range regardless of input quality, whereas the regression head produces well-calibrated continuous scores across the full $[0,1]$ range.
This confirms that replacing the language modeling head with a dedicated regression head is essential for reliable evaluation.

\paragraph{Computational Efficiency.}
Unlike generative LLM-as-a-judge metrics such as GREEN~\cite{ostmeier2024green} that require autoregressive decoding of score explanations, \ourseval{} produces all five attribute scores in a single forward pass through the 3B Qwen2.5-3B~\cite{bai2025qwen25vltechnicalreport} backbone.
\Cref{tab:efficiency} compares the computational cost against existing medical text evaluation metrics on a single NVIDIA B200 GPU.
\ourseval{} achieves $\sim$88$\times$ faster inference than the 27B teacher verifier from which it is distilled and $\sim$34$\times$ faster than GREEN, while being the only deployable-scale metric that is both visually grounded and reference-free across diverse imaging modalities.

\begin{table}[h!]
\centering
\caption{Computational comparison of medical text evaluation metrics (single NVIDIA B200, batch 1, bfloat16).
GREEN~\cite{ostmeier2024green} is included as a representative LLM-as-a-judge baseline designed for chest X-ray reports.
Autoregressive baselines assume 200--300 generated tokens.}
\label{tab:efficiency}
\footnotesize
\begin{tabular}{lrrcc}
\toprule
Metric & Params & ms/sample & Image & Ref-free \\
\midrule
BERTScore~\cite{zhang2019bertscore}        & 0.07B & $\sim$12   & \ding{55}  & \ding{55}  \\
RaTEScore~\cite{zhao2024ratescore}         & 0.29B & $\sim$220  & \ding{55}  & \ding{55}  \\
RadEval~\cite{xu2025radeval}               & 0.55B & $\sim$880  & \ding{55}  & \ding{55}  \\
GREEN~\cite{ostmeier2024green}             & 7B    & $\sim$3.2k & \ding{55}  & \ding{55}  \\
\ourseval{} (ours)                         & 3.0B  & $\sim$95   & \checkmark & \checkmark \\
\midrule
27B Teacher (HOVER Step~3)                 & 27B   & $\sim$8.4k & \checkmark & \checkmark \\
\bottomrule
\end{tabular}
\end{table}

Both the evaluator weights and all training data are released for full reproducibility.


\section{Task Definitions and Evaluation Subset}
\label{app:task_bc}

\paragraph{Task~A (Primary).}
Given a medical image with ROI bounding box, the model generates a lay description following the standardized three-segment structure (\Cref{app:prompt}).
All 33 models are evaluated on a shared modality-stratified 5,000-sample subset of the test split: MRI (2,957), Pathology (1,500), CT (521), X-Ray (13), PET (6), Dermoscopy (2), and Endoscopy (1).
Subset sample IDs are released in \texttt{\_shared\_ids.json}.

\paragraph{Task~B (Expert-Guided Lay Generation).}
The model receives the medical image, ROI bounding box, and the ground-truth expert caption.
This isolates register transfer capability by removing the need for the model to independently identify clinical findings.

\paragraph{Task~C (Text-Only Lay Generation).}
The model receives only the expert caption without the image.
This quantifies the contribution of visual grounding by testing whether text-only rewriting can preserve clinical accuracy.

\paragraph{Ablation Subset.}
Tasks~B and~C are evaluated on a modality-stratified 500-sample subset drawn from the Task~A evaluation pool using the top-3 proprietary models.
Table~\ref{tab:task_ablation} reveals two key findings:
(1)~providing the ground-truth expert caption (Task~B) consistently boosts all five \ourseval{} attributes over the image-only baseline (Task~A), with gains of $+1.3$ to $+8.3$ points in overall score, confirming that lay simplification becomes substantially easier when clinical findings need not be independently identified;
(2)~removing the image entirely (Task~C) incurs minimal degradation relative to Task~B for Gemini~2.5 Flash ($-0.2$) and even improves GPT-5.4-mini ($+1.2$), suggesting that once the expert caption is available, visual grounding contributes little, and the bottleneck lies in register transfer, not image understanding.
Claude~Opus~4.7 exhibits a larger B$\to$C drop ($-2.1$), indicating it leverages the image more actively for cross-referencing during simplification.

\begin{table}[h!]
\centering
\caption{\textbf{Task ablation on a shared 500-sample subset.}
Task~A scores are from the 5K evaluation (\Cref{tab:leaderboard_full}); Tasks~B and~C are evaluated on the same 500 samples.
Best in \textbf{bold}, second \underline{underlined} per task.
$\dagger$\,=\,thinking enabled.}
\label{tab:task_ablation}
\footnotesize
\begin{tabular}{l c c ccccc}
\toprule
\textbf{Model} & \textbf{Task} & \textbf{$S$} & Mod & Anat & Find & Fact & Read \\
\midrule
\multicolumn{8}{l}{\cellcolor{gray!8}\textit{Task~A: Image $\to$ Lay}} \\
Gemini 2.5 Flash$^\dagger$  & A & \textbf{70.6} & \textbf{67.6} & \textbf{75.4} & \textbf{78.5} & \textbf{62.0} & \textbf{69.6} \\
Claude Opus 4.7              & A & \underline{64.1} & \underline{60.3} & \underline{67.3} & \underline{71.6} & \underline{55.4} & \underline{65.9} \\
GPT-5.4-mini                 & A & 61.2 & 59.0 & 64.4 & 68.1 & 53.4 & 61.3 \\
\midrule
\multicolumn{8}{l}{\cellcolor{gray!8}\textit{Task~B: Image + Expert $\to$ Lay}} \\
Gemini 2.5 Flash$^\dagger$  & B & \textbf{73.5} & \textbf{70.3} & \textbf{77.3} & \textbf{80.9} & \textbf{66.0} & \underline{72.8} \\
Claude Opus 4.7              & B & \underline{72.4} & \underline{68.8} & \underline{76.5} & \underline{80.3} & \underline{63.8} & \textbf{72.9} \\
GPT-5.4-mini                 & B & 62.5 & 60.8 & 64.7 & 69.5 & 54.3 & 63.0 \\
\midrule
\multicolumn{8}{l}{\cellcolor{gray!8}\textit{Task~C: Expert Text $\to$ Lay (no image)}} \\
Gemini 2.5 Flash$^\dagger$  & C & \textbf{73.3} & \textbf{69.5} & \textbf{77.2} & \textbf{80.8} & \textbf{65.4} & \textbf{73.5} \\
Claude Opus 4.7              & C & \underline{70.3} & \underline{66.2} & \underline{74.1} & \underline{78.3} & \underline{61.5} & \underline{71.6} \\
GPT-5.4-mini                 & C & 63.7 & 62.2 & 66.1 & 70.8 & 55.7 & 63.5 \\
\bottomrule
\end{tabular}
\end{table}

\begin{figure}[h!]
\centering
\includegraphics[width=\textwidth]{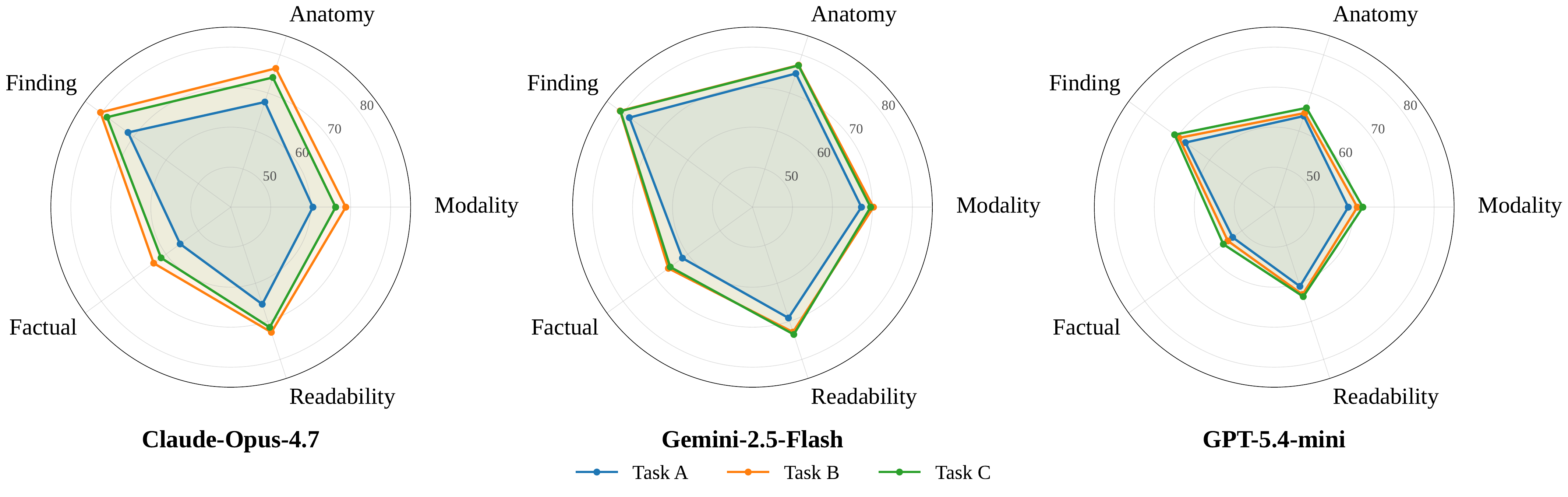}
\caption{\textbf{Per-attribute \ourseval{} scores across Tasks~A, B, and C for the top-3 proprietary models ($n$\,=\,500).}
Providing the expert caption (Task~B) uniformly expands the score polygon relative to the image-only baseline (Task~A).
Removing the image (Task~C) incurs negligible degradation for Gemini~2.5~Flash and GPT-5.4-mini, whereas Claude~Opus~4.7 shows a visible contraction in Anatomy and Finding, suggesting stronger reliance on visual grounding.}
\label{fig:radar_task_abc}
\end{figure}

\section{Qualitative Examples}
\label{app:qualitative}

This section presents qualitative examples of expert-to-lay rewriting produced by the \hover{} pipeline across all eight imaging modalities in \oursdata{}.
For each modality, we show the medical image with ROI annotation, the original expert caption with \ehigh{medical terminology highlighted}, and the corresponding lay caption with \lhigh{patient-friendly substitutions highlighted}.
These examples illustrate how the pipeline preserves clinical content while systematically replacing jargon with accessible language.

\paragraph{MRI.}
This brain MRI sample (BraTS2018, T2-weighted) demonstrates a transverse section with a small region of abnormal signal intensity near the midline.
The expert caption employs standard neuroradiology terminology (\ehigh{cerebral hemispheres}, \ehigh{ventricular system}, \ehigh{edema}), all of which the pipeline maps to plain-language equivalents while preserving the anatomical specificity of the description.

\begin{figure}[H]
\centering
\begin{minipage}[c]{0.35\linewidth}
  \tcbox[
    boxrule=0.4pt, colback=white, colframe=black!60,
    arc=4pt, boxsep=0pt, left=0pt, right=0pt, top=0pt, bottom=0pt
  ]{%
    \includegraphics[width=\linewidth]{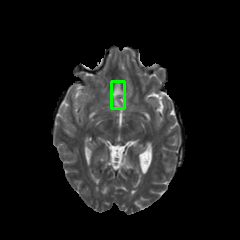}%
  }
\end{minipage}%
\hspace{12pt}%
\begin{minipage}[c]{0.35\linewidth}
  \footnotesize
  \begin{tabular}{@{}l@{\hspace{8pt}}l@{}}
  \multicolumn{2}{@{}l}{\textbf{Disorders}} \\
    \textit{T046 Pathologic Function} & \texttt{C0013604} edema \\
    \textit{T046 Pathologic Function} & \texttt{C0030660} pathological process \\[4pt]
  \multicolumn{2}{@{}l}{\textbf{Anatomy}} \\
    \textit{T023 Body Part, Organ} & \texttt{C0006104} human brain \\
  \end{tabular}
\end{minipage}
\end{figure}

\begin{tcolorbox}[
  enhanced, title={\small Expert Caption}, 
  colback=white, colframe=black!70, 
  coltitle=white, colbacktitle=black!70,
  boxrule=0.4pt, arc=2pt, fonttitle=\bfseries\small,
  left=4pt, right=4pt, top=2pt, bottom=2pt
]
\small
The image is a transverse section \ehigh{Magnetic Resonance Imaging (MRI)} scan of the \ehigh{human brain}, showcasing the \ehigh{cerebral hemispheres}, \ehigh{ventricular system}, and surrounding brain structures without any medical devices present.
The region of interest, located centrally and in the upper-middle area of the image, occupying approximately 0.6\% of the area, exhibits \ehigh{abnormal signal intensity}, which is indicative of \ehigh{edema}.
This region is situated near the midline structures of the brain, possibly affecting or being affected by adjacent tissues due to its proximity, suggesting a potential relationship where the \ehigh{edema} could be causing pressure effects on nearby structures or could be a result of a \ehigh{pathological process} involving adjacent tissues.
\end{tcolorbox}

\begin{tcolorbox}[
  enhanced, title={\small Lay Caption}, 
  colback=white, colframe=black!70, 
  coltitle=white, colbacktitle=black!70,
  boxrule=0.4pt, arc=2pt, fonttitle=\bfseries\small,
  left=4pt, right=4pt, top=2pt, bottom=2pt
]
\small
The image is a \lhigh{cross-section picture} of the \lhigh{brain}, taken using a special machine called an MRI, showing the \lhigh{brain}'s different parts and the spaces around them without any medical equipment in the way.
The highlighted area, located centrally and in the upper-middle part of the image, shows a bright spot indicating \lhigh{swelling (edema)}.
This area is near the middle of the \lhigh{brain}, possibly affecting or being affected by nearby tissues, suggesting the swelling could be causing pressure on surrounding structures or is a result of a \lhigh{disease process} in that area.
\end{tcolorbox}

\paragraph{CT.}
This abdominal CT sample (FLARE23) shows a pancreatic lesion characterized by abnormal texture.
The expert caption uses organ-specific clinical terms (\ehigh{pancreatic cancer}, \ehigh{localized disease}), which the pipeline rewrites into direct, conversational language while retaining the diagnostic specificity of the finding.

\begin{figure}[H]
\centering
\begin{minipage}[c]{0.35\linewidth}
  \tcbox[
    boxrule=0.4pt, colback=white, colframe=black!60,
    arc=4pt, boxsep=0pt, left=0pt, right=0pt, top=0pt, bottom=0pt
  ]{%
    \includegraphics[width=\linewidth]{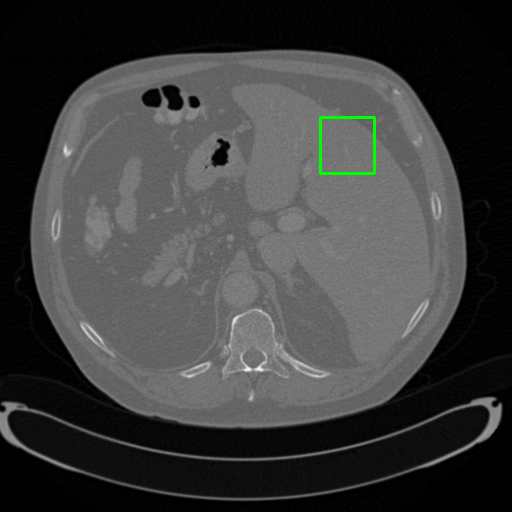}%
  }
\end{minipage}%
\hspace{12pt}%
\begin{minipage}[c]{0.35\linewidth}
  \footnotesize
  \begin{tabular}{@{}l@{\hspace{8pt}}l@{}}
  \multicolumn{2}{@{}l}{\textbf{Disorders}} \\
    \textit{T191 Neoplastic Process} & \texttt{C0006826} cancer \\
    \textit{T047 Disease or Syndrome} & \texttt{C0277565} localized disease \\[4pt]
  \multicolumn{2}{@{}l}{\textbf{Anatomy}} \\
    \textit{T023 Body Part, Organ} & \texttt{C0030274} pancreas \\
  \end{tabular}
\end{minipage}
\end{figure}

\begin{tcolorbox}[
  enhanced, title={\small Expert Caption}, 
  colback=white, colframe=black!70, 
  coltitle=white, colbacktitle=black!70,
  boxrule=0.4pt, arc=2pt, fonttitle=\bfseries\small,
  left=4pt, right=4pt, top=2pt, bottom=2pt
]
\small
The image is an \ehigh{abdominal CT scan} showing the liver, \ehigh{kidneys}, and \ehigh{pancreas}.
The liver is located in the right-center, upper-middle section of the image, with the \ehigh{kidneys} situated below and the \ehigh{pancreas} positioned centrally.
The region of interest, indicative of \ehigh{pancreatic cancer}, is characterized by an \ehigh{abnormal texture and size}, located in the \ehigh{pancreas}.
This abnormal area may be related to the surrounding pancreatic tissue, potentially representing a \ehigh{localized disease} process within the \ehigh{pancreas} that could affect the entire organ.
\end{tcolorbox}

\begin{tcolorbox}[
  enhanced, title={\small Lay Caption}, 
  colback=white, colframe=black!70, 
  coltitle=white, colbacktitle=black!70,
  boxrule=0.4pt, arc=2pt, fonttitle=\bfseries\small,
  left=4pt, right=4pt, top=2pt, bottom=2pt
]
\small
This is a \lhigh{CT scan of your abdomen} showing the liver, kidneys, and pancreas.
The liver is in the upper right, with the kidneys below and the pancreas in the center.
There is an \lhigh{abnormal area} in the pancreas that \lhigh{looks different} in texture and size, which suggests cancer.
This \lhigh{spot} is part of the pancreas and could potentially \lhigh{affect the whole organ}.
\end{tcolorbox}

\paragraph{X-Ray.}
This chest X-ray (NIH ChestX-ray) demonstrates a pneumothorax visible as a translucent area lacking lung markings.
Notably, the lay caption introduces the parenthetical term ``collapsed lung (pneumothorax)'' on first mention, following the pipeline's retention rule for original medical terms, which enables patients to cross-reference with their clinical reports.

\begin{figure}[H]
\centering
\begin{minipage}[c]{0.35\linewidth}
  \tcbox[
    boxrule=0.4pt, colback=white, colframe=black!60,
    arc=4pt, boxsep=0pt, left=0pt, right=0pt, top=0pt, bottom=0pt
  ]{%
    \includegraphics[width=\linewidth]{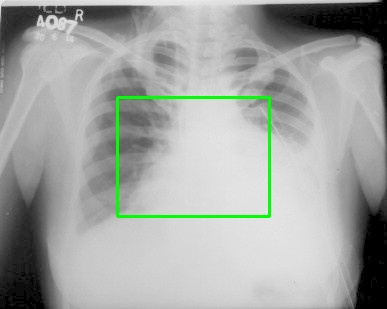}%
  }
\end{minipage}%
\hspace{12pt}%
\begin{minipage}[c]{0.35\linewidth}
  \footnotesize
  \begin{tabular}{@{}l@{\hspace{8pt}}l@{}}
  \multicolumn{2}{@{}l}{\textbf{Anatomy}} \\
    \textit{T024 Tissue} & \texttt{C0032225} pleural \\
    \textit{T024 Tissue} & \texttt{C0819757} lung tissue \\[4pt]
  \multicolumn{2}{@{}l}{\textbf{Disorders}} \\
    \textit{T046 Pathologic Function} & \texttt{C0009566} complications \\
  \end{tabular}
\end{minipage}
\end{figure}

\begin{tcolorbox}[
  enhanced, title={\small Expert Caption}, 
  colback=white, colframe=black!70, 
  coltitle=white, colbacktitle=black!70,
  boxrule=0.4pt, arc=2pt, fonttitle=\bfseries\small,
  left=4pt, right=4pt, top=2pt, bottom=2pt
]
\small
The image is a chest X-ray showing the lungs, with the heart and upper spine also visible.
The region of interest, located centrally and in the middle of the image, encompasses approximately 15\% of the area and appears to show a translucent area without lung markings, indicative of \ehigh{pneumothorax}.
This unusual transparency suggests the presence of air in the \ehigh{pleural space}, which is not normally seen in healthy \ehigh{lung tissue}.
The proximity of this region to the heart and upper spine could imply a potential impact on these structures, possibly leading to a shift in the mediastinum or other \ehigh{complications} if the \ehigh{pneumothorax} were to \ehigh{increase in size}.
\end{tcolorbox}

\begin{tcolorbox}[
  enhanced, title={\small Lay Caption}, 
  colback=white, colframe=black!70, 
  coltitle=white, colbacktitle=black!70,
  boxrule=0.4pt, arc=2pt, fonttitle=\bfseries\small,
  left=4pt, right=4pt, top=2pt, bottom=2pt
]
\small
The image is a chest X-ray showing the lungs, with the heart and upper spine also visible.
The region of interest, located centrally and in the middle of the image, encompasses approximately 15\% of the area and appears to show a translucent area without lung markings, indicative of a \lhigh{collapsed lung (pneumothorax)}.
This unusual transparency suggests the presence of air in the \lhigh{space around the lungs (pleural space)}, which is not normally seen in healthy lung tissue.
The proximity of this region to the heart and upper spine could imply a potential impact on these structures, possibly leading to a shift in the mediastinum or other complications if the collapsed lung were to \lhigh{get bigger}.
\end{tcolorbox}

\paragraph{Ultrasound.}
This transesophageal echocardiogram shows a postoperative cardiac abnormality suggestive of a pseudoaneurysm.
The pipeline translates cardiac chamber terminology (\ehigh{left ventricle}, \ehigh{right ventricle}) into intuitive anatomical descriptions and replaces the imaging modality name with its common equivalent.

\begin{figure}[H]
\centering
\begin{minipage}[c]{0.35\linewidth}
  \tcbox[
    boxrule=0.4pt, colback=white, colframe=black!60,
    arc=4pt, boxsep=0pt, left=0pt, right=0pt, top=0pt, bottom=0pt
  ]{%
    \includegraphics[width=\linewidth]{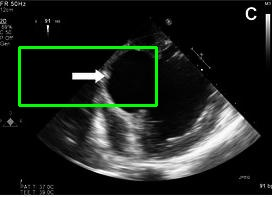}%
  }
\end{minipage}%
\hspace{12pt}%
\begin{minipage}[c]{0.35\linewidth}
  \footnotesize
  \begin{tabular}{@{}l@{\hspace{8pt}}l@{}}
  \multicolumn{2}{@{}l}{\textbf{Anatomy}} \\
    \textit{T023 Body Part, Organ} & \texttt{C0225897} left ventricle \\
    \textit{T023 Body Part, Organ} & \texttt{C0225883} right ventricle \\
    \textit{T024 Tissue} & \texttt{C1272575} cardiac tissue \\
  \end{tabular}
\end{minipage}
\end{figure}

\begin{tcolorbox}[
  enhanced, title={\small Expert Caption}, 
  colback=white, colframe=black!70, 
  coltitle=white, colbacktitle=black!70,
  boxrule=0.4pt, arc=2pt, fonttitle=\bfseries\small,
  left=4pt, right=4pt, top=2pt, bottom=2pt
]
\small
The image is a \ehigh{transesophageal echocardiogram} showing the heart and \ehigh{surrounding structures}.
The heart is located centrally with the \ehigh{left ventricle} visible on the right side of the image and the \ehigh{right ventricle} on the left side due to the perspective of the image.
The region of interest, located in the upper-middle left-center of the image, occupies approximately 15.0\% of the area and shows a periphery-and postoperative area that appears to have a different texture and intensity compared to the surrounding \ehigh{cardiac tissue}, suggesting the presence of an abnormality such as a pseudoaneurysm.
This abnormal area is adjacent to the heart and could potentially be affecting or be affected by the cardiac function, as abnormalities in this region can impact the heart's ability to perform its vital functions.
\end{tcolorbox}

\begin{tcolorbox}[
  enhanced, title={\small Lay Caption}, 
  colback=white, colframe=black!70, 
  coltitle=white, colbacktitle=black!70,
  boxrule=0.4pt, arc=2pt, fonttitle=\bfseries\small,
  left=4pt, right=4pt, top=2pt, bottom=2pt
]
\small
The image is an \lhigh{ultrasound picture} of the heart and the \lhigh{areas around it}.
The heart is in the middle with the \lhigh{left lower heart chamber} visible on the right side of the image and the \lhigh{right lower heart chamber} on the left side due to \lhigh{the way the picture was taken}.
The region of interest, located in the upper-middle left-center of the image, occupies approximately 15.0\% of the area and shows an area near a surgical site that appears to have a different texture and intensity compared to the surrounding \lhigh{heart tissue}, suggesting the presence of an abnormality such as a pseudoaneurysm.
This abnormal area is next to the heart and could potentially be affecting or be affected by the heart's function, as abnormalities in this region can impact the heart's ability to perform its vital functions.
\end{tcolorbox}

\paragraph{Pathology.}
This histological section of the thymus (TCGA) presents a small cluster of cells with irregular nuclear features.
The lay caption demonstrates the largest register shift among all modalities, replacing dense histopathology terminology (\ehigh{histological image}, \ehigh{cell morphology}, \ehigh{neoplastic process}) with everyday descriptions.
This aligns with the per-modality benchmark finding that Pathology consistently exhibits the widest Expert-Lay Gap.

\begin{figure}[H]
\centering
\begin{minipage}[c]{0.35\linewidth}
  \tcbox[
    boxrule=0.4pt, colback=white, colframe=black!60,
    arc=4pt, boxsep=0pt, left=0pt, right=0pt, top=0pt, bottom=0pt
  ]{%
    \includegraphics[width=\linewidth]{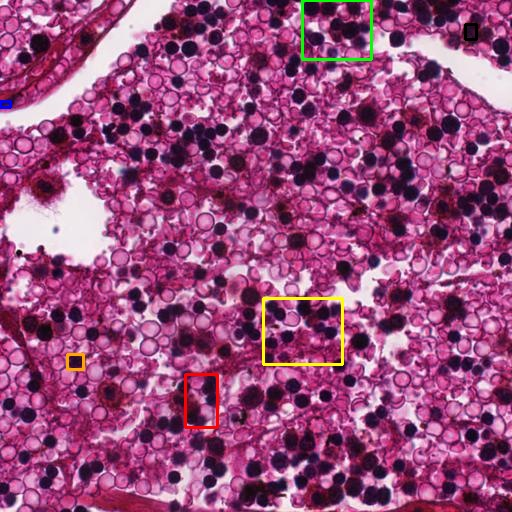}%
  }
\end{minipage}%
\hspace{12pt}%
\begin{minipage}[c]{0.35\linewidth}
  \footnotesize
  \begin{tabular}{@{}l@{\hspace{8pt}}l@{}}
  \multicolumn{2}{@{}l}{\textbf{Anatomy}} \\
    \textit{T023 Body Part, Organ} & \texttt{C0040113} thymus \\[4pt]
  \multicolumn{2}{@{}l}{\textbf{Disorders}} \\
    \textit{T049 Cell Dysfunction} & \texttt{C0333954} irregular nuclei \\
    \textit{T191 Neoplastic Process} & \texttt{C0027651} neoplasm \\
  \end{tabular}
\end{minipage}
\end{figure}

\begin{tcolorbox}[
  enhanced, title={\small Expert Caption}, 
  colback=white, colframe=black!70, 
  coltitle=white, colbacktitle=black!70,
  boxrule=0.4pt, arc=2pt, fonttitle=\bfseries\small,
  left=4pt, right=4pt, top=2pt, bottom=2pt
]
\small
The \ehigh{histological image} displays a section of the \ehigh{thymus}, with the region of interest located horizontally in the right-center and vertically in the lower-middle, occupying approximately 0.6\% of the area.
This region shows a cluster of cells with abnormal features, such as \ehigh{irregular nuclei} and \ehigh{altered staining patterns}, which are indicative of a \ehigh{pathological condition}.
The affected area is characterized by a stark contrast in \ehigh{cell morphology} compared to the surrounding tissue, suggesting a \ehigh{localized disease process}.
The unusual characteristics within this region, such as the altered nuclei and staining, may be related to the surrounding thymic tissue, potentially indicating a \ehigh{neoplastic process} that could be influencing or arising from the adjacent normal tissue.
\end{tcolorbox}

\begin{tcolorbox}[
  enhanced, title={\small Lay Caption}, 
  colback=white, colframe=black!70, 
  coltitle=white, colbacktitle=black!70,
  boxrule=0.4pt, arc=2pt, fonttitle=\bfseries\small,
  left=4pt, right=4pt, top=2pt, bottom=2pt
]
\small
This \lhigh{microscope image} shows a section of the \lhigh{thymus gland}.
A small area in the lower-right center contains a group of cells that look abnormal, with \lhigh{irregular shapes} and \lhigh{unusual coloring}, suggesting a disease.
This area looks very different from the healthy tissue around it, pointing to a problem in just that spot.
These changes might be linked to the nearby tissue, possibly indicating an \lhigh{abnormal growth} starting there.
\end{tcolorbox}

\paragraph{Endoscopy.}
This colonoscopic image (CVC-ClinicDB) shows a polypoid lesion on the colonic mucosa.
The \hover{} verifier additionally flagged an expert caption error inherited from MedTrinity-25M: the original annotation labeled this as an esophagogastroduodenoscopy (EGD), whereas the visible colonic haustra clearly indicate a colonoscopy.
The corrected expert caption and the lay caption both reflect the accurate modality.

\begin{figure}[H]
\centering
\begin{minipage}[c]{0.35\linewidth}
  \tcbox[
    boxrule=0.4pt, colback=white, colframe=black!60,
    arc=4pt, boxsep=0pt, left=0pt, right=0pt, top=0pt, bottom=0pt
  ]{%
    \includegraphics[width=\linewidth]{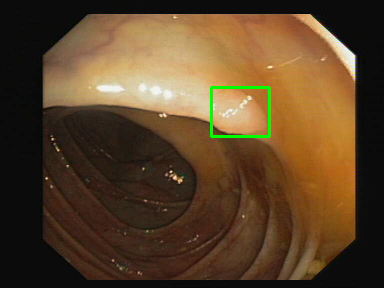}%
  }
\end{minipage}%
\hspace{12pt}%
\begin{minipage}[c]{0.35\linewidth}
  \footnotesize
  \begin{tabular}{@{}l@{\hspace{8pt}}l@{}}
  \multicolumn{2}{@{}l}{\textbf{Procedures}} \\
    \textit{T060 Diagnostic Procedure} & \texttt{C0009378} colonoscopy \\[4pt]
  \multicolumn{2}{@{}l}{\textbf{Anatomy}} \\
    \textit{T023 Body Part, Organ} & \texttt{C0009368} colon \\[4pt]
  \multicolumn{2}{@{}l}{\textbf{Disorders}} \\
    \textit{T191 Neoplastic Process} & \texttt{C0006826} cancer \\
  \end{tabular}
\end{minipage}
\end{figure}

\begin{tcolorbox}[
  enhanced, title={\small Expert Caption}, 
  colback=white, colframe=black!70, 
  coltitle=white, colbacktitle=black!70,
  boxrule=0.4pt, arc=2pt, fonttitle=\bfseries\small,
  left=4pt, right=4pt, top=2pt, bottom=2pt
]
\small
This \ehigh{colonoscopic image} displays the \ehigh{colonic lumen} obtained during an \ehigh{endoscopic examination} of the \ehigh{gastrointestinal tract} lining.
A focal area in the upper right aspect exhibits altered texture and coloration relative to the surrounding tissue, suggestive of a \ehigh{polypoid lesion}.
This abnormality is distinct from the adjacent normal \ehigh{colonic mucosa}.
This polyp represents a mucosal growth that could potentially affect nearby tissue or undergo \ehigh{malignant transformation}, as such lesions are recognized early signs of \ehigh{colorectal neoplasia}.
\end{tcolorbox}

\begin{tcolorbox}[
  enhanced, title={\small Lay Caption}, 
  colback=white, colframe=black!70, 
  coltitle=white, colbacktitle=black!70,
  boxrule=0.4pt, arc=2pt, fonttitle=\bfseries\small,
  left=4pt, right=4pt, top=2pt, bottom=2pt
]
\small
The image shows the \lhigh{inside of the colon}, taken during a procedure to examine the \lhigh{digestive tract lining}.
A specific area in the upper-right center looks different in texture and color than the surrounding tissue, suggesting a \lhigh{growth called a polyp}.
This abnormal spot is distinct from the normal lining.
This polyp is a growth arising from the colon lining that could potentially affect nearby tissue or, in some cases, develop into \lhigh{colon cancer}, as polyps can be early signs of this disease.
\end{tcolorbox}

\paragraph{PET.}
This PET scan (PMC-OA) shows an area of increased FDG uptake in the upper abdomen.
The verifier flagged an upstream expert caption error: the original annotation identified the region as upper abdomen/lower thorax, whereas the clinical context indicates the axilla/shoulder region.
The lay caption demonstrates how the pipeline handles complex radiopharmaceutical terminology, translating ``(18)F-fluorodeoxyglucose'' into ``a special sugar'' while preserving the original term in parentheses.

\begin{figure}[H]
\centering
\begin{minipage}[c]{0.35\linewidth}
  \tcbox[
    boxrule=0.4pt, colback=white, colframe=black!60,
    arc=4pt, boxsep=0pt, left=0pt, right=0pt, top=0pt, bottom=0pt
  ]{%
    \includegraphics[width=\linewidth]{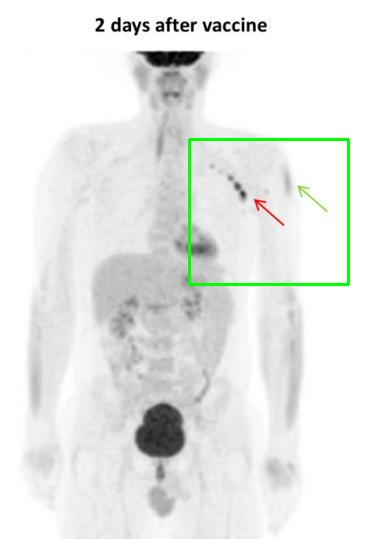}%
  }
\end{minipage}%
\hspace{12pt}%
\begin{minipage}[c]{0.35\linewidth}
  \footnotesize
  \begin{tabular}{@{}l@{\hspace{8pt}}l@{}}
  \multicolumn{2}{@{}l}{\textbf{Procedures}} \\
    \textit{T060 Diagnostic Procedure} & \texttt{C0032743} PET scan \\[4pt]
  \multicolumn{2}{@{}l}{\textbf{Anatomy}} \\
    \textit{T023 Body Part, Organ} & \texttt{C0023884} liver \\
    \textit{T023 Body Part, Organ} & \texttt{C0022646} kidney \\[4pt]
  \multicolumn{2}{@{}l}{\textbf{Disorders}} \\
    \textit{T033 Finding} & \texttt{C0332439} FDG uptake \\
  \end{tabular}
\end{minipage}
\end{figure}

\begin{tcolorbox}[
  enhanced, title={\small Expert Caption}, 
  colback=white, colframe=black!70, 
  coltitle=white, colbacktitle=black!70,
  boxrule=0.4pt, arc=2pt, fonttitle=\bfseries\small,
  left=4pt, right=4pt, top=2pt, bottom=2pt
]
\small
The image is a coronal section from a \ehigh{PET (Positron Emission Tomography)} scan, showing various organs, including parts of the torso and possibly the \ehigh{lower thoracic region} of the body.
The region of interest, located in the \ehigh{upper abdomen}, shows an area of increased uptake of \ehigh{(18)F-fluorodeoxyglucose (FDG)}, which is indicative of a benign or \ehigh{malignant process}.
The unusual uptake is significant as it may suggest a \ehigh{nonmalignant lesion} with potential clinical implications, such as an \ehigh{inflammatory process}.
The region's relative position to other organs could be related to the disease process, possibly due to a benign condition such as inflammation rather than a malignant process.
\end{tcolorbox}

\begin{tcolorbox}[
  enhanced, title={\small Lay Caption}, 
  colback=white, colframe=black!70, 
  coltitle=white, colbacktitle=black!70,
  boxrule=0.4pt, arc=2pt, fonttitle=\bfseries\small,
  left=4pt, right=4pt, top=2pt, bottom=2pt
]
\small
The image is a side view from a \lhigh{special imaging test (PET scan)}, showing various organs including parts of the torso.
The region of interest, located in the \lhigh{upper belly area}, shows an area of increased uptake of a \lhigh{special sugar} ((18)F-fluorodeoxyglucose or FDG), which can be a sign of a \lhigh{non-cancerous} or cancerous process.
The unusual uptake may suggest a \lhigh{non-cancerous} abnormality, such as \lhigh{swelling (inflammation)}.
The area's position relative to other organs could be related to the disease process, possibly due to a non-cancerous condition rather than cancer.
\end{tcolorbox}

\paragraph{Dermoscopy.}
This dermoscopic image (ISIC 2018) captures a skin lesion with visible inflammatory features.
The pipeline replaces dermatological terms (\ehigh{epidermis}, \ehigh{dermis}, \ehigh{nodule}, \ehigh{pustule}) with intuitive descriptions (``top layer,'' ``deeper layer,'' ``bump,'' ``pimple''), achieving a PASS verdict (10/10) on the first verification attempt.

\begin{figure}[H]
\centering
\begin{minipage}[c]{0.35\linewidth}
  \tcbox[
    boxrule=0.4pt, colback=white, colframe=black!60,
    arc=4pt, boxsep=0pt, left=0pt, right=0pt, top=0pt, bottom=0pt
  ]{%
    \includegraphics[width=\linewidth]{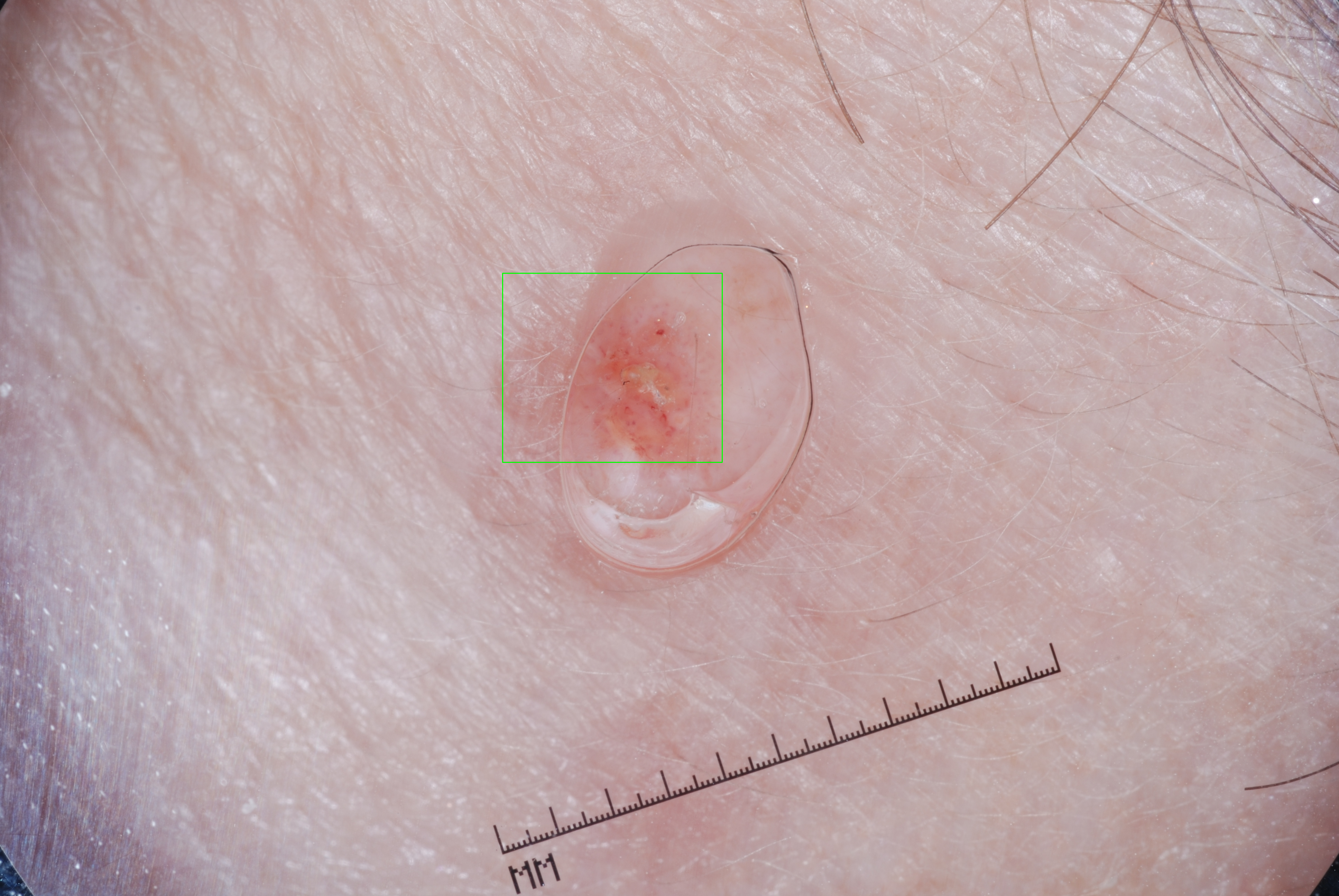}%
  }
\end{minipage}%
\hspace{12pt}%
\begin{minipage}[c]{0.35\linewidth}
  \footnotesize
  \begin{tabular}{@{}l@{\hspace{8pt}}l@{}}
  \multicolumn{2}{@{}l}{\textbf{Disorders}} \\
    \textit{T047 Disease or Syndrome} & \texttt{C0037284} skin lesion \\
    \textit{T046 Pathologic Function} & \texttt{C0021368} inflammatory response \\[4pt]
  \multicolumn{2}{@{}l}{\textbf{Anatomy}} \\
    \textit{T024 Tissue} & \texttt{C1123023} skin structure \\
  \end{tabular}
\end{minipage}
\end{figure}

\begin{tcolorbox}[
  enhanced, title={\small Expert Caption}, 
  colback=white, colframe=black!70, 
  coltitle=white, colbacktitle=black!70,
  boxrule=0.4pt, arc=2pt, fonttitle=\bfseries\small,
  left=4pt, right=4pt, top=2pt, bottom=2pt
]
\small
The \ehigh{dermoscopy image} displays a close-up view of a \ehigh{skin lesion}, with the \ehigh{epidermis} and \ehigh{dermis} visible.
The lesion is located centrally and occupies a small area of the skin's surface.
Within the region of interest, there is a notable redness and a distinct texture difference compared to the surrounding tissue, suggesting an \ehigh{inflammatory response}.
This area is characterized by a change in color and possibly texture, with the presence of a small, white, semi-transparent, slightly elevated \ehigh{nodule} or \ehigh{pustule}, which is indicative of a \ehigh{pathological change}.
\end{tcolorbox}

\begin{tcolorbox}[
  enhanced, title={\small Lay Caption}, 
  colback=white, colframe=black!70, 
  coltitle=white, colbacktitle=black!70,
  boxrule=0.4pt, arc=2pt, fonttitle=\bfseries\small,
  left=4pt, right=4pt, top=2pt, bottom=2pt
]
\small
The dermoscopy image shows a close-up view of an \lhigh{abnormal area} on the skin, with the \lhigh{top layer} and \lhigh{deeper layer} visible.
The abnormal area is located in the center and takes up a small part of the skin's surface.
Within this area, there is noticeable redness and a different texture compared to the surrounding skin, suggesting that the skin may be \lhigh{reacting to something}.
This area has a change in color and possibly texture, with a small, white, semi-transparent \lhigh{bump} or \lhigh{pimple}, which could mean that something is wrong with the skin.
\end{tcolorbox}

\newpage

\definecolor{claudeorange}{HTML}{D4764E}
\definecolor{geminiblue}{HTML}{1A73E8}
\definecolor{navyblue}{HTML}{1B3A6B}

\section{Qualitative Model Comparison}
\label{app:model_comparison}

\begin{figure}[h!]
\centering
\includegraphics[width=0.85\textwidth]{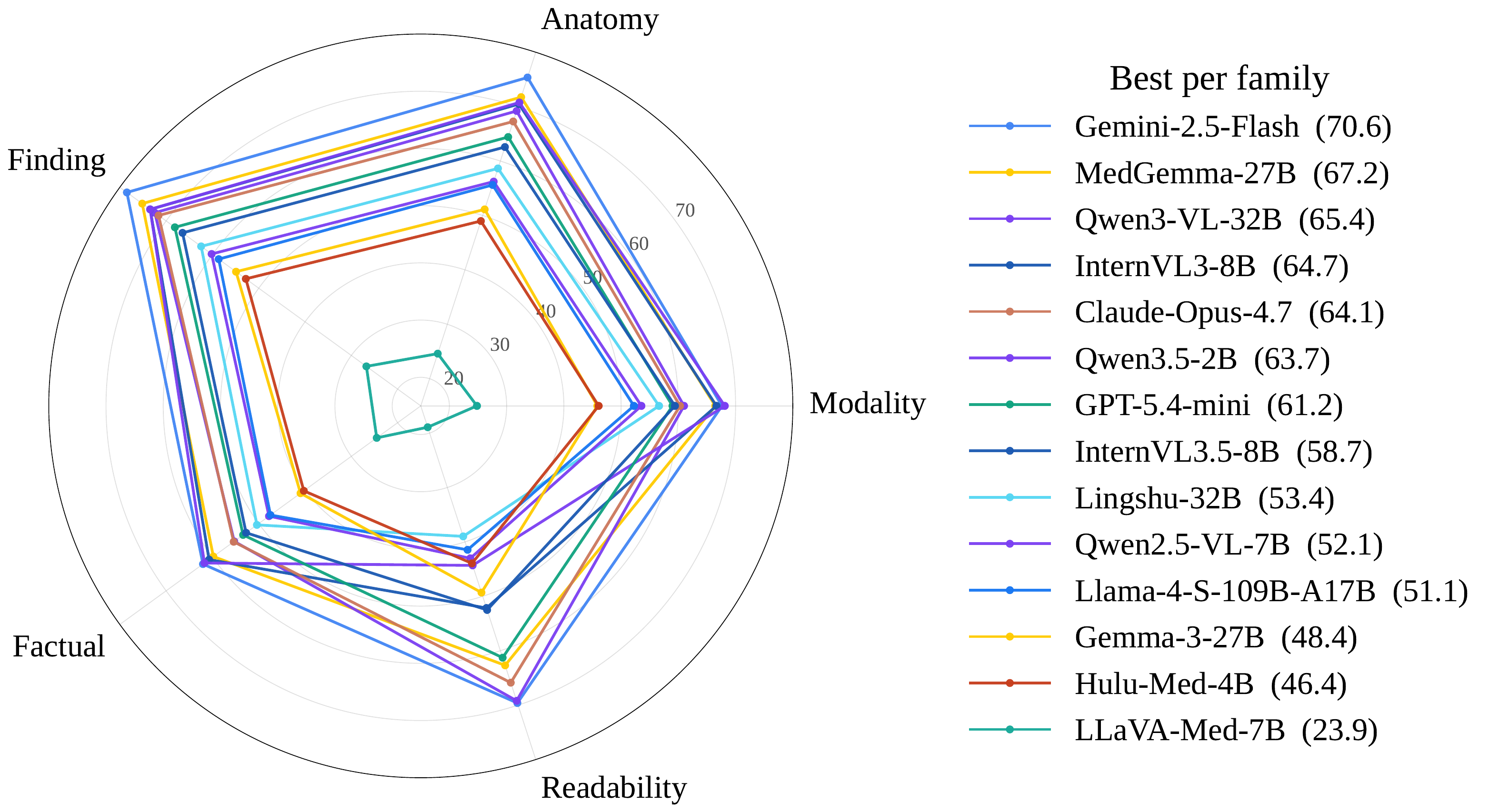}
\caption{\textbf{Per-attribute \ourseval{} profile for the best-scoring model in each family (Task~A, lay register, $n$\,=\,5,000).}
Line style distinguishes model categories: solid (proprietary), dashed (open-source generalist), dotted (medical).
Top-tier models cluster tightly across clinical fidelity attributes (Anatomy, Finding, Factual) but diverge on Readability.
Medical VLMs exhibit a characteristic Readability deficit relative to their clinical scores.}
\label{fig:radar_family}
\end{figure}

To complement the numerical leaderboard in \Cref{tab:leaderboard_full}, \Cref{fig:radar_family} visualizes the per-attribute \ourseval{} profile for the best-scoring model in each family, revealing that top-tier models cluster tightly on clinical fidelity attributes yet diverge sharply on Readability, a pattern difficult to discern from tabular scores alone.
We then present side-by-side lay caption predictions from four representative VLMs on four test-set samples spanning CT, MRI, Pathology, and Chest X-Ray.
Models are selected to cover distinct categories: Gemini~2.5~Flash (top-ranked proprietary), GPT-5.5 and Claude~Opus~4.7 (proprietary), and MedGemma-27B (top medical).
The \oursfull{} ground-truth lay caption is shown as reference.
All outputs are shown verbatim without post-processing, generated under Task~A (image + ROI only, no expert caption provided) with identical prompts and greedy decoding.
Predictions exceeding 200 words are truncated with ``\dots''; full outputs are available in the released repository.

\newpage

\begin{figure}[H]
\centering
\begin{minipage}[c]{0.20\linewidth}
  \centering
  \tcbox[boxrule=0.4pt, colback=white, colframe=black!60,
    arc=4pt, boxsep=0pt, left=0pt, right=0pt, top=0pt, bottom=0pt]{%
    \includegraphics[width=\linewidth]{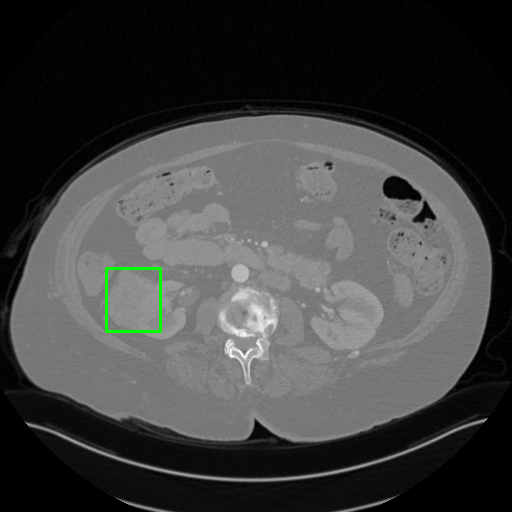}%
  }
\end{minipage}%
\hfill%
\begin{minipage}[c]{0.77\linewidth}
  \begin{tcolorbox}[
    enhanced, title={\scriptsize\textbf{Abdominal CT (FLARE23) --- Pancreatic Lesion}},
    colback=black!3, colframe=black!50, coltitle=white, colbacktitle=black!65,
    boxrule=0.3pt, arc=1.5pt, fonttitle=\bfseries\scriptsize,
    left=3pt, right=3pt, top=1pt, bottom=1pt]
    \scriptsize\textbf{Expert Caption} (\textit{124 words}):\enspace 
    The image is a transverse section of an abdominal CT scan, showing various organs such as the liver, spleen, kidneys, and segments of the intestines. The region of interest, located left-center and in the middle of the image, appears to have a different texture and density compared to the surrounding tissue, which may indicate a pathological change. This region's relative position to other organs suggests it could be affecting or affected by adjacent structures due to its proximity, potentially indicating a spread of disease or a localized process. The abnormal area's characteristics, such as its size and texture, are consistent with the presence of a pancreatic cancer, as described in the provided medical knowledge, which often presents as a low-density mass on CT scans.
  \end{tcolorbox}
\end{minipage}
\end{figure}

\vspace{-10pt}
\begin{tcolorbox}[
  enhanced, breakable,
  title={\small\textbf{Model Predictions (Lay Register, Task~A)}},
  colback=white, colframe=black!60, coltitle=white, colbacktitle=black!60,
  boxrule=0.3pt, arc=2pt, fonttitle=\bfseries\small,
  left=3pt, right=3pt, top=2pt, bottom=2pt]
  \noindent{\scriptsize\textcolor{black!70}{\textbf{\oursfull{} Ground Truth}}
  \hfill\textcolor{gray}{\textit{Reference $\cdot$ 126 words}}}\\[-3pt]
  {\scriptsize The image is a cross-section of a CT scan of the belly area, showing organs like the liver, spleen, kidneys, and parts of the intestines. The region of interest, located left-center and in the middle of the image, appears to have a different texture and density compared to the surrounding tissue, which may indicate a problem with the body's cells. This region's relative position to other organs suggests it could be affecting or affected by nearby structures due to its proximity, potentially indicating a spread of disease or a localized process. The abnormal area's characteristics, such as its size and texture, are consistent with the presence of pancreatic cancer (pancreatic carcinoma), as described in medical knowledge, which often appears as a low-density lump on CT scans.}
  \tcbline
  \noindent{\scriptsize\textcolor{geminiblue}{\textbf{Gemini 2.5 Flash}}
  \hfill\textcolor{gray}{\textit{Proprietary $\cdot$ 409 words}}}\\[-3pt]
  {\scriptsize Here's an explanation of your medical image: --- 1. What type of image this is and what body part is shown. This image is a CT scan (Computed Tomography scan). A CT scan is a special type of X-ray that takes detailed pictures, or "slices," of your body from different angles. It helps doctors see inside your body, including organs, bones, and soft tissues, much more clearly than a regular X-ray. This particular image shows a cross-section of your abdomen (belly area). Imagine looking at a slice of your body as if you were lying on your back and we're looking up from your feet. You can see your spine (backbone) in the center-back, along with various organs and muscles. 2. What the highlighted area shows and where it is located. The bright green box highlights an area within your right kidney. Your kidneys are two bean-shaped organs that sit on either side of your spine, just below your rib cage. Their main job is to filter waste from your blood and produce urine. Within the green box, you can see a mass or growth inside the kidney. This means there's an area of tissue that looks different from the\,\dots}
  \tcbline
  \noindent{\scriptsize\textcolor{black!80}{\textbf{GPT-5.5}}
  \hfill\textcolor{gray}{\textit{Proprietary $\cdot$ 128 words}}}\\[-3pt]
  {\scriptsize This is a CT scan (a detailed X-ray picture) of the abdomen, shown as a cross-section slice. It includes organs such as the intestines, spine, blood vessels, and kidneys. The green box is around the right kidney area. On CT images, the patient's right side appears on the left side of the picture. The highlighted region shows an area of kidney tissue/near-kidney tissue that looks different from the surrounding structures. This means there is a spot or area in the right kidney region that the doctor wants to evaluate. It could represent something benign, such as a cyst (a fluid-filled sac), or another type of lesion (an abnormal-looking area). The radiology report and sometimes additional imaging help determine exactly what it is and whether any follow-up is needed.}
  \tcbline
  \noindent{\scriptsize\textcolor{claudeorange}{\textbf{Claude Opus 4.7}}
  \hfill\textcolor{gray}{\textit{Proprietary $\cdot$ 368 words}}}\\[-3pt]
  {\scriptsize Understanding Your Medical Image 1. What This Image Is This is a CT scan (a special type of X-ray that takes detailed cross-section pictures of the inside of your body, like looking at a slice of bread). The image shows a view of your abdomen (belly area), taken as if you were lying on your back and we're looking up at you from your feet. You can see your spine at the bottom center, and your internal organs and intestines surrounding it. 2. What the Green Box Shows The green box highlights an area on the right side of your abdomen (which appears on the left side of the image because of how CT scans are viewed). This area is in the soft tissue near your kidney and the muscles of your back/side wall. Inside the box, there appears to be a slightly different-looking spot --- it looks a bit rounded and has a slightly different shade compared to the tissue around it, which is why your doctor wants to draw your attention to it. 3. What This Finding Means The highlighted spot represents an area that looks different from the surrounding normal tissue. This could be several things, such\,\dots}
  \tcbline
  \noindent{\scriptsize\textcolor{navyblue}{\textbf{MedGemma-27B}}
  \hfill\textcolor{gray}{\textit{Medical $\cdot$ 208 words}}}\\[-3pt]
  {\scriptsize Okay, let's break down this medical image. This is a CT (Computed Tomography) scan image. It's like a detailed X-ray that shows cross-sections of your body. This particular image shows a slice through the middle of your abdomen (belly), specifically focusing on the area around your kidneys, spleen, and part of your intestines. The green box highlights a specific area within the abdomen. It points to a structure that appears to be a kidney. The kidney is located towards the back of the abdomen, on either side of the spine. The highlighted kidney is on the left side in this image. The highlighted kidney appears to be larger than normal. This is called hydronephrosis (swelling of a kidney due to a build-up of urine). It suggests that there might be a blockage somewhere in the urinary tract (the system that carries urine from the kidneys to the bladder), preventing urine from draining properly from that kidney. This blockage could be caused by various things, like a kidney stone, a tumor, or a narrowing of the ureter (the tube that carries urine from the kidney to the bladder). Your doctor will likely need further tests to figure out the exact cause\,\dots}
\end{tcolorbox}

\clearpage

\vspace{-4pt}
\begin{figure}[H]
\centering
\begin{minipage}[c]{0.20\linewidth}
  \centering
  \tcbox[boxrule=0.4pt, colback=white, colframe=black!60,
    arc=4pt, boxsep=0pt, left=0pt, right=0pt, top=0pt, bottom=0pt]{%
    \includegraphics[width=\linewidth]{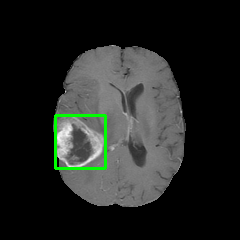}%
  }
\end{minipage}%
\hfill%
\begin{minipage}[c]{0.77\linewidth}
  \begin{tcolorbox}[
    enhanced, title={\scriptsize\textbf{Brain MRI T1ce (BraTS2018) --- Enhancing Tumor with Infiltration}},
    colback=black!3, colframe=black!50, coltitle=white, colbacktitle=black!65,
    boxrule=0.3pt, arc=1.5pt, fonttitle=\bfseries\scriptsize,
    left=3pt, right=3pt, top=1pt, bottom=1pt]
    \scriptsize\textbf{Expert Caption} (\textit{113 words}):\enspace 
    The image is a magnetic resonance imaging (MRI) scan of the brain, showing a cross-sectional view with various brain structures visible, such as the cerebral cortex, ventricles, and possibly the basal ganglia. The region of interest, located left-center horizontally and in the middle vertically, occupies approximately 4.6\% of the area and exhibits abnormal tissue characteristics indicative of an enhancing tumor, such as altered signal intensity, which may suggest the presence of a mass effect or contrast enhancement. This abnormal region is likely affecting or being affected by the surrounding brain tissue due to its mass effect or infiltration, which is common in brain tumors as they can disrupt normal brain function and anatomy.
  \end{tcolorbox}
\end{minipage}
\end{figure}

\vspace{-10pt}
\begin{tcolorbox}[
  enhanced, breakable,
  title={\small\textbf{Model Predictions (Lay Register, Task~A)}},
  colback=white, colframe=black!60, coltitle=white, colbacktitle=black!60,
  boxrule=0.3pt, arc=2pt, fonttitle=\bfseries\small,
  left=3pt, right=3pt, top=2pt, bottom=2pt]
  \noindent{\scriptsize\textcolor{black!70}{\textbf{\oursfull{} Ground Truth}}
  \hfill\textcolor{gray}{\textit{Reference $\cdot$ 115 words}}}\\[-3pt]
  {\scriptsize The image is an MRI scan of the brain, showing a cross-sectional view with various brain structures visible, such as the outer layer of the brain (cerebral cortex), brain ventricles, and possibly the basal ganglia. The region of interest, located left-center horizontally and in the middle vertically, occupies approximately 4.6\% of the area and exhibits abnormal tissue characteristics indicative of an enhancing lump (tumor), such as altered signal intensity, which may suggest the presence of a swelling or contrast enhancement. This abnormal region is likely affecting or being affected by the surrounding brain tissue due to its swelling or infiltration, which is common in brain lumps as they can disrupt normal brain function and anatomy.}
  \tcbline
  \noindent{\scriptsize\textcolor{geminiblue}{\textbf{Gemini 2.5 Flash}}
  \hfill\textcolor{gray}{\textit{Proprietary $\cdot$ 329 words}}}\\[-3pt]
  {\scriptsize Okay, let's walk through this image together. It's completely normal to have questions, and we'll explain everything in simple terms. --- 1. What type of image this is and what body part is shown. This is a special type of picture called an MRI (Magnetic Resonance Imaging). An MRI uses strong magnets and radio waves to create very detailed images of the soft tissues inside your body, like your brain. It's excellent for showing structures that don't show up well on regular X-rays. This particular image shows a "slice" or cross-section of your brain, as if we're looking down from the top of your head. You can see the different parts of the brain tissue. --- 2. What the highlighted area shows and where it is located. The bright green box highlights an area on the right side of your brain. Within that box, you can see an abnormal growth or mass. Doctors refer to this as a tumor (an abnormal collection of cells that can grow in the body). It looks irregular in shape and has different shades within it -- some brighter areas and some darker areas. The brighter areas often indicate where the growth is more active\,\dots}
  \tcbline
  \noindent{\scriptsize\textcolor{black!80}{\textbf{GPT-5.5}}
  \hfill\textcolor{gray}{\textit{Proprietary $\cdot$ 174 words}}}\\[-3pt]
  {\scriptsize This is an MRI scan (magnetic resonance imaging, a test that uses magnets to make detailed pictures) of the brain, shown as a horizontal ``slice'' through the head. The green box marks an area on one side of the brain that looks brighter and different from the surrounding brain tissue. On this type of brain MRI, a bright area like this can represent abnormal brain tissue. Because MRI images are usually displayed as if looking up from the patient's feet, the highlighted area is likely on the right side of the brain. This bright area may suggest a stroke (an area of brain injury caused by reduced blood flow), especially if this is a diffusion MRI sequence, where bright signal can mean restricted diffusion (water movement in the brain tissue is limited, often seen in recent injury). In simple terms, part of the brain in that area may not be getting normal blood flow or may have been recently injured. A radiologist and your doctor would interpret this together with symptoms and other scans.}
  \tcbline
  \noindent{\scriptsize\textcolor{claudeorange}{\textbf{Claude Opus 4.7}}
  \hfill\textcolor{gray}{\textit{Proprietary $\cdot$ 352 words}}}\\[-3pt]
  {\scriptsize Understanding Your Brain Image 1. What Type of Image This Is This is an MRI scan (Magnetic Resonance Imaging --- a type of picture that uses magnets and radio waves to show the inside of your body without using radiation) of your brain. You are looking at a horizontal "slice" of the brain, as if viewing it from above. The light gray areas are brain tissue, and the dark areas are spaces normally filled with fluid. 2. What the Highlighted Area Shows Inside the green box, on the right side of the image (which actually represents the left side of your brain --- brain images are flipped like a mirror), there is a bright, irregular white area surrounded by a darker zone. - The bright white spot is an abnormal mass or growth in the brain tissue. - The darker ring around it is edema (swelling caused by extra fluid leaking into the surrounding healthy brain tissue). - You can also see that the nearby fluid-filled spaces in the middle of the brain (called ventricles) look slightly pushed or squeezed, which is called a mass effect (when something takes up space and presses on nearby structures). 3. What This Finding\,\dots}
  \tcbline
  \noindent{\scriptsize\textcolor{navyblue}{\textbf{MedGemma-27B}}
  \hfill\textcolor{gray}{\textit{Medical $\cdot$ 196 words}}}\\[-3pt]
  {\scriptsize Okay, let's break down this medical image. This is a Magnetic Resonance Imaging (MRI) scan. MRI uses strong magnets and radio waves to create detailed pictures of the inside of your body. This particular scan shows your brain. The green box highlights a specific area within your brain. It shows a region that appears brighter than the surrounding brain tissue. This brighter area is located in the right side of your brain, specifically in the area called the basal ganglia (a group of structures deep within the brain involved in movement control, learning, and other functions). It looks like a somewhat irregular, bright spot. This bright spot in the highlighted area suggests a lesion (an area of abnormal tissue). Based on its appearance on this MRI, it could be several things, such as a stroke (damage to brain tissue due to lack of blood supply), a tumor (abnormal growth of cells), inflammation, or possibly an old injury. The exact nature of this lesion would require further investigation and correlation with your symptoms and medical history. Your doctor will use this image, along with other information, to determine the cause and plan the best course of action.}
\end{tcolorbox}

\clearpage

\vspace{-4pt}
\begin{figure}[H]
\centering
\begin{minipage}[c]{0.20\linewidth}
  \centering
  \tcbox[boxrule=0.4pt, colback=white, colframe=black!60,
    arc=4pt, boxsep=0pt, left=0pt, right=0pt, top=0pt, bottom=0pt]{%
    \includegraphics[width=\linewidth]{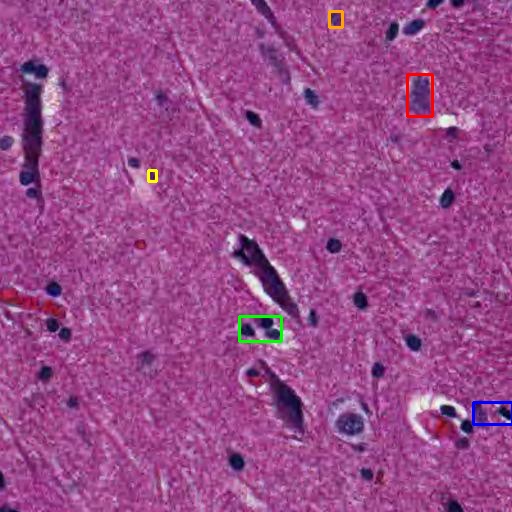}%
  }
\end{minipage}%
\hfill%
\begin{minipage}[c]{0.77\linewidth}
  \begin{tcolorbox}[
    enhanced, title={\scriptsize\textbf{Brain Histopathology (TCGA) --- Neoplastic Process}},
    colback=black!3, colframe=black!50, coltitle=white, colbacktitle=black!65,
    boxrule=0.3pt, arc=1.5pt, fonttitle=\bfseries\scriptsize,
    left=3pt, right=3pt, top=1pt, bottom=1pt]
    \scriptsize\textbf{Expert Caption} (\textit{137 words}):\enspace 
    The histological image displays a section of brain tissue, where the region of interest, located at the right-center horizontally and lower-middle vertically, occupies approximately 0.4\% of the area and is characterized by a cluster of abnormal nuclei that are indicative of a neoplastic process. This area, distinguished by its distinct cellular morphology, is situated in the brain parenchyma, away from the main tumor mass and does not involve any medical devices. The abnormal nuclei within this region suggest a potential neoplasm, which may be related to the surrounding brain tissue either as a secondary effect of the primary brain tumor or as a separate pathological process. The specific nature of these nuclei, which are not part of the main tumor mass, could imply a metastatic spread or a localized area of tumor activity within the brain tissue.
  \end{tcolorbox}
\end{minipage}
\end{figure}

\vspace{-10pt}
\begin{tcolorbox}[
  enhanced, breakable,
  title={\small\textbf{Model Predictions (Lay Register, Task~A)}},
  colback=white, colframe=black!60, coltitle=white, colbacktitle=black!60,
  boxrule=0.3pt, arc=2pt, fonttitle=\bfseries\small,
  left=3pt, right=3pt, top=2pt, bottom=2pt]
  \noindent{\scriptsize\textcolor{black!70}{\textbf{\oursfull{} Ground Truth}}
  \hfill\textcolor{gray}{\textit{Reference $\cdot$ 146 words}}}\\[-3pt]
  {\scriptsize The microscopic image shows a section of brain tissue, where the area of interest, located at the right-center horizontally and lower-middle vertically, takes up about 0.4\% of the area and has a group of abnormal cell centers (nuclei) that suggest an abnormal growth process (neoplastic process). This area, which looks different from the surrounding cells, is in the brain tissue itself, away from the main lump (tumor mass) and doesn't involve any medical devices. The abnormal nuclei within this region indicate a possible abnormal growth, which may be related to the surrounding brain tissue either as a secondary effect of the primary brain tumor or as a separate disease process. The specific nature of these nuclei, which are not part of the main lump (tumor mass), could imply that cancer has spread (metastatic spread) or there's a localized area of abnormal growth within the brain tissue.}
  \tcbline
  \noindent{\scriptsize\textcolor{geminiblue}{\textbf{Gemini 2.5 Flash}}
  \hfill\textcolor{gray}{\textit{Proprietary $\cdot$ 266 words}}}\\[-3pt]
  {\scriptsize Okay, let's look at this image together. 1. What type of image this is and what body part is shown. This is a microscopic image, which means it's a picture taken through a very powerful magnifying glass. To get this image, a very thin slice of tissue was taken from your body and specially stained with colors (pink and purple) so that different parts of the cells become visible. This particular image shows a small piece of your brain tissue. 2. What the highlighted area shows and where it is located. The green box highlights a single cell (the basic building block of your body). Inside this cell, you can see two distinct, dark purple, somewhat rounded structures. This cell is located within the brain tissue itself, surrounded by many other brain cells and the supporting material between them. 3. What this finding means in simple terms. What the green box shows is a cell that is in the process of dividing. This means the cell is making a copy of itself to create two new cells. Cells divide for many normal reasons in the body: to grow, to repair damaged tissue, or to replace old cells. In the brain,\,\dots}
  \tcbline
  \noindent{\scriptsize\textcolor{black!80}{\textbf{GPT-5.5}}
  \hfill\textcolor{gray}{\textit{Proprietary $\cdot$ 154 words}}}\\[-3pt]
  {\scriptsize This is a microscope image of a tissue slide stained with H\&E stain (a common dye that makes cells and tissue easier to see). It shows brain tissue, specifically the pink background material where brain cells and their connections are located. The green box highlights a small, pale ``empty-looking'' space within the brain tissue background. This is called a vacuole (a tiny clear space or hole seen under the microscope). It is located between brain cells in the surrounding nerve tissue, called the neuropil (the network of nerve fibers and connections between brain cells). The highlighted area suggests a small area where the brain tissue looks ``spongy'' or has developed tiny spaces. Pathologists call this spongiform change (a sponge-like appearance caused by many tiny holes in the tissue). This kind of change can be seen in certain brain diseases, but the exact meaning depends on the full slide, other tests, and the patient's symptoms.}
  \tcbline
  \noindent{\scriptsize\textcolor{claudeorange}{\textbf{Claude Opus 4.7}}
  \hfill\textcolor{gray}{\textit{Proprietary $\cdot$ 349 words}}}\\[-3pt]
  {\scriptsize Understanding Your Medical Image 1. What This Image Shows This is a histology slide (a very thin slice of body tissue stained with special dyes and viewed under a microscope). The pink and purple colors come from a common stain called H\&E, which helps doctors see the different parts of cells clearly. The tissue shown here is from the brain --- specifically, it appears to be brain tissue showing the supportive background material and various cells that make up brain structure. 2. What the Highlighted Area Shows Inside the green box, located in the lower-middle portion of the image, you can see a small dark, elongated structure. This is a neuron (a brain nerve cell that sends and receives signals) or possibly a small blood vessel cut lengthwise. The dark purple/blue staining shows the nucleus (the control center of the cell that holds genetic material), and the surrounding pink area is the cell body. The cell sits within a sea of pink fibers, which is the neuropil (the dense network of nerve fiber connections that fills the spaces between brain cells). 3. What This Finding Means in Simple Terms The structure highlighted in the green box appears to be a\,\dots}
  \tcbline
  \noindent{\scriptsize\textcolor{navyblue}{\textbf{MedGemma-27B}}
  \hfill\textcolor{gray}{\textit{Medical $\cdot$ 215 words}}}\\[-3pt]
  {\scriptsize Okay, let's break down this medical image. This is a microscopic image, meaning it was taken using a microscope. It shows a sample of brain tissue. You can see the different types of cells that make up the brain, like the roundish neurons (nerve cells) and the supporting cells (glia). The green box highlights a specific area within the brain tissue. This area shows a collection of cells that look different from the surrounding normal brain tissue. These cells are tightly packed together, have large, dark nuclei (the control center of the cell), and appear somewhat abnormal in shape and arrangement. This cluster is located within the brain tissue, likely within the white matter (the area with many nerve fibers). The highlighted area shows a cluster of abnormal cells. This pattern is highly suggestive of a type of brain tumor called a glioma (a tumor arising from glial cells, the supporting cells of the brain). Specifically, the appearance is often associated with a high-grade glioma, like glioblastoma (a fast-growing, aggressive type of brain tumor). The cells are dividing rapidly and don't look like normal brain cells. This finding would require further investigation, likely including more detailed analysis and possibly imaging\,\dots}
\end{tcolorbox}

\clearpage

\vspace{-4pt}
\begin{figure}[H]
\centering
\begin{minipage}[c]{0.20\linewidth}
  \centering
  \tcbox[boxrule=0.4pt, colback=white, colframe=black!60,
    arc=4pt, boxsep=0pt, left=0pt, right=0pt, top=0pt, bottom=0pt]{%
    \includegraphics[width=\linewidth]{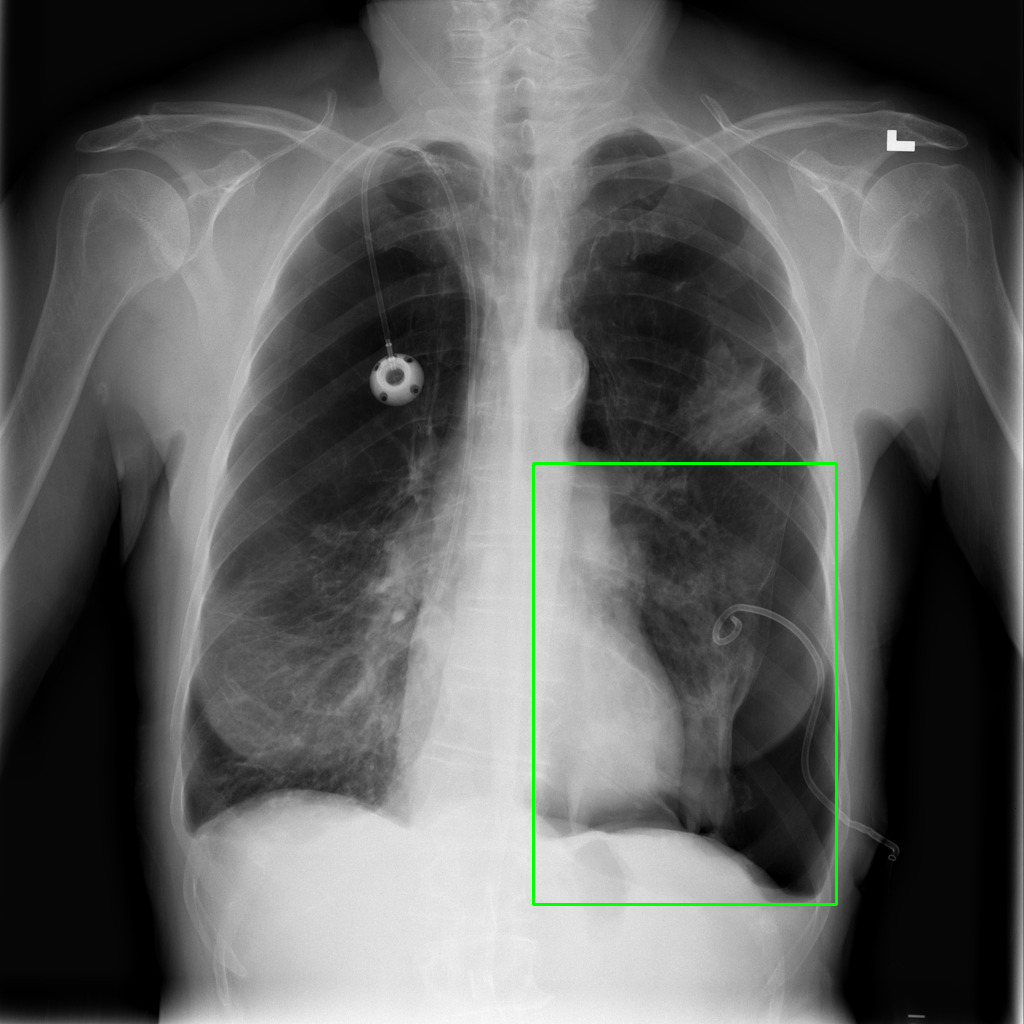}%
  }
\end{minipage}%
\hfill%
\begin{minipage}[c]{0.77\linewidth}
  \begin{tcolorbox}[
    enhanced, title={\scriptsize\textbf{Chest X-Ray (NIH CXR) --- Pneumothorax}},
    colback=black!3, colframe=black!50, coltitle=white, colbacktitle=black!65,
    boxrule=0.3pt, arc=1.5pt, fonttitle=\bfseries\scriptsize,
    left=3pt, right=3pt, top=1pt, bottom=1pt]
    \scriptsize\textbf{Expert Caption} (\textit{180 words}):\enspace 
    The image is a chest X-ray showing the thoracic cavity with the lungs, heart, and the bony structures of the chest wall. A region of interest is located at right-center part of the image horizontally and lower-middle part of image vertically, occupying approximately 12.7\% of the total area. The lungs are the primary organs visible, with the heart situated centrally between them. The region of interest is located in the lower right lung, adjacent to the diaphragm. Within this region, there is a visible translucent area without lung markings, indicative of pneumothorax. This area is unusual due to its lack of vascular and bronchial structures typically seen in healthy lung tissue, suggesting a collapsed lung segment. The affected area's position relative to the rest of the lung tissue suggests that it could be the result of air accumulating in the pleural space, potentially causing a portion of the lung to collapse. This could be due to a rupture in the lung tissue or a breach in the chest wall, leading to the abnormal collection of air in the pleural space.
  \end{tcolorbox}
\end{minipage}
\end{figure}

\vspace{-10pt}
\begin{tcolorbox}[
  enhanced, breakable,
  title={\small\textbf{Model Predictions (Lay Register, Task~A)}},
  colback=white, colframe=black!60, coltitle=white, colbacktitle=black!60,
  boxrule=0.3pt, arc=2pt, fonttitle=\bfseries\small,
  left=3pt, right=3pt, top=2pt, bottom=2pt]
  \noindent{\scriptsize\textcolor{black!70}{\textbf{\oursfull{} Ground Truth}}
  \hfill\textcolor{gray}{\textit{Reference $\cdot$ 191 words}}}\\[-3pt]
  {\scriptsize The image is a chest X-ray showing the chest area with the lungs, heart, and the bony structures of the chest wall. A region of interest is located at right-center part of the image horizontally and lower-middle part of image vertically, occupying approximately 12.7\% of the total area. The lungs are the primary organs visible, with the heart situated centrally between them. The region of interest is located in the lower right lung, adjacent to the diaphragm. Within this region, there is a visible translucent area without lung markings, indicative of air in the space around the lung (pneumothorax). This area is unusual due to its lack of blood vessels and airways typically seen in healthy lung tissue, suggesting a collapsed lung segment. The affected area's position relative to the rest of the lung tissue suggests that it could be the result of air accumulating in the space around the lung, potentially causing a portion of the lung to collapse. This could be due to a rupture in the lung tissue or a breach in the chest wall, leading to the abnormal collection of air in the space around the lung.}
  \tcbline
  \noindent{\scriptsize\textcolor{geminiblue}{\textbf{Gemini 2.5 Flash}}
  \hfill\textcolor{gray}{\textit{Proprietary $\cdot$ 219 words}}}\\[-3pt]
  {\scriptsize Here is an explanation of your medical image: This is a chest X-ray, which is a special type of image that uses a small amount of radiation to create a picture of the inside of your chest. It shows your lungs, heart, ribs, and the surrounding structures. The green box highlights the lower and middle part of your right lung (the lung on your right side). In this area, instead of the usual clear, dark appearance of healthy lung tissue, we can see some cloudy or hazy patches. There is also a thin, curving tube visible within this area, which is a chest tube (a flexible tube inserted into the chest to remove air or fluid). The cloudy or hazy patches in your right lung suggest there is some inflammation or fluid buildup. This could be due to an infection, like pneumonia (an infection that inflames the air sacs in one or both lungs), or possibly some collapse of lung tissue (called atelectasis). The chest tube is in place to help drain any fluid or air that might be causing these issues, which helps your lung to heal and expand properly. Your doctor will use this X-ray, along with your\,\dots}
  \tcbline
  \noindent{\scriptsize\textcolor{black!80}{\textbf{GPT-5.5}}
  \hfill\textcolor{gray}{\textit{Proprietary $\cdot$ 187 words}}}\\[-3pt]
  {\scriptsize This is a chest X-ray (a picture made using a small amount of radiation). It shows the chest, including both lungs, the heart, ribs, and the upper part of the diaphragm (the breathing muscle under the lungs). The green box is around the lower part of the left side of the chest. In this area, there appears to be a very dark space along the outside/lower part of the lung, with a curved tube visible nearby. This is most consistent with a pneumothorax (air trapped in the space between the lung and chest wall), and the tube may be a chest drain (a tube used to remove air or fluid from around the lung). Air seems to have collected where it should not be---around the lung rather than inside it. This can press on the lung and make it partly collapse, which may cause shortness of breath or chest pain. The tube is likely there to let the trapped air out so the lung can expand again. Your medical team would use this X-ray to check the position of the tube and whether the lung is re-expanding.}
  \tcbline
  \noindent{\scriptsize\textcolor{claudeorange}{\textbf{Claude Opus 4.7}}
  \hfill\textcolor{gray}{\textit{Proprietary $\cdot$ 377 words}}}\\[-3pt]
  {\scriptsize Understanding Your Chest X-Ray 1. What This Image Shows This is a chest X-ray (a picture taken using small amounts of radiation to see inside your body). It shows your chest area, including your lungs, heart, ribs, collarbones, and the upper part of your spine. You can also see some medical devices on the image --- a round object on the upper left side appears to be a port (a small device placed under the skin to give medications) and there are some thin lines that look like tubes or wires. 2. What the Highlighted Area Shows The green box is drawn around the lower portion of your left lung and the area near your heart (on the right side of the image, but this is actually the left side of your body --- X-rays are viewed as if you're facing the doctor). In this area, you can see: - A hazy or cloudy white area instead of the dark, clear appearance that healthy lung tissue normally shows - The outline of the heart and diaphragm (the dome-shaped muscle below your lungs that helps you breathe) appears less sharp than it should - A curved tube-like device is visible in\,\dots}
  \tcbline
  \noindent{\scriptsize\textcolor{navyblue}{\textbf{MedGemma-27B}}
  \hfill\textcolor{gray}{\textit{Medical $\cdot$ 204 words}}}\\[-3pt]
  {\scriptsize Okay, let's break down this medical image. This is a chest X-ray (a type of radiograph, which uses X-rays to create an image). It shows the inside of your chest, including your lungs, heart, ribs, and some of the upper abdomen. The green box highlights a specific area within your right lung. Inside this box, you can see a dense, white, irregular-shaped area. This area appears to be a mass or consolidation (a filling of the air spaces in the lung with fluid, cells, or other substances). It's located in the upper part of your right lung. This finding, a mass or consolidation in the lung, is concerning. It could represent several things, including: * Infection: Like pneumonia (inflammation of the lung tissue), which can cause consolidation. * Tumor: A cancerous or non-cancerous growth. * Other conditions: Such as inflammation or ...}
\end{tcolorbox}

\section{Reproducibility and Hyperparameters}
\label{app:repro}

All code is released at the anonymous repository \url{https://anonymous.4open.science/r/MedLayXPlain-5D27}.
Every setting in this section was extracted from the production scripts included therein.

\paragraph{Evaluated Models.}
We benchmark 33 VLM configurations spanning four categories (\Cref{tab:leaderboard_full}).
\textit{Proprietary}~(6): GPT-5.5~\cite{openai2026gpt55}, GPT-5.4-mini~\cite{openai2026gpt54}, Claude Opus~4.7~\cite{anthropic2026opus47}, Claude Haiku~4.5~\cite{anthropic2025haiku45}, Gemini~3~Flash~\cite{google2026gemini3}, and Gemini~2.5~Flash~\cite{google2025gemini25}.
\textit{Open-source generalist}~(21): Qwen3.5-\{0.8B, 2B, 4B, 9B, 27B\}~\cite{qwen3.5}, Qwen3-VL-\{2B, 4B, 8B, 30B-A3B, 32B\}~\cite{bai2025qwen3}, Qwen3.6-35B-A3B~\cite{qwen36_35b_a3b} (with and without thinking), Qwen2.5-VL-7B~\cite{wang2024qwen2}, InternVL3-8B and InternVL3.5-\{2B, 4B, 8B, 14B, 38B\}~\cite{zhu2025internvl3}, Gemma-4-31B~\cite{gemmateam2025gemma3technicalreport}, and Llama-4-Scout~\cite{meta2025llama4}.
\textit{Medical}~(6): MedGemma-\{4B, 27B\}~\cite{sellergren2025medgemma}, Lingshu-\{7B, 32B\}~\cite{xu2025lingshu}, Hulu-Med-4B~\cite{jiang2025hulu}, and LLaVA-Med-7B~\cite{li2023llava}.
All models are evaluated under identical three-segment prompt templates with temperature~0 and max 1,024 output tokens.
Prompt templates are provided in \Cref{app:prompt}.
\ourseval{} (Qwen2.5-VL-3B) belongs to a different model generation than any benchmark participant, mitigating evaluation circularity.

\begin{table}[h!]
\centering
\caption{Hyperparameters for HOVER Step~2 constrained rewriting.}
\label{tab:hyper_step2}
\begin{tabular}{ll}
\toprule
Model                   & Llama-3.1-70B-Instruct~\cite{grattafiori2024llama} \\
Backend                 & vLLM~\cite{kwon2023efficient} \\
Temperature             & 0.3 \\
Top-$p$                 & 0.95 \\
Max generation tokens   & 768 \\
Max retries on failure  & 3 \\
Timeout per request     & 180\,s \\
Concurrency             & Sequential (checkpoint resume every 50 rows) \\
Sentence-count constraint & Soft (recorded for downstream verifier) \\
\bottomrule
\end{tabular}
\end{table}

\begin{table}[h!]
\centering
\caption{Hyperparameters for HOVER Step~3 cross-model visual verification.}
\label{tab:hyper_step3}
\begin{tabular}{ll}
\toprule
Model                   & Qwen3.5-27B \\
Backend                 & SGLang~\cite{zheng2024sglang} (OpenAI-compatible vision endpoint) \\
Tensor parallelism      & 8 \\
Context length          & 131,072 tokens \\
Temperature             & 0.1 \\
Max generation tokens   & 2,048 \\
Thinking mode           & Disabled \\
Image preprocessing     & Longest edge $\leq 1,024$\,px (Lanczos); JPEG quality\,=\,85 \\
Async concurrency       & 32 \\
Per-request timeout     & 300\,s (3 retries) \\
Pass threshold          & $S = 10/10$ \\
Fail threshold          & $S < 7/10$ \\
Max revision retries    & 2 \\
\bottomrule
\end{tabular}
\end{table}

\begin{table}[h!]
\centering
\caption{Hyperparameters for \ourseval{} distillation training.}
\label{tab:hyper_medlayeval}
\begin{tabular}{ll}
\toprule
Teacher                 & Qwen3.5-27B verifier (frozen labels from HOVER Step~3) \\
Student backbone        & Qwen2.5-VL-3B-Instruct + LoRA \\
LoRA rank / alpha       & 16 / 32 \\
LoRA target modules     & All attention and feed-forward projections (q, k, v, o, gate, up, down) \\
LoRA dropout            & 0.05 \\
Regression head         & Linear($d$, 256) $\to$ GELU $\to$ Dropout(0.1) $\to$ Linear(256, 5) $\to$ Sigmoid \\
Head precision          & float32 (backbone in bfloat16) \\
Loss                    & Inverse-frequency weighted MSE on $[0, 1]$ targets \\
Optimizer               & AdamW, weight decay\,=\,0.01, gradient clipping\,=\,1.0 \\
Learning rate           & $1 \times 10^{-5}$, cosine schedule \\
Warmup                  & 200 linear steps \\
Per-device batch size   & 4 \\
Gradient accumulation   & 4 \\
Distributed training    & 8-GPU DDP (NCCL) \\
Effective batch size    & $4 \times 8 \times 4 = 128$ \\
Epochs                  & 3 \\
Max sequence length     & 2,048 \\
Image max resolution    & $448 \times 448$ \\
Caption truncation      & 1,500 characters per caption \\
Mixed precision         & bfloat16 backbone, float32 head \\
Attention implementation & SDPA \\
GPUs                    & $4 \times$ NVIDIA B200 \\
\bottomrule
\end{tabular}
\end{table}

\begin{table}[h!]
\centering
\caption{Inference settings applied uniformly across all benchmarked VLMs.
The inference backend defaults to vLLM where the architecture is supported, falling back to HuggingFace Transformers otherwise.
Data is sharded 8-way across GPUs.}
\label{tab:inference_settings}
\begin{tabular}{ll}
\toprule
Decoding            & Temperature\,=\,0.0, max tokens\,=\,1,024, greedy \\
Image preprocessing & Longest-edge resize to 512\,px (bicubic) \\
vLLM settings       & bfloat16, GPU memory utilisation\,=\,0.9, max concurrent seqs\,=\,8 \\
HF fallback         & bfloat16, single-GPU per shard \\
Batch size (vLLM)   & 16 prompts per call (1 image per prompt) \\
Tensor parallelism  & 1 (default); 4 for Llama-4-Scout-17B-16E \\
Data sharding       & 8 shards over the 24,590-sample test split \\
\bottomrule
\end{tabular}
\end{table}

\vspace{-8pt}
\section{Comprehensive Benchmark Analysis}
\label{app:bench}
\vspace{-4pt}

\subsection{Thinking-Mode Reasoning Trace}
\label{app:thinking-trace}

Qwen3.6-35B-A3B is the only benchmarked model evaluated in both thinking and non-thinking configurations.
The non-thinking variant achieves a substantially higher \ourseval{} score (61.5 vs 47.7), a gap of 13.8 points that makes it the single largest configuration-level difference in the benchmark.
Two factors drive this:
(i)~budget consumption, as chain-of-thought deliberation consumes the 1,024-token budget before the patient-facing answer is produced;
and (ii)~ungrounded self-talk, where visible deliberation loops do not converge because the model lacks an external grounding signal, producing self-correcting anatomical reasoning that never reaches a final answer.
This finding suggests that reasoning-mode inference, while beneficial for logic-heavy tasks, is counterproductive for visually grounded lay description generation under constrained token budgets.

\subsection{Reference-Based Metric Analysis}
\label{app:metric_disagreement}

The main text identifies systematic divergence between reference-based metrics and \ourseval{} (\S\ref{sec:results}).
Here we analyze each metric family in detail.

\paragraph{GREEN penalizes lay register.}
GREEN~\cite{ostmeier2024green} detects clinically significant errors by comparing generated text against expert-level references.
Because it is trained on radiology reports written in professional register, replacing expert terms with plain-language equivalents is flagged as missing clinical content.
This explains the inverted ranking: models that retain more expert jargon (e.g., Qwen3.6-35B-A3B, GREEN\,=\,28.7) score higher than models that successfully simplify (e.g., Gemini~2.5~Flash, GREEN\,=\,18.5; Claude~Opus~4.7, GREEN\,=\,13.6).
Across all 33 models, GREEN and \ourseval{} Readability are negatively correlated, confirming that GREEN actively penalizes the simplification that lay captioning requires.

\paragraph{RaTEScore lacks discriminative power.}
RaTEScore~\cite{zhao2024ratescore} performs entity-level matching robust to medical synonyms.
While this design is valuable for expert report evaluation, it provides minimal separation for the expert-to-lay task: 27 of 33 models fall within a narrow 48--55 band.
This compression occurs because entity preservation is largely maintained across all models regardless of register, as even poorly simplified captions retain the core medical entities.

\paragraph{BLEU-4 and ROUGE-L reflect lexical shift.}
N-gram overlap metrics are structurally misaligned with simplification, where replacing clinical terms with plain-language equivalents necessarily reduces overlap with expert references.
The overall BLEU-4 range (1.4--6.4) is too compressed to meaningfully rank models, and high scores paradoxically indicate failure: InternVL3.5-2B achieves the top BLEU-4 (6.4) and ROUGE-L (26.1) by reproducing expert vocabulary without simplification.

\paragraph{Flesch--Kincaid grade reflects verbosity.}
FK grade measures surface-level readability via sentence length and syllable count.
Models producing shorter outputs achieve lower FK grades (Qwen3.5-4B: 6.4) regardless of actual comprehensibility, while verbose models score higher (LLaVA-Med: 16.3).
FK does not correlate with \ourseval{} Readability: Qwen3.5-4B scores only 38.4 on \ourseval{} Readability despite its low FK, because brevity alone does not ensure comprehensible explanations.

\paragraph{SARI is near-uniform.}
SARI~\cite{xu2016optimizing} measures edit operations against a reference simplification.
All 33 models fall within a 5.5-point range (24.6--30.1), offering negligible separation between the top-ranked and mid-tier models on this task.


\end{document}